\documentclass[11pt]{article}
\usepackage{pgfplots}
\pgfplotsset{compat=1.18}
\usetikzlibrary{patterns, calc, positioning}
\usepackage{acl}
\usepackage{times}
\usepackage{latexsym}
\usepackage[T1]{fontenc}
\usepackage[utf8]{inputenc}
\usepackage{microtype}
\usepackage{inconsolata}
\usepackage{graphicx}
\usepackage{subcaption}
\usepackage{booktabs}
\usepackage{svg}
\usepackage{multirow}
\usepackage{authblk}

\usepackage{amsmath,amsfonts,bm}

\def\eqref#1{equation~\ref{#1}}

\def\1{\bm{1}}

\DeclareMathAlphabet{\mathsfit}{\encodingdefault}{\sfdefault}{m}{sl}
\SetMathAlphabet{\mathsfit}{bold}{\encodingdefault}{\sfdefault}{bx}{n}

\usepackage{array}
\usepackage{graphicx}     
\usepackage[export]{adjustbox}
\usepackage{enumitem}     
\usepackage{url}
\usepackage{wrapfig}
\usepackage{textcomp}
\usepackage{float}
\usepackage{amsmath}
\usepackage{xcolor}
\usepackage{amssymb} 
\usepackage{tikz}
\usepackage{stackengine}
\usepackage{afterpage}
\usepackage[table]{xcolor} 
\usepackage{authblk} 
\usepackage{natbib}
\usepackage{caption} 
\usepackage[many]{tcolorbox}
\definecolor{headercolor}{rgb}{0.9, 0.9, 0.9} 
\definecolor{groupcolor}{rgb}{0.97, 0.97, 0.97} 
\definecolor{highlightcolor}{rgb}{0.90, 0.95, 1.0} 

\usepackage{tabularray}
\UseTblrLibrary{functional}
\UseTblrLibrary{booktabs}
\usepackage{makecell}
\usepackage{siunitx}
\UseTblrLibrary{siunitx}
\usepackage{longtable}

\usepackage{listings}
\definecolor{v3blue}{HTML}{378ADD}
\definecolor{r1coral}{HTML}{D85A30}
\definecolor{gapgreen}{HTML}{1D9E75}
\definecolor{gapred}{HTML}{E24B4A}

\title{Reinforcement Learning Improves Traversal of \\ Parametric Knowledge in LLMs}
\date{\today}

\author{
  \textbf{Renfei Zhang}\textsuperscript{1}, 
  \textbf{Manasa Kaniselvan}\textsuperscript{2}\protect\thanks{Work partly done at Meta.}, 
  \textbf{Rylan Schaeffer}\textsuperscript{3}, 
  \textbf{Niloofar Mireshghallah}\textsuperscript{1,*}\\ 
  \textsuperscript{1}Carnegie Mellon University, 
  \textsuperscript{2}MIT, 
  \textsuperscript{3}Stanford University \\
  \small{\textbf{Correspondence:} \texttt{\{az5, nmireshg\}@andrew.cmu.edu}}
}

\begin{document}
\pagestyle{empty}
\pagenumbering{gobble}
\maketitle
\begin{abstract}
Reinforcement learning (RL) is often credited with improving reasoning at the expense of factual knowledge. We instead find that reasoning models outperform their instruction-tuned versions on factual recall by accessing existing parametric knowledge more effectively. Across five model families, structured prompting—which explicitly guides models through hierarchical traversal—recovers most of this gap, suggesting
  that much of the missing knowledge is latent rather than absent. Controlled RL experiments further support this: training on unseen, non-extractable facts improves recall of held-out, frequent but previously inaccessible facts, ruling out simple data exposure. Decomposing the training objective further attributes this gain to iterated on-policy exploration. The same mechanism appears
  behaviorally and internally: the reasoning advantage grows with retrieval depth,
  while layerwise analysis finds similar factual representations but divergent
  query representations. Distilled models, in contrast, often imitate
  self-correction without acquiring the exploration needed for navigation. Together, these findings suggest that improving factual recall in LLMs depends not only on expanding what models know but also on teaching them to navigate it—motivating future post-training methods that optimize traversal.
\end{abstract}

\afterpage{\afterpage{%
\begin{table*}[!t]
\centering
\caption{Majority vote accuracy comparison of Instruct vs. Reasoning models on PopQA, IPC, and MedConceptsQA datasets across three prompt templates (QA, CoT, Structured). $\Delta$ rows show the reasoning advantage; \colorbox{orange!30}{orange} shading intensity reflects gap magnitude. Bold indicates best within each model pair.}
\label{tab:majority_vote_compact}
\scalebox{0.70}{%
\begin{tabular}{l l | ccc | ccc | ccc}
\toprule
& & \multicolumn{3}{c|}{\textbf{PopQA}} & \multicolumn{3}{c|}{\textbf{IPC}} & \multicolumn{3}{c}{\textbf{MedConceptsQA}} \\
\textbf{Model} & \textbf{Type} & \textbf{QA} & \textbf{CoT} & \textbf{Struct.} & \textbf{QA} & \textbf{CoT} & \textbf{Struct.} & \textbf{QA} & \textbf{CoT} & \textbf{Struct.} \\
\midrule
\multirow{3}{*}{\textbf{Qwen2.5-32B}}
& Instruct  & 0.252          & 0.288          & 0.310          & 0.300 & \textbf{0.370} & \textbf{0.400} & 0.379          & 0.475          & 0.469 \\
& Reasoning & \textbf{0.386} & \textbf{0.380} & \textbf{0.375} & \textbf{0.350}          & 0.360          & 0.360          & \textbf{0.482} & \textbf{0.513} & \textbf{0.505} \\
& $\Delta$  & \cellcolor{orange!62}\textbf{+0.134} & \cellcolor{orange!52}\textbf{+0.092} & \cellcolor{orange!42}\textbf{+0.065} & \cellcolor{orange!28}\textbf{+0.050} & \cellcolor{orange!10}\textbf{-0.010} & \cellcolor{orange!10}\textbf{-0.040} & \cellcolor{orange!46}\textbf{+0.103} & \cellcolor{orange!34}\textbf{+0.038} & \cellcolor{orange!32}\textbf{+0.036} \\
\cmidrule{1-11}
\multirow{3}{*}{\textbf{Qwen3-8B}}
& Instruct  & 0.205          & 0.252          & \textbf{0.262} & 0.266          & 0.277          & 0.372          & 0.338          & 0.337          & 0.406 \\
& Reasoning & \textbf{0.237} & \textbf{0.267} & \textbf{0.262} & \textbf{0.298} & \textbf{0.394} & \textbf{0.394} & \textbf{0.400} & \textbf{0.424} & \textbf{0.421} \\
& $\Delta$  & \cellcolor{orange!18}\textbf{+0.032} & \cellcolor{orange!12}\textbf{+0.015} & \cellcolor{orange!10}\textbf{+0.000} & \cellcolor{orange!18}\textbf{+0.032} & \cellcolor{orange!50}\textbf{+0.117} & \cellcolor{orange!12}\textbf{+0.022} & \cellcolor{orange!35}\textbf{+0.062} & \cellcolor{orange!42}\textbf{+0.087} & \cellcolor{orange!22}\textbf{+0.015} \\
\cmidrule{1-11}
\multirow{3}{*}{\textbf{Qwen3-30B-A3B}}
& Instruct  & 0.231          & 0.279          & 0.302          & 0.309          & 0.319          & 0.436          & 0.398          & 0.404          & \textbf{0.489} \\
& Reasoning & \textbf{0.288} & \textbf{0.317} & \textbf{0.312} & \textbf{0.426} & \textbf{0.372} & \textbf{0.457} & \textbf{0.498} & \textbf{0.506} & 0.476 \\
& $\Delta$  & \cellcolor{orange!32}\textbf{+0.058} & \cellcolor{orange!22}\textbf{+0.037} & \cellcolor{orange!12}\textbf{+0.010} & \cellcolor{orange!50}\textbf{+0.117} & \cellcolor{orange!28}\textbf{+0.053} & \cellcolor{orange!12}\textbf{+0.021} & \cellcolor{orange!50}\textbf{+0.100} & \cellcolor{orange!51}\textbf{+0.102} & \cellcolor{orange!10}\textbf{-0.013} \\
\cmidrule{1-11}
\multirow{3}{*}{\textbf{Qwen3-235B-A22B}}
& Instruct  & 0.429          & 0.491          & 0.541          & 0.372          & 0.383          & \textbf{0.596} & 0.542          & 0.548          & \textbf{0.631} \\
& Reasoning & \textbf{0.494} & \textbf{0.562} & \textbf{0.567} & \textbf{0.457} & \textbf{0.404} & 0.574          & \textbf{0.641} & \textbf{0.656} & 0.580 \\
& $\Delta$  & \cellcolor{orange!42}\textbf{+0.065} & \cellcolor{orange!50}\textbf{+0.071} & \cellcolor{orange!18}\textbf{+0.026} & \cellcolor{orange!42}\textbf{+0.085} & \cellcolor{orange!12}\textbf{+0.021} & \cellcolor{orange!10}\textbf{-0.022} & \cellcolor{orange!44}\textbf{+0.099} & \cellcolor{orange!46}\textbf{+0.108} & \cellcolor{orange!10}\textbf{-0.051} \\
\cmidrule{1-11}
\multirow{3}{*}{\textbf{DeepSeek-V3}}
& Instruct  & 0.430          & 0.505          & \textbf{0.569} & 0.415          & 0.596          & 0.553          & 0.541          & 0.632          & 0.717 \\
& Reasoning & \textbf{0.513} & \textbf{0.557} & \textbf{0.569} & \textbf{0.638} & \textbf{0.670} & \textbf{0.574} & \textbf{0.778} & \textbf{0.790} & \textbf{0.792} \\
& $\Delta$  & \cellcolor{orange!50}\textbf{+0.083} & \cellcolor{orange!32}\textbf{+0.052} & \cellcolor{orange!10}\textbf{+0.000} & \cellcolor{orange!50}\textbf{+0.223} & \cellcolor{orange!38}\textbf{+0.074} & \cellcolor{orange!12}\textbf{+0.021} & \cellcolor{orange!75}\textbf{+0.237} & \cellcolor{orange!57}\textbf{+0.158} & \cellcolor{orange!38}\textbf{+0.075} \\
\cmidrule{1-11}
\multirow{3}{*}{\textbf{DeepSeek-V3.1}}
& Instruct  & 0.429          & 0.506          & 0.565          & 0.468          & 0.500          & \textbf{0.606} & 0.526          & 0.595          & \textbf{0.785} \\
& Reasoning & \textbf{0.429} & \textbf{0.508} & \textbf{0.567} & \textbf{0.479} & \textbf{0.511} & 0.585          & \textbf{0.794} & \textbf{0.805} & 0.774 \\
& $\Delta$  & \cellcolor{orange!10}\textbf{+0.003} & \cellcolor{orange!10}\textbf{+0.002} & \cellcolor{orange!10}\textbf{+0.002} & \cellcolor{orange!12}\textbf{+0.011} & \cellcolor{orange!12}\textbf{+0.011} & \cellcolor{orange!10}\textbf{-0.021} & \cellcolor{orange!85}\textbf{+0.268} & \cellcolor{orange!67}\textbf{+0.210} & \cellcolor{orange!10}\textbf{-0.011} \\
\cmidrule{1-11}
\multirow{3}{*}{\textbf{Kimi-K2}}
& Instruct  & 0.507          & 0.533          & 0.566          & 0.511          & 0.521          & 0.543          & 0.595          & 0.634          & 0.750 \\
& Reasoning & \textbf{0.580} & \textbf{0.662} & \textbf{0.652} & \textbf{0.575} & \textbf{0.585} & \textbf{0.564} & \textbf{0.834} & \textbf{0.829} & \textbf{0.830} \\
& $\Delta$  & \cellcolor{orange!38}\textbf{+0.073} & \cellcolor{orange!62}\textbf{+0.129} & \cellcolor{orange!42}\textbf{+0.086} & \cellcolor{orange!32}\textbf{+0.064} & \cellcolor{orange!32}\textbf{+0.064} & \cellcolor{orange!12}\textbf{+0.021} & \cellcolor{orange!75}\textbf{+0.239} & \cellcolor{orange!69}\textbf{+0.195} & \cellcolor{orange!40}\textbf{+0.080} \\
\cmidrule{1-11}
\multirow{3}{*}{\textbf{Kimi-K2.5}}
& Instruct  & 0.507          & 0.573          & 0.649          & 0.521          & 0.500          & \textbf{0.553} & 0.699          & 0.734          & 0.875 \\
& Reasoning & \textbf{0.597} & \textbf{0.700} & \textbf{0.682} & \textbf{0.574} & \textbf{0.574} & \textbf{0.585} & \textbf{0.905} & \textbf{0.915} & \textbf{0.911} \\
& $\Delta$  & \cellcolor{orange!50}\textbf{+0.100} & \cellcolor{orange!62}\textbf{+0.127} & \cellcolor{orange!18}\textbf{+0.033} & \cellcolor{orange!28}\textbf{+0.053} & \cellcolor{orange!38}\textbf{+0.074} & \cellcolor{orange!10}\textbf{+0.032} & \cellcolor{orange!72}\textbf{+0.206} & \cellcolor{orange!65}\textbf{+0.181} & \cellcolor{orange!28}\textbf{+0.036} \\
\bottomrule
\end{tabular}}
\end{table*}%
}}

\section{Introduction}
Large Language Models (LLMs) acquire vast parametric knowledge during pretraining, encoding facts and concepts across model parameters. Post-training techniques, including supervised fine-tuning (SFT), distillation, Reinforcement Learning from Human Feedback (RLHF) and RL, then transform these base models into instruction-following agents capable of complex reasoning \citep{bai2022constitutional,wang2025octothinker,yu2025dapo}. While these methods improve reasoning and align with user preferences, a growing body of evidence reveals a trade-off known as the ``alignment tax'' \citep{askell2021general,lin2024mitigating, sorensen2025spectrum}: models sacrifice factual recall to optimize for other objectives, leading to degraded performance on knowledge-intensive benchmarks \citep{gekhman2024does,yuan2024towards}. 

\begin{figure*}[t]
\centering
\includegraphics[width=\textwidth, height=0.15\textheight]{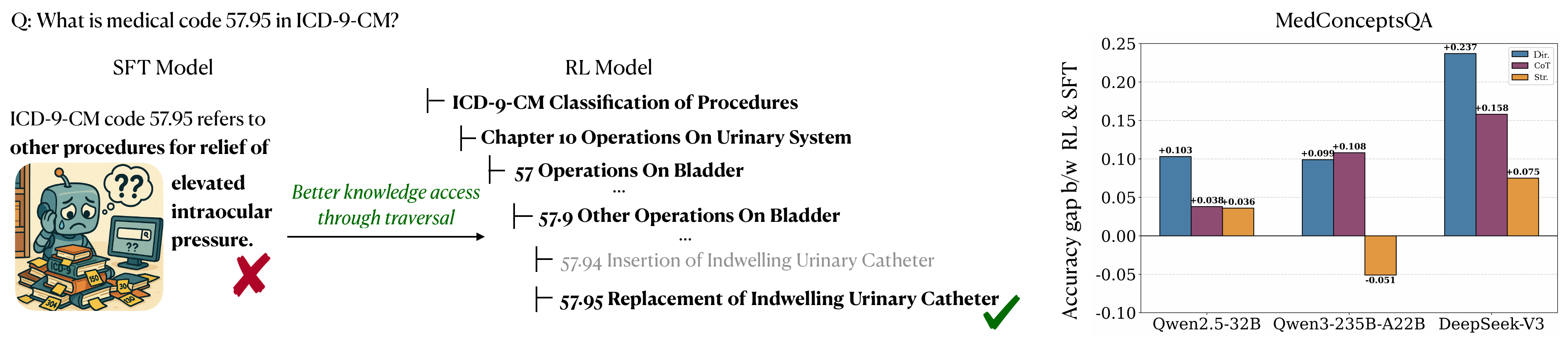}
\caption{%
  \textbf{RL improves LLM's parametric knowledge access through hierarchical traversal.}
  \textbf{(Left)} Example query ``What is medical code 57.95 in ICD-9-CM?''
  The instruct model retrieves an incorrect description via direct memorization,
  while the reasoning model systematically traverses the knowledge hierarchy to
  arrive at the correct answer. Final answers and intermediate codes visited during
  reasoning are in bold.
  \textbf{(Right)} Majority-voted accuracy on MedConceptsQA with three prompt
  strategies: Dir. (direct answering), CoT (chain-of-thought), and Str.\
  (structured hierarchy traversal). When instruct models are explicitly instructed to
  traverse the knowledge hierarchy before answering, their accuracy gap relative to
  reasoning models quickly narrows, sometimes even reverses. This indicates that instruct models
  possess the parametric knowledge but lack effective access methods to surface it.
}
\label{fig:overview}
\end{figure*}

However, existing work has examined a limited number of recall tasks and post-training methods. This leaves open two broader questions: \textbf{\textit{do these degradation patterns generalize across the major forms of knowledge retrieval—entity-relations, taxonomies, and ontologies? And how do knowledge-recall abilities differ across distilled, instruction-tuned, and reasoning models \footnote{Throughout, we use \emph{instruct} and \emph{reasoning} to denote publicly released model variants, whose recipes may combine multiple post-training stages. Comparisons between these variants are therefore observational and do not isolate the effect of RL. We reserve \emph{SFT} and \emph{RL} for the controlled knowledge-injection experiments described in Section~\ref{sec:controlled_injection}.}?}}

To address these questions, we study knowledge retrieval tasks which require models to navigate both explicit and implicit structures in parametric memory. Surprisingly, reasoning models can outperform their instruct counterparts by 13 and 24 percentage points on PopQA and MedConceptsQA respectively \citep{mallen2023trustlanguagemodelsinvestigating,shoham2024medconceptsqaopensourcemedical}, directly challenging the conventional wisdom that RL sacrifices memorization for reasoning \citep{ghosh2024closer,chu2025sft}. We hypothesize these models succeed through systematic hierarchical navigation rather than direct recall. For example, in structured domains, such as medical code lookup, retrieval may require traversing a formal hierarchy (Figure~\ref{fig:overview}): to identify that ICD-9-CM code 57.95 refers to ``Replacement of indwelling urinary catheter,'' reasoning models systematically traverse the knowledge hierarchy (Chapter 11 $\rightarrow$ codes 57.0-57.99 $\rightarrow$ specific procedure). In contrast, instruct models often attempt direct recall, which often fails due to the vast code space. In flat factual domains, such as entity-relation QA, there is no explicit taxonomy, but reasoning models still benefit from implicit navigation by recalling related entities, relations, or contexts that lead to the answer. Therefore, we propose that \textbf{RL improves navigation of existing parametric knowledge rather than adding new factual content}.\footnote{\emph{Knowledge access} (or \emph{retrieval}) denotes
  eliciting facts in model parameters, whereas \emph{hierarchical
  traversal} (or \emph{navigation}) refers to accessing them through intermediate
  levels of knowledge hierarchy; Table~\ref{tab:terminology} provides the
  complete mapping.} 

To disentangle knowledge acquisition from navigation, we design a series of experiments and uncover four findings:

\begin{itemize}[itemsep=2pt, parsep=0pt, topsep=2pt]
    \item \textbf{Instruct models already possess the knowledge but lack the navigation policy to access it.} Inspired by work showing prompt optimization can match RL gains \citep{khattab2023dspy, agrawal2025gepa, ziems2025multi}, we develop structured prompting that explicitly guides models through hierarchical traversal; it reduces the 24pp gap between DeepSeek-V3 and DeepSeek-R1 to 7pp on MedConceptsQA (Figure~\ref{fig:overview}, right). Together with manual inspection of instruct model responses, this result suggests that information is present but only reliably accessible with proper navigation. Response-distribution analysis corroborates this: hierarchical navigation pushes the knowledge distribution from mostly incorrect to mostly correct, while reasoning models---having internalized the navigation policy---change little because they already operate near their performance ceiling. Applying RL to Qwen3-8B-/30B-A3B-Instruct on unseen facts further improves recall on held-out facts that are frequent but previously non-extractable, isolating RL's effect on access to latent parametric knowledge. 

    \item \textbf{The RL gain on knowledge access mostly comes from iterated on-policy exploration. } Holding the data and hyperparameters fixed, we
    compare six post-training objectives spanning supervised finetuning, rejection-sampling,
    and group relative policy optimization (GRPO). Knowledge access improves when self-generated rollouts track the current policy, and the contrastive
    advantage of GRPO adds little once iteration is present.

    \item \textbf{Reasoning models' traversal advantage grows with retrieval depth.} We introduce two depth-stratified datasets, IPC-CAR and DBO-CAR, that require recalling the paths from a given pair of nodes to their nearest common ancestor. Using a path-matching score that measures traversal fidelity, we find the reasoning--instruct gap widens as depth increases---indicating that reasoning's advantage emerges specifically under greater retrieval complexity, a signature of learned traversal that scales with task complexity.

    \item \textbf{Post-training reshapes how models process questions while leaving factual representations intact.} To provide internal validation, we conduct layerwise representational analysis~\citep{agarwal2024policy,mukherjee2025reinforcement,skean2025layer,huan2025does} on matched query--answer pairs, comparing queries (e.g., ``What is medical code 57.95?'') against declarative statements (e.g., ``Code 57.95 refers to urinary catheter replacement''). We find that declarative statements maintain high cosine similarity between base and reasoning models, while queries diverge. This asymmetry shows that post-training techniques such as RL primarily transform how models process questions while leaving factual knowledge representations intact (Figure~\ref{fig:divergence_analysis}).

    \item \textbf{Distillation lacks the exploration behavior needed to navigate latent knowledge.} Contrary to prior findings that distillation can outperform RL on reasoning tasks~\citep{liu2025deepseek, yue2025limitofrlvr}, distilled models often underperform reasoning models on knowledge recall, sometimes falling below base and instruct models. Manual inspection of model outputs shows that they imitate self-correction without truly abandoning incorrect hypotheses, lacking the trial-and-error exploration RL instills. This prevents them from navigating the hierarchy to surface latent knowledge.
\end{itemize}

In summary, our experiments show that reasoning models excel at knowledge-retrieval tasks because they learn hierarchical navigation strategies, rather than simply acquiring new knowledge. Future methods should focus not only on injecting or distilling knowledge, but also on learning how to better retrieve and navigate the knowledge already encoded in model parameters.

\section{Structured Prompting Reveals Latent Knowledge in Instruct Models}
We evaluate whether structured prompting improves access to parametric knowledge across two settings: flat factual associations, where useful intermediate cues are implicit, and formally organized code systems, where the path to an answer is explicit.  These settings allow us to test whether the benefits of navigation generalize across different organizations of knowledge. Across direct, chain-of-thought (CoT), and structured prompts, structured prompting shrinks the instruct–reasoning accuracy gap, showing the effectiveness of hierarchical traversal at re-surfacing the knowledge. Response distribution and reasoning trace analyses corroborate that
instruct models possess the underlying knowledge but lack the navigation policy that
RL internalizes.

\subsection{Experimental Setup}
\noindent \textbf{Models.} To make our conclusions robust, we evaluate eight instruct–reasoning model pairs spanning five families: Qwen2.5-32B \citep{team2024qwen2}, Qwen3 (8B/30B-A3B/235B-A22B) \citep{yang2025qwen3}, DeepSeek-V3/V3.1 \citep{liu2024deepseek, liu2025deepseek}, and Kimi-K2/K2.5 \citep{kimiteam2026kimik25visualagentic, kimiteam2026kimik2openagentic}, covering both dense and mixture-of-experts (MoE) architectures. 

\noindent \textbf{Datasets.} Our evaluation focuses on factual knowledge recall rather than logical reasoning. The benchmarks span relational (PopQA), taxonomic (International Patent Classification (IPC)-Lookup) and ontological (MedConceptsQA) knowledge, with both multiple-choice and open-ended formats. Specifically, PopQA organize facts as entity-relation pairs, while IPC-Lookup and MedConceptsQA encode facts within explicit hierarchies. 
\begin{itemize}[itemsep=0pt, parsep=0pt, topsep=2pt]
    \item \textbf{PopQA} \citep{mallen2023trustlanguagemodelsinvestigating}: an open-ended, entity-centric QA dataset for evaluating long-tail factual recall, without explicit hierarchical structure. 
    \item \textbf{IPC-Lookup:}
    a multiple-choice dataset of 94 questions requiring models to identify the correct description of a given patent code, drawn from the most recent IPC taxonomy, Version 2026.01, released by \citet{wipo2026ipc}. \footnote{We use Version 2026.01 and select codes intended to be stable across recent releases (2024.01--2026.01), minimizing any confound from models trained before the latest revision.} Each question presents a patent code alongside four candidate descriptions. Further details are in Appendix~\ref{app:ipc_lookup}. 
    \item \textbf{MedConceptsQA} \citep{shoham2024medconceptsqaopensourcemedical}: a multiple-choice benchmark testing recall of code descriptions for clinical ontologies including diagnoses (ICD9/10-CM), procedures (ICD9/10-PROC), and drugs (ATC).
\end{itemize}

\noindent \textbf{Prompting.} We use three prompt templates:
\begin{itemize}[itemsep=0pt, parsep=0pt, topsep=2pt]
     \item \textbf{Direct QA} asks the model to output only the final answer, without explanation.
     \item \textbf{CoT} asks the model to provide a final answer with a free-form explanation, without imposing any procedural constraints.
     \item \textbf{Structured Prompting} asks the model to traverse the relevant hierarchy of facts or codes before giving the final answer. This tests whether explicit structured recall can reduce the performance gap between instruct and reasoning models (see Appendix~\ref{app: prompts} for detail).
\end{itemize}

\noindent \textbf{Evaluation.} We sample from all models using a temperature of 0.8, a top-p of 0.7, and a maximum of 10,000 tokens across three independent runs. Performance is reported as both majority-voted and mean accuracy (± standard deviation), where majority voting selects the most frequent answer per question. Here we prioritize majority-voted accuracy over pass@k because our goal is to measure reliable knowledge access rather than knowledge boundary. We describe the full protocol in Appendix~\ref{app:evaluation}.

\subsection{Structured Prompting Narrows the Gap Between Instruct and Reasoning Models on Knowledge Retrieval} 
\label{sec:prompting_strategies}
Structured prompting explicitly instructs models to traverse unstructured factual data through relevant entities and structured codes through neighbors and predecessors. This substantially reduces the accuracy gap between instruct and reasoning models on knowledge retrieval tasks (Table~\ref{tab:majority_vote_compact}). On MedConceptsQA, the DeepSeek-V3--DeepSeek-R1 gap shrinks from +23.7 pp under direct QA to +7.5 pp with structured prompting. PopQA shows similar pattern: the gap decreases from +12.7 pp to +3.3 pp for Kimi-K2.5, and from +13.4 pp to +6.5 pp for Qwen2.5-32B. These results show that hierarchical navigation helps instruct models better access the latent parametric knowledge. We further analyze the implications of this finding in Section~\ref{sec: sft_access_knowledge}.

\subsection{Instruct Models Already Have Parametric Knowledge But Needs Better Access}
\label{sec: sft_access_knowledge}
\begin{figure}[t]
\centering
\includegraphics[width=\columnwidth, keepaspectratio]{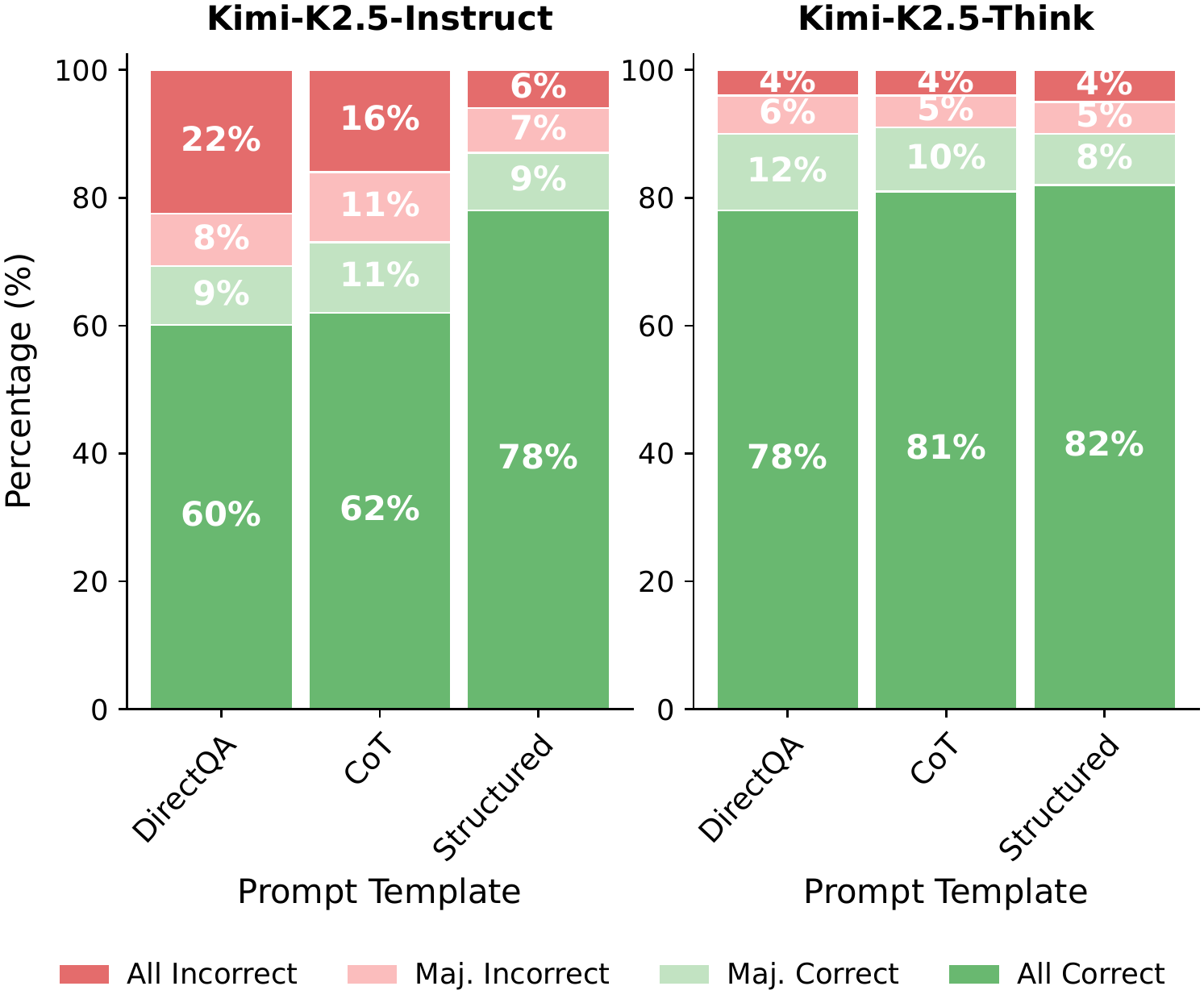}
\caption{Comparative distribution analysis of Kimi-K2.5-Instruct and Kimi-K2.5-Think on MedConceptsQA.}
\label{fig:majvote_stackedbar}
\end{figure}

\textbf{Response Distribution Analysis.} Section~\ref{sec:prompting_strategies} shows that providing instruct models with traversal algorithms can narrow the performance gap between instruct and reasoning models. To further investigate this, we examine changes in correctness distribution. Figure~\ref{fig:majvote_stackedbar} reports majority-voting results for Kimi-K2.5-Instruct and Kimi-K2.5-Think over three independent runs on MedConceptsQA, categorized into four classes: \textit{All Incorrect} (0/3), \textit{Majority Incorrect} (1/3), \textit{Majority Correct} (2/3), and \textit{All Correct} (3/3). Switching from direct QA to structured prompting shows marked redistribution in Kimi-K2.5-Instruct where \textit{All Incorrect} drops from 29 \% to 6 \% and \textit{All Correct} increases from 60\% to 78\%. Kimi-K2.5-Think shows no significant shift (e.g.,  \textit{All Incorrect} stays at 4\%), suggesting it already operates near its performance ceiling. This suggests that the accuracy gains come not from acquiring new knowledge, but from effectively surfacing the parametric knowledge already encoded in instruct models.

\noindent \textbf{Reasoning Trajectories Analysis.} We present the responses from instruct and reasoning models in Figures~\ref{fig:reasoning_traces_kimi_popqa}--\ref{fig:reasoning_traces}. The instruct model only retrieves the correct answer under structured prompting by explicitly recalling the code or concept hierarchy, whereas the reasoning model correctly answers the question via spontaneous traversal. This shows that reasoning model internalizes the navigation policy required for accessing parametric knowledge, while structured prompting provides this guidance externally for instruct models. 

\section{Isolating RL’s Effect on Unlocking Inaccessible Knowledge Through Controlled Injection}
\label{sec:controlled_injection}

To isolate the effect of RL on parametric knowledge access from data exposure, we construct a controlled knowledge-injection setting using PopQA. We perform RL exclusively on facts that models have never seen during pre-training and cannot extract through direct retrieval, then evaluate whether this improves access to a disjoint set of facts that are frequently seen in pre-training but similarly inaccessible. If RL improves accuracy on these held-out high-exposure facts, it suggests that RL improves the model’s ability to access latent parametric knowledge.

\subsection{Experimental Setup}
\noindent \textbf{Controlled Knowledge Injection.} To estimate how frequently each entity–answer pair co-occurs in pretraining corpus, we use Infini-gram, an n-gram language model over large text corpora \citep{liu2025infinigramscalingunboundedngram}. The zero-co-occurrence examples that a model cannot answer correctly across three independent runs are then used as a proxy for unseen facts. Here we focus on Qwen3-8B-/30B-A3B-Instruct, evaluate them on the low-frequency examples and retain the subset that the models answer incorrectly as the RL training set. Similarly, examples with high co-occurrence counts that remain non-extractable are also retained as the held-out test set. The two subsets are subsequently filtered to use disjoint relations, eliminating relation-overlap confounds. We then apply RL to the rare, non-extractable subset until convergence and evaluate the trained models on this test set. To further isolate the RL contribution, we also train the models with SFT on the same data similarly.  Additional experimental details are reported in Appendix~\ref{sec:rl_exp}.

\subsection{RL Unlocks Latent Knowledge Better Than SFT}
RL on the rare, non-extractable data improves access to the high-frequency, inaccessible held-out set from 0\% to 10.1\% on Qwen3-8B-Instruct and from $0\%$ to $7.05\%$ on Qwen3-30B-A3B-Instruct, while SFT on the same data yields only 0.96\% and $0.64\%$. Because the training and test splits are relation-disjoint and the test facts were never seen during fine-tuning, this gain cannot be explained by data exposure or relation overlap. This suggests that RL improves access to latent parametric knowledge already encoded during pretraining.

\subsection{On-Policy Exploration Drives the Gain}

\begin{figure}[t]
    \centering
    \includegraphics[width=\linewidth]
    {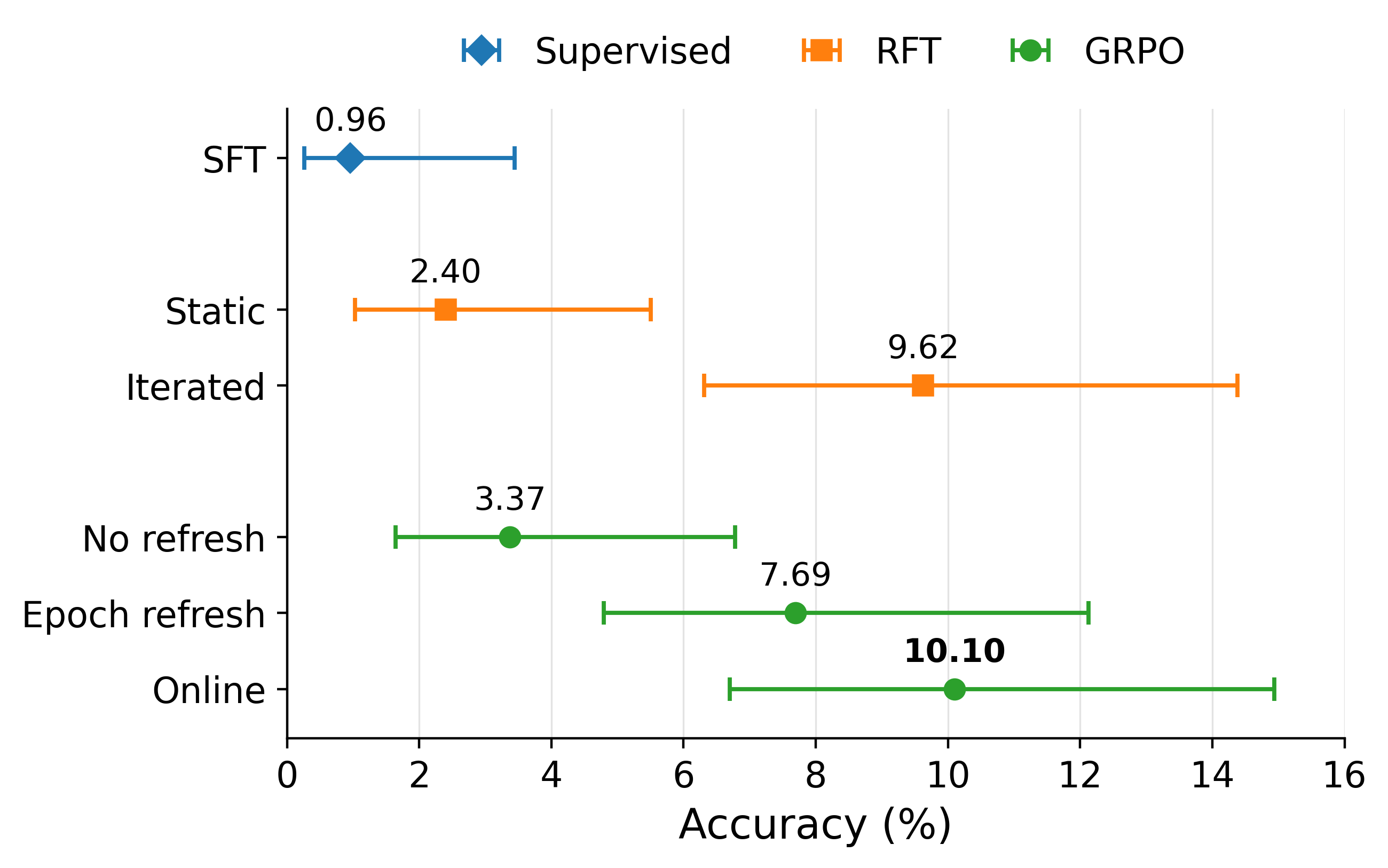}
    \caption{
    Decomposing RL on the controlled knowledge-injection task using
    Qwen3-8B-Instruct. Points indicate accuracy, and horizontal error bars
    indicate 95\% Wilson confidence intervals.
    }
    \label{fig:rl_decomposition}
\end{figure}


\label{sec:rl_decomposition}
The comparison above separates RL from SFT but does not identify which ingredient
of RL matters. We therefore decompose the training objective along three axes---%
whether the training data is self-generated, how often it is
refreshed, and whether incorrect rollouts are discarded or instead supply a negative
gradient---holding the data and hyperparameters fixed on
Qwen3-8B-Instruct so that only the objective changes
(Figure~\ref{fig:rl_decomposition}). 

Rejection-sampling fine-tuning (RFT) samples multiple rollouts for each question, retains only those that produce a correct answer, and trains on the retained rollouts with the standard supervised likelihood objective. Static RFT generates this training set once from the initial policy and keeps it fixed throughout training, whereas Iterated RFT regenerates rollouts from the current policy per batch.  

Group Relative Policy Optimization (GRPO) \citep{shao2024deepseekmathpushinglimitsmathematical} is an RL algorithm that scores a
group of rollouts per question and updates the policy toward those scoring above
the group mean and away from those below it.  Within GRPO, we change how frequently the self-generated rollouts are refreshed while leaving the rest unchanged. Offline GRPO reuses rollouts sampled
once from the initial policy; epoch-refresh GRPO regenerates them from the latest policy once per training epoch; and Online GRPO generates fresh rollouts from the current policy before every update. These variants therefore interpolate from fully stale to fully current training.

  
Three patterns emerge. First, switching from externally supplied answers to the
model's own correct rollouts is necessary but not sufficient: replacing the externally supplied target answer with the model's own correct samples raises recall only from $0.96\%$ to $2.40\%$. Second, iteration on self-generated data accounts for
most of the effect: a single round of rejection-sampling fine-tuning barely moves
recall (Static RFT, $2.40\%$), whereas repeating the sample--filter--finetune
loop recovers most of it (Iterated RFT, $9.62\%$). The same ordering holds within
GRPO, where recall rises monotonically as rollouts become fresher
($3.37\% \rightarrow 7.69\% \rightarrow 10.1\%$). Third, GRPO's contrastive
advantage---which pushes above-average rollouts per question up while pushing below-average ones down---contributes little
once iteration is present: Online GRPO ($10.1\%$) and Iterated RFT ($9.62\%$) are
statistically indistinguishable in held-out accuracy. Table~\ref{tab:rl_decomposition_pairwise} reports pairwise confidence intervals supporting these comparisons. Improved knowledge access thus comes from iterated on-policy exploration, which is precisely what distilled models lack (Section~\ref{sec:distill_analysis}).

\section{Reasoning Improves Hierarchical Navigation Under Greater Retrieval Complexity}
\label{sec:retrieval_complexity}
\vspace{-0.3em}

\begin{figure}[H]
    \centering
    \hspace*{0.03\columnwidth}
    \includegraphics[width=0.9\columnwidth]{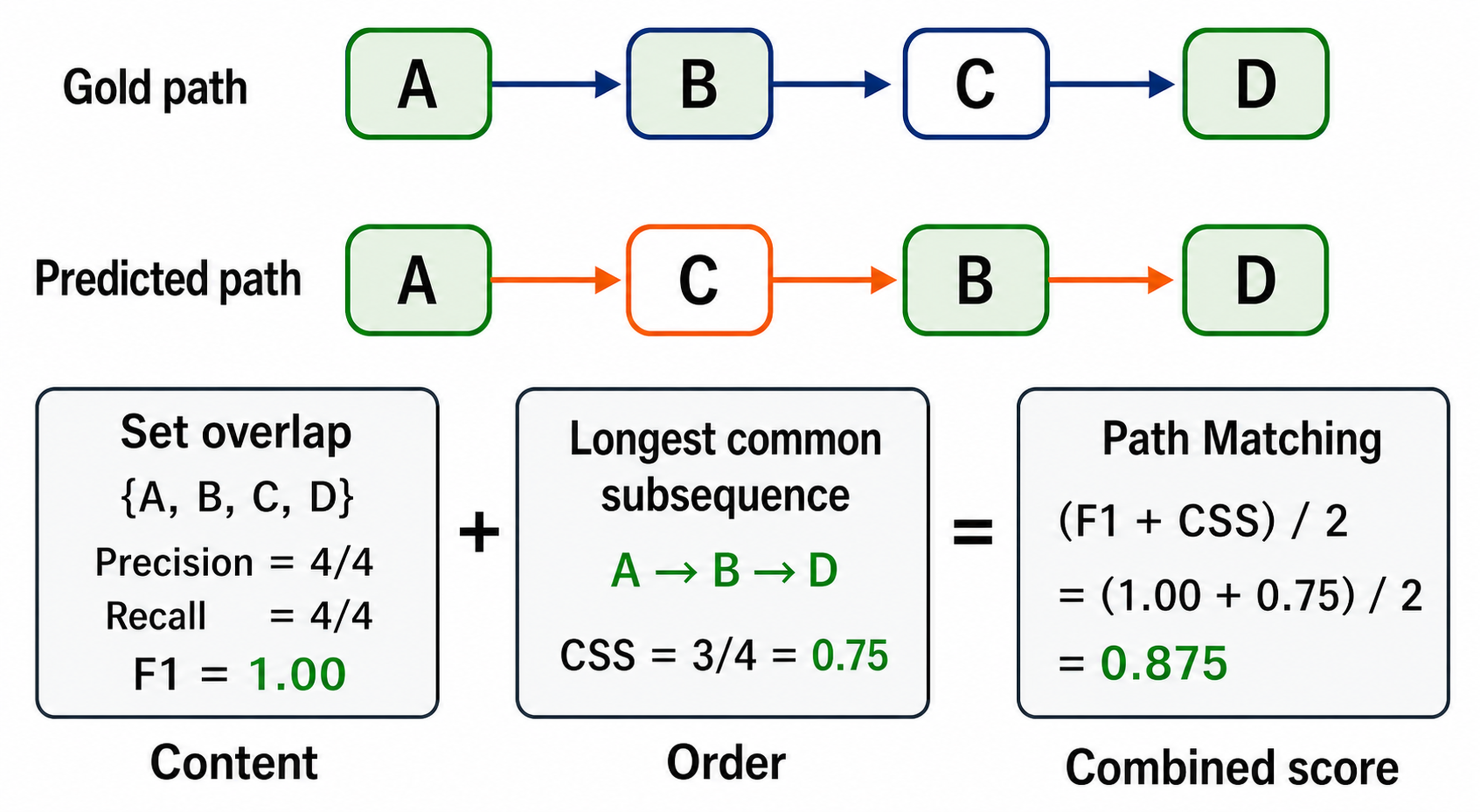}
    \caption{Illustration of the Path Matching score.}
    \label{fig:path_matching_illustration}
\end{figure}

\noindent Section~\ref{sec:prompting_strategies} showed that structured prompting recovers most of the instruct--reasoning gap on knowledge retrieval, but the benchmarks used there do not systematically vary the depth of traversal required to reach an answer. To probe this, we introduce IPC-CAR and DBO-CAR---two datasets stratified by retrieval depth---and a path matching metric to score traversal quality. Results show that reasoning models exhibit superior hierarchical traversal as depth increases.

\begin{figure*}[t]
\centering
\begin{tikzpicture}
\begin{axis}[
    name=pmRaw,
    width=8cm, height=6cm,
    ybar, bar width=10pt,
    ymin=0, ymax=1.15,
    ytick={0,0.2,0.4,0.6,0.8,1.0},
    enlarge x limits=0.18,
    symbolic x coords={2,3,4,5,6},
    xtick=data,
    xlabel={\texttt{Depths}},
    ylabel={Path Matching},
    title={\textbf{Path Matching} (raw) $\uparrow$},
    title style={font=\small},
    legend style={
        at={(0.96,0.96)}, anchor=north east,
        font=\footnotesize, draw=none, fill=none,
        /tikz/every even column/.append style={column sep=6pt}
    },
    legend columns=2,
    ymajorgrids, grid style={gray!20},
    tick label style={font=\footnotesize},
    label style={font=\footnotesize},
]
\addplot[fill=v3blue, draw=v3blue!80!black] coordinates {
    (2,0.865) (3,0.576) (4,0.454) (5,0.416) (6,0.327)
};
\addplot[fill=r1coral, draw=r1coral!80!black] coordinates {
    (2,0.725) (3,0.549) (4,0.473) (5,0.465) (6,0.434)
};
\legend{ds\_v3, ds\_r1}
\node[font=\fontsize{6}{7}\selectfont, xshift=-4pt, yshift=1pt, anchor=south] at (axis cs:2,0.865) {0.865};
\node[font=\fontsize{6}{7}\selectfont, xshift=-6pt, yshift=1.8pt, anchor=south] at (axis cs:3,0.576) {0.576};
\node[font=\fontsize{6}{7}\selectfont, xshift=-8pt, yshift=0.2pt, anchor=south] at (axis cs:4,0.454) {0.454};
\node[font=\fontsize{6}{7}\selectfont, xshift=-6pt, yshift=0.1pt, anchor=south] at (axis cs:5,0.416) {0.416};
\node[font=\fontsize{6}{7}\selectfont, xshift=-6pt, yshift=0.5pt, anchor=south] at (axis cs:6,0.327) {0.327};

\node[font=\fontsize{6}{7}\selectfont, xshift=6pt, yshift=0.3pt, anchor=south] at (axis cs:2,0.725) {0.725};
\node[font=\fontsize{6}{7}\selectfont, xshift=6pt, yshift=0.1pt, anchor=south] at (axis cs:3,0.549) {0.549};
\node[font=\fontsize{6}{7}\selectfont, xshift=4pt, yshift=4pt, anchor=south] at (axis cs:4,0.473) {0.473};
\node[font=\fontsize{6}{7}\selectfont, xshift=4pt, yshift=2pt, anchor=south] at (axis cs:5,0.465) {0.465};
\node[font=\fontsize{6}{7}\selectfont, xshift=4pt, yshift=2pt, anchor=south] at (axis cs:6,0.434) {0.434};
\end{axis}

\begin{axis}[
    name=pmGap,
    at={($(pmRaw.east)+(1.7cm,0)$)}, anchor=west,
    width=8cm, height=6cm,
    ybar, bar width=14pt,
    ymin=-0.18, ymax=0.14,
    ytick={-0.16,-0.12,-0.08,-0.04,0,0.04,0.08,0.12},
    yticklabel style={/pgf/number format/.cd, fixed, precision=2, zerofill},
    enlarge x limits=0.18,
    symbolic x coords={2,3,4,5,6},
    xtick={2,3,4,5,6},
    xlabel={\texttt{Depths}},
    ylabel={Gap (R1 $-$ V3)},
    title={\textbf{Path Matching} gap},
    title style={font=\small},
    ymajorgrids, grid style={gray!20},
    extra y ticks={0}, extra y tick labels={},
    extra y tick style={grid=major, grid style={black, thick}},
    tick label style={font=\footnotesize},
    label style={font=\footnotesize},
]
\addplot[ybar, bar width=14pt, draw=black!40, fill=gapred] coordinates {
    (2,-0.140) (3,-0.027) (4,nan) (5,nan) (6,nan)
};
\addplot[ybar, bar width=14pt, draw=black!40, fill=gapgreen] coordinates {
    (2,nan) (3,nan) (4,0.019) (5,0.049) (6,0.107)
};
\node[font=\tiny] at (axis cs:2,-0.155) {$-$0.140};
\node[font=\tiny] at (axis cs:3,-0.043) {$-$0.027};
\node[font=\tiny] at (axis cs:4, 0.033) {+0.019};
\node[font=\tiny] at (axis cs:5, 0.064) {+0.049};
\node[font=\tiny] at (axis cs:6, 0.123) {+0.107};
\node[font=\tiny, gray] at (axis cs:2,-0.170) {n=20};
\node[font=\tiny, gray] at (axis cs:3,-0.170) {n=20};
\node[font=\tiny, gray] at (axis cs:4,-0.170) {n=20};
\node[font=\tiny, gray] at (axis cs:5,-0.170) {n=20};
\node[font=\tiny, gray] at (axis cs:6,-0.170) {n=20};
\end{axis}

\begin{axis}[
    name=accRaw,
    at={($(pmRaw.south)+(0,-1.9cm)$)}, anchor=north,
    width=8cm, height=6cm,
    ybar, bar width=10pt,
    ymin=0, ymax=1.15,
    ytick={0,0.2,0.4,0.6,0.8,1.0},
    enlarge x limits=0.18,
    symbolic x coords={2,3,4,5,6},
    xtick=data,
    xlabel={\texttt{Depths}},
    ylabel={Accuracy},
    title={\textbf{Accuracy} (raw) $\uparrow$},
    title style={font=\small},
    ymajorgrids, grid style={gray!20},
    tick label style={font=\footnotesize},
    label style={font=\footnotesize},
]
\addplot[fill=v3blue, draw=v3blue!80!black] coordinates {
    (2,0.900) (3,0.850) (4,0.700) (5,0.750) (6,0.750)
};
\addplot[fill=r1coral, draw=r1coral!80!black] coordinates {
    (2,0.800) (3,0.700) (4,0.600) (5,0.800) (6,0.800)
};
\node[font=\fontsize{6}{7}\selectfont, xshift=-4pt, yshift=2pt, anchor=south] at (axis cs:2,0.900) {0.900};
\node[font=\fontsize{6}{7}\selectfont, xshift=-4pt, yshift=2pt, anchor=south] at (axis cs:3,0.850) {0.850};
\node[font=\fontsize{6}{7}\selectfont, xshift=-4pt, yshift=2pt, anchor=south] at (axis cs:4,0.700) {0.700};
\node[font=\fontsize{6}{7}\selectfont, xshift=-6pt, yshift=0pt, anchor=south] at (axis cs:5,0.750) {0.750};
\node[font=\fontsize{6}{7}\selectfont, xshift=-6pt, yshift=0pt, anchor=south] at (axis cs:6,0.750) {0.750};

\node[font=\fontsize{6}{7}\selectfont, xshift=6pt, yshift=0.2pt, anchor=south] at (axis cs:2,0.800) {0.800};
\node[font=\fontsize{6}{7}\selectfont, xshift=6pt, yshift=0.2pt, anchor=south] at (axis cs:3,0.700) {0.700};
\node[font=\fontsize{6}{7}\selectfont, xshift=6pt, yshift=0.2pt, anchor=south] at (axis cs:4,0.600) {0.600};
\node[font=\fontsize{6}{7}\selectfont, xshift=4pt, yshift=0.2pt, anchor=south] at (axis cs:5,0.800) {0.800};
\node[font=\fontsize{6}{7}\selectfont, xshift=4pt, yshift=0.2pt, anchor=south] at (axis cs:6,0.800) {0.800};
\end{axis}

\begin{axis}[
    name=accGap,
    at={($(accRaw.east)+(1.7cm,0)$)}, anchor=west,
    width=8cm, height=6cm,
    ybar, bar width=14pt,
    ymin=-0.22, ymax=0.12,
    ytick={-0.20,-0.15,-0.10,-0.05,0,0.05,0.10},
    yticklabel style={/pgf/number format/.cd, fixed, precision=2, zerofill},
    enlarge x limits=0.18,
    symbolic x coords={2,3,4,5,6},
    xtick={2,3,4,5,6},
    xlabel={\texttt{Depths}},
    ylabel={Gap (R1 $-$ V3)},
    title={\textbf{Accuracy} gap},
    title style={font=\small},
    ymajorgrids, grid style={gray!20},
    extra y ticks={0}, extra y tick labels={},
    extra y tick style={grid=major, grid style={black, thick}},
    tick label style={font=\footnotesize},
    label style={font=\footnotesize},
]
\addplot[ybar, bar width=14pt, draw=black!40, fill=gapred] coordinates {
    (2,-0.100) (3,-0.150) (4,-0.100)
};
\addplot[ybar, bar width=14pt, draw=black!40, fill=gapgreen] coordinates {
    (5,0.050) (6,0.050)
};
\node[font=\tiny] at (axis cs:2,-0.115) {$-$0.100};
\node[font=\tiny] at (axis cs:3,-0.165) {$-$0.150};
\node[font=\tiny] at (axis cs:4,-0.115) {$-$0.100};
\node[font=\tiny] at (axis cs:5, 0.065) {+0.050};
\node[font=\tiny] at (axis cs:6, 0.065) {+0.050};
\node[font=\tiny, gray] at (axis cs:2,-0.205) {n=20};
\node[font=\tiny, gray] at (axis cs:3,-0.205) {n=20};
\node[font=\tiny, gray] at (axis cs:4,-0.205) {n=20};
\node[font=\tiny, gray] at (axis cs:5,-0.205) {n=20};
\node[font=\tiny, gray] at (axis cs:6,-0.205) {n=20};
\end{axis}

\end{tikzpicture}

\caption{%
  Path matching score and final-answer accuracy of DeepSeek-V3 (Instruct) versus DeepSeek-R1 (Reasoning) on IPC-CAR, stratified by depth.
  \textbf{Left column:} raw scores; arrows ($\uparrow$) indicate higher is better.
  \textbf{Right column:} gap (R1 $-$ V3) plotted on a tight y-axis centered on zero, with positive gaps (R1 better) in green and negative gaps (R1 worse) in red.
  Each level has $n{=}20$ questions.
}
\label{fig:ipc_car_v3_r1}
\end{figure*}

\subsection{Experimental Setup}

\noindent \textbf{Datasets.} 
\begin{itemize}[itemsep=0pt, parsep=0pt, topsep=2pt]
    \item \textbf{IPC Common Ancestor Retrieval (IPC-CAR)}: a dataset used to evaluate model's fine-grained recall ability under controlled retrieval complexity. Each question presents two leaf IPC codes and asks the model to identify their nearest common ancestor (NCA). We define retrieval depth as the total number of edges that must be recalled along the two upward paths. Since each leaf is at least one edge away from the NCA, the minimum depth is two. 
    \item \textbf{DBpedia Ontology Common Ancestor Retrieval (DBO-CAR)}: Derived from Wikipedia infoboxes, the community-curated DBpedia Ontology covers 768 classes across domains like people, organizations, and species \citep{dbpedia_ontology}. We use it to construct DBO-CAR, an ontology-grounded counterpart to IPC-CAR that tests recall over CamelCase concept names rather than alphanumeric patent codes. Full construction details for IPC/DBO-CAR are in Section~\ref{app:ipc_car} and~\ref{app:dbo_car}.
\end{itemize}

\noindent \textbf{Path Matching Score.}
To evaluate the quality of predicted hierarchical paths, we propose the path matching score, which combines two metrics:

\begin{itemize}[itemsep=0pt, parsep=0pt, topsep=2pt]
    \item \textbf{F1 Score:} The harmonic mean of precision $P$ and recall $R$ over the set of predicted vs.\ ground-truth ancestors, $F_1 = \frac{2 \times P \times R}{P + R}$ \citep{buckland1994relationship}.
    \item \textbf{Common Subsequence Score (CSS):} Evaluates path order fidelity as the length of the Longest Common Subsequence (LCS) \citep{paterson1994longest} between the predicted and ground-truth paths, normalized by ground-truth length: $\text{CSS} = \frac{|\text{LCS}(\text{predicted}, \text{ground truth})|}{|\text{ground truth ancestors}|}$.
\end{itemize}

\noindent The Path Matching score is the mean of the two components,
$\text{PathMatch} = \frac{F_1 + \text{CSS}}{2}$, capturing both set-level accuracy
and order fidelity in hierarchical recall, as illustrated in
Figure~\ref{fig:path_matching_illustration}.


\subsection{Reasoning Models Show Superior Path Recall When Traversal Depth Increases} 
\noindent We evaluate DeepSeek-V3 (Instruct) and DeepSeek-R1 (Reasoning) on
IPC-CAR, stratified by retrieval depth (Figure~\ref{fig:ipc_car_v3_r1}).

\begin{figure*}[t]
\centering
\hspace*{-0.3\textwidth}
\includegraphics[width=1.3\textwidth]{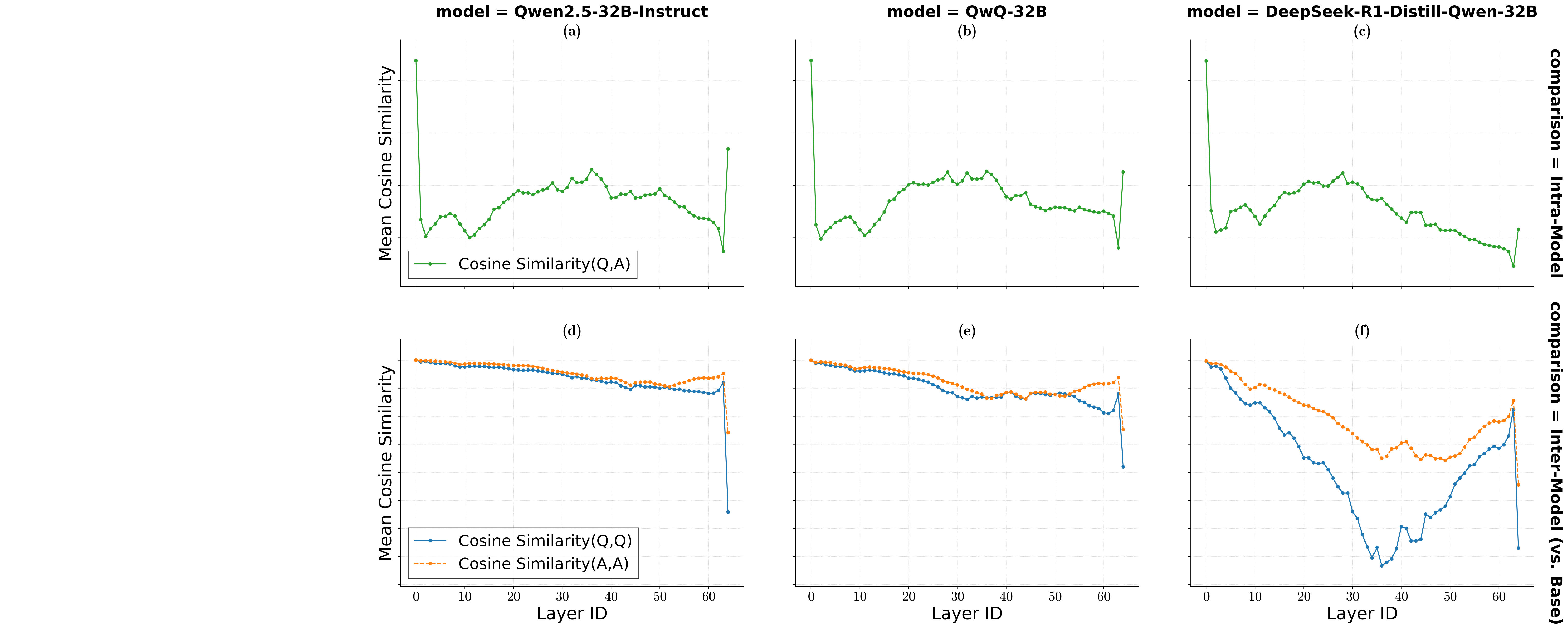}
\caption{Layerwise cosine similarity of question (Q) and answer (A) representations on ICD9PROC. \textbf{Top row (a–c)}: intra-model Q–A similarity within the instruct, reasoning, and distilled variants of Qwen2.5. \textbf{Bottom row (d–f)}: inter-model Q–Q and A–A similarity between each variant and the base model.}
\label{fig:divergence_analysis}
\end{figure*}

A clear crossover emerges: V3 outperforms R1 at shallow depths (PathMatch
$0.865$ vs.\ $0.725$ at depth 2), but R1 overtakes V3 from depth 4 onward,
with the gap widening to $+0.107$ at depth 6. While final-answer accuracy
shows the same directional crossover, only PathMatch captures the depth-dependent widening. R1's advantage under greater retrieval complexity shows procedural navigation beyond simple memorization. DBO-CAR shows similar pattern (Appendix~\ref{app:dbo_car_results}).

\section{Mechanistic and Failure-Mode Analysis}
\subsection{Layerwise Representation Analysis}
\label{sec:representation_analysis}
To localize how post-training reshapes internal knowledge processing, we analyze layerwise activations on MedConceptsQA using contrastive question--answer probes,
comparing the base model against its instruct, reasoning, and distilled variants.

\subsubsection{Experimental Setup}
\noindent \textbf{Probe Construction.}
We sample 100 question--answer pairs from the \texttt{ICD9PROC} vocabulary
in MedConceptsQA. Each pair consists of a factual question and its
ground-truth description, formatted as a declarative statement. For
example, code \texttt{87.15} yields:
\vspace{-0.2em}

\begin{quote}
\textit{Question:} What is the description of the medical code
\texttt{87.15} in \texttt{ICD9PROC}?

\textit{Answer:} The description of the medical code \texttt{87.15} in
\texttt{ICD9PROC} is contrast radiogram of sinus.
\end{quote}
\vspace{-0.2em}
\noindent For a model with $L$ layers and hidden dimension $d$, we process each probe's question and answer independently. At every layer $\ell$, we extract the final-token hidden state $\mathbf{h}_\ell \in \mathbb{R}^d$, which summarizes the full preceding context at that layer.

\noindent \textbf{Analyses.} We compute cosine similarity between
representations under two regimes:
\begin{itemize}[itemsep=2pt, topsep=0pt, parsep=0pt]
    \item \textbf{Inter-model}: For each
    layer, we compare $\mathbf{Q}^{\text{base}}$ to
    $\mathbf{Q}^{\text{post-trained}}$ and $\mathbf{A}^{\text{base}}$ to
    $\mathbf{A}^{\text{post-trained}}$, measuring how much post-training
    moves question vs.\ answer representations.
    \item \textbf{Intra-model}: For each
    model in isolation, we compare $\mathbf{Q}$ to $\mathbf{A}$
    layer-by-layer, tracing how internal activations evolve from encoding
    a question to producing an answer.
\end{itemize}

\noindent Similarity at layer $\ell$, cosine similarity for source pair $(a, b)$ is
$\cos^{(a,b)}(\ell) = \frac{1}{N}\sum_{i=1}^{N}
\frac{\mathbf{h}_{\ell}^{(a)}(i)^{\top}\mathbf{h}_{\ell}^{(b)}(i)}
{\|\mathbf{h}_{\ell}^{(a)}(i)\|_2\,\|\mathbf{h}_{\ell}^{(b)}(i)\|_2}$,
averaged over $N=100$ probes (formal notation in Appendix~\ref{app:similarity_metric}).

\noindent \textbf{Models.} We compare Qwen2.5-32B (base) against its instruct, reasoning, and distilled variants. Mistral-Small-24B results are in Appendix~\ref{app:cos_sim_app}.

\subsubsection{Instruct and Reasoning Models Reshape Question Processing, Not Stored Knowledge}
Figure~\ref{fig:divergence_analysis} shows layerwise similarity trends; Tables~\ref{tab:layerwise_similarity} and~\ref{tab:within_model_qa_similarity}
report inter- and intra-model values aggregated into early (1--21), middle (22--43), and late (44--64) bands. We highlight two findings:
\begin{itemize}[itemsep=2pt, topsep=2pt, parsep=0pt]
    \item Within each model, question and answer representations start highly similar but diverge in later layers (e.g., the reasoning variant's Q--A similarity drops from $0.878$ at layer 0 to $0.161$ at layer 63; Figure~\ref{fig:divergence_analysis}), indicating that representations accumulate increasingly distinct features.
    \item Instruct and reasoning models stay closely aligned with the base (e.g., mean Q/A similarity: 0.966/0.976 for instruct; Table \ref{tab:layerwise_similarity}), whereas the distilled model diverges substantially (mean Q/A: 0.808/0.892), mirroring its degraded knowledge retrieval (Section~\ref{sec:distill_analysis}). Notably, question representations diverge more than answers across all post-trained models, suggesting that performance gains arise primarily from refining question understanding rather than reorganizing factual knowledge.
\end{itemize}

 
\subsection{Distillation Captures Surface Patterns Without Navigation}
\label{sec:distill_analysis}
\begin{table}[h]
\centering
\caption{Qwen2.5-32B model series majority-vote accuracy on MedConceptsQA.}
\label{tab:qwen32b_medconceptsqa_mv}
\resizebox{0.70\columnwidth}{!}{%
\begin{tabular}{lccc}
\toprule
\textbf{Model} & \textbf{QA} & \textbf{CoT} & \textbf{Structured} \\
\midrule
Base      & 0.221 & 0.332 & 0.404 \\
Distilled & 0.375 & 0.380 & 0.447 \\
Instruct  & 0.379 & 0.475 & 0.469 \\
Reasoning & \textbf{0.482} & \textbf{0.513} & \textbf{0.505} \\
\bottomrule
\end{tabular}}
\end{table}

\noindent In addition to SFT and RL, distillation — training on CoT traces from a capable teacher model — is another widely used approach to improving model performance \citep{liu2025deepseek, yue2025limitofrlvr}. To analyze the distillation effectivess, we examine the Qwen2.5-32B model series, comparing four variants: base, instruct, reasoning, and distilled, which is trained on responses from DeepSeek-R1. As shown in Table~\ref{tab:qwen32b_medconceptsqa_mv}, on MedConceptsQA, the distilled model outperforms its base yet remains below both the instruct and reasoning versions, even under structured prompting. Additional results are reported in Table~\ref{tab:qwen32b_both_datasets}.

To investigate this, we analyze 40 responses from DeepSeek-R1-Distill-Qwen2.5-32B that lead to wrong answers. We find that models fail to perform proper hierarchical traversal: whereas DeepSeek-R1 reconstructs multi-level paths (e.g., ICD9PROC $\to$ 81 $\to$ 81.3 $\to$ 81.36), the distilled model frequently stops at shallow prefixes or bypasses code traversal entirely. Examining these traces, we identify two failure modes. The most prevalent (80\%) happens when distilled models appear to simulate self-correction linguistically using keywords such as ``wait'', yet never truly uncommit from their initial hypothesis. Here the model forms a strong initial prior and filters alternatives through shallow pattern matching against its initial choice. For the remaining 20\%, model backtracks across multiple answers before selecting the wrong one. This suggests that, unlike reasoning models, which learn a navigation policy through trial-and-error exploration, SFT-based distillation captures only surface-level self-correction patterns without learning to uncommit from mistakes. As a result, distilled model lacks the navigation skills to unlock the parametric knowledge. Additional reasoning trajectories for each failure mode are provided in Table~\ref{tab:distill_errors}.

\section{Conclusion}
We showed that RL improves how LLMs \emph{access} knowledge rather than \emph{what} they store. Structured prompting recovers most of the instruct--reasoning gap on knowledge retrieval, showing that instruct models already possess the underlying knowledge but lack the navigation policy that RL internalizes. It therefore provides a low-cost, training-free alternative for recovering much of
  the reasoning advantage by explicitly guiding instruct models to
  traverse knowledge already stored in their parameters. A controlled knowledge-injection experiment confirms that RL on unseen data can re-surface high-exposure facts that are previously inaccessible. Across RFT and GRPO variants, performance improves as
  self-generated rollouts are refreshed more frequently, while Online GRPO and
  Iterated RFT perform similarly. This points to iterated on-policy exploration as the main driver of
  the gain. On depth-stratified IPC-CAR and DBO-CAR benchmarks, the instruct--reasoning gap widens with retrieval depth, indicating that reasoning models internalize a scalable traversal capacity. Layerwise analysis shows that post-training primarily reshapes query representations while leaving factual representations stable, and failure-mode analysis  explains why distilled models, which imitate self-correction without acquiring exploratory behavior, fail to inherit this navigation ability. Together, these findings reframe the instruct--reasoning gap on knowledge-intensive tasks as an access bottleneck rather than a knowledge gap, and suggest that future studies should optimize retrieval procedures alongside knowledge content.

\section*{Limitations}
First, we use co-occurrence counts as a proxy for whether data was encountered during pretraining, but it is not a definitive measure of training exposure. Second, while our controlled knowledge-injection experiment attempts to isolate the effect of RL on knowledge access, it cannot fully account for all the differences of a given instruct/reasoning model pair due to their closed training recipes. 

\section*{Ethical considerations}
In this paper, we only use open source datasets or databases and do not collect any sensitive data. This work studies how post-training affects access to parametric knowledge. The main potential risk is that improving parametric knowledge access may also make models more likely to retrieve incorrect or outdated facts. Our evaluations are carried out on verifiable benchmarks, and improved benchmark performance should not be interpreted as reliable factuality in real-world deployment.

\section*{Acknowledgments}
We gratefully acknowledge the Foresight Institute for supporting this work.

\bibliography{custom}

@misc{shao2024deepseekmathpushinglimitsmathematical,
 author = {Zhihong Shao and Peiyi Wang and Qihao Zhu and Runxin Xu and Junxiao Song and Xiao Bi and Haowei Zhang and Mingchuan Zhang and Y. K. Li and Y. Wu and Daya Guo},
 journal = {ArXiv preprint},
 title = {DeepSeekMath: Pushing the Limits of Mathematical Reasoning in Open Language Models},
 url = {https://arxiv.org/abs/2402.03300},
 volume = {abs/2402.03300},
 year = {2024}
}

@inproceedings{
lambert2025tulu3pushingfrontiers,
title={Tulu 3: Pushing Frontiers in Open Language Model Post-Training},
author={Nathan Lambert and Jacob Morrison and Valentina Pyatkin and Shengyi Huang and Hamish Ivison and Faeze Brahman and Lester James Validad Miranda and Alisa Liu and Nouha Dziri and Xinxi Lyu and Yuling Gu and Saumya Malik and Victoria Graf and Jena D. Hwang and Jiangjiang Yang and Ronan Le Bras and Oyvind Tafjord and Christopher Wilhelm and Luca Soldaini and Noah A. Smith and Yizhong Wang and Pradeep Dasigi and Hannaneh Hajishirzi},
booktitle={Second Conference on Language Modeling},
year={2025},
url={https://openreview.net/forum?id=i1uGbfHHpH}
}

@misc{xie2025logicrlunleashingllmreasoning,
 author = {Tian Xie and Zitian Gao and Qingnan Ren and Haoming Luo and Yuqian Hong and Bryan Dai and Joey Zhou and Kai Qiu and Zhirong Wu and Chong Luo},
 journal = {ArXiv preprint},
 title = {Logic-RL: Unleashing LLM Reasoning with Rule-Based Reinforcement Learning},
 url = {https://arxiv.org/abs/2502.14768},
 volume = {abs/2502.14768},
 year = {2025}
}

@inproceedings{
wang2025reinforcementlearningreasoninglarge,
title={Reinforcement Learning for Reasoning in Large Language Models with One Training Example},
author={Yiping Wang and Qing Yang and Zhiyuan Zeng and Liliang Ren and Liyuan Liu and Baolin Peng and Hao Cheng and Xuehai He and Kuan Wang and Jianfeng Gao and Weizhu Chen and Shuohang Wang and Simon Shaolei Du and yelong shen},
booktitle={The Thirty-ninth Annual Conference on Neural Information Processing Systems 2025, {NeurIPS} 2025, San Diego, CA, USA, December 2–7, 2025},
year={2025},
publisher={OpenReview.net},
url={https://openreview.net/forum?id=IBrRNLr6JA}
}

@inproceedings{
zhao2025echochamberrlposttraining,
title={Echo Chamber: {RL} Post-training Amplifies Behaviors Learned in Pretraining},
author={Rosie Zhao and Alexandru Meterez and Sham M. Kakade and Cengiz Pehlevan and Samy Jelassi and Eran Malach},
booktitle={Second Conference on Language Modeling},
year={2025},
url={https://openreview.net/forum?id=dp4KWuSDzj}
}

@inproceedings{
liu2025understandingr1zeroliketrainingcritical,
title={Understanding R1-Zero-Like Training: A Critical Perspective},
author={Zichen Liu and Changyu Chen and Wenjun Li and Penghui Qi and Tianyu Pang and Chao Du and Wee Sun Lee and Min Lin},
booktitle={Second Conference on Language Modeling},
year={2025},
url={https://openreview.net/forum?id=5PAF7PAY2Y}
}

@inproceedings{dang2025assessing,
 author = {Xingyu Dang and Christina Baek and J Zico Kolter and Aditi Raghunathan},
 booktitle = {Scaling Self-Improving Foundation Models without Human Supervision},
 title = {Assessing Diversity Collapse in Reasoning},
 url = {https://openreview.net/forum?id=AMiKsHLjQh},
 year = {2025}
}

@inproceedings{yue2025limitofrlvr,
 author = {Yue, Yang and Chen, Zhiqi and Lu, Rui and Zhao, Andrew and Wang, Zhaokai and Yue, Yang and Song, Shiji and Huang, Gao},
 booktitle = {Advances in Neural Information Processing Systems},
 title = {Does Reinforcement Learning Really Incentivize Reasoning Capacity in {LLMs} Beyond the Base Model?},
 year = {2025}
}

@article{shoham2024medconceptsqaopensourcemedical,
title = {MedConceptsQA: Open source medical concepts QA benchmark},
journal = {Computers in Biology and Medicine},
volume = {182},
pages = {109089},
year = {2024},
issn = {0010-4825},
doi = {https://doi.org/10.1016/j.compbiomed.2024.109089},
url = {https://www.sciencedirect.com/science/article/pii/S0010482524011740},
author = {Ofir Ben Shoham and Nadav Rappoport}
}

@misc{yang2025qwen3,
      title={Qwen3 Technical Report}, 
      author={An Yang and Anfeng Li and Baosong Yang and Beichen Zhang and Binyuan Hui and Bo Zheng and Bowen Yu and Chang Gao and Chengen Huang and Chenxu Lv and Chujie Zheng and Dayiheng Liu and Fan Zhou and Fei Huang and Feng Hu and Hao Ge and Haoran Wei and Huan Lin and Jialong Tang and Jian Yang and Jianhong Tu and Jianwei Zhang and Jianxin Yang and Jiaxi Yang and Jing Zhou and Jingren Zhou and Junyang Lin and Kai Dang and Keqin Bao and Kexin Yang and Le Yu and Lianghao Deng and Mei Li and Mingfeng Xue and Mingze Li and Pei Zhang and Peng Wang and Qin Zhu and Rui Men and Ruize Gao and Shixuan Liu and Shuang Luo and Tianhao Li and Tianyi Tang and Wenbiao Yin and Xingzhang Ren and Xinyu Wang and Xinyu Zhang and Xuancheng Ren and Yang Fan and Yang Su and Yichang Zhang and Yinger Zhang and Yu Wan and Yuqiong Liu and Zekun Wang and Zeyu Cui and Zhenru Zhang and Zhipeng Zhou and Zihan Qiu},
      year={2025},
      eprint={2505.09388},
      archivePrefix={arXiv},
      primaryClass={cs.CL},
      url={https://arxiv.org/abs/2505.09388}, 
}

@misc{liu2024deepseek,
      title={DeepSeek-V3 Technical Report}, 
      author={DeepSeek-AI and Aixin Liu and Bei Feng and Bing Xue and Bingxuan Wang and Bochao Wu and Chengda Lu and Chenggang Zhao and Chengqi Deng and Chenyu Zhang and Chong Ruan and Damai Dai and Daya Guo and Dejian Yang and Deli Chen and Dongjie Ji and Erhang Li and Fangyun Lin and Fucong Dai and Fuli Luo and Guangbo Hao and Guanting Chen and Guowei Li and H. Zhang and Han Bao and Hanwei Xu and Haocheng Wang and Haowei Zhang and Honghui Ding and Huajian Xin and Huazuo Gao and Hui Li and Hui Qu and J. L. Cai and Jian Liang and Jianzhong Guo and Jiaqi Ni and Jiashi Li and Jiawei Wang and Jin Chen and Jingchang Chen and Jingyang Yuan and Junjie Qiu and Junlong Li and Junxiao Song and Kai Dong and Kai Hu and Kaige Gao and Kang Guan and Kexin Huang and Kuai Yu and Lean Wang and Lecong Zhang and Lei Xu and Leyi Xia and Liang Zhao and Litong Wang and Liyue Zhang and Meng Li and Miaojun Wang and Mingchuan Zhang and Minghua Zhang and Minghui Tang and Mingming Li and Ning Tian and Panpan Huang and Peiyi Wang and Peng Zhang and Qiancheng Wang and Qihao Zhu and Qinyu Chen and Qiushi Du and R. J. Chen and R. L. Jin and Ruiqi Ge and Ruisong Zhang and Ruizhe Pan and Runji Wang and Runxin Xu and Ruoyu Zhang and Ruyi Chen and S. S. Li and Shanghao Lu and Shangyan Zhou and Shanhuang Chen and Shaoqing Wu and Shengfeng Ye and Shengfeng Ye and Shirong Ma and Shiyu Wang and Shuang Zhou and Shuiping Yu and Shunfeng Zhou and Shuting Pan and T. Wang and Tao Yun and Tian Pei and Tianyu Sun and W. L. Xiao and Wangding Zeng and Wanjia Zhao and Wei An and Wen Liu and Wenfeng Liang and Wenjun Gao and Wenqin Yu and Wentao Zhang and X. Q. Li and Xiangyue Jin and Xianzu Wang and Xiao Bi and Xiaodong Liu and Xiaohan Wang and Xiaojin Shen and Xiaokang Chen and Xiaokang Zhang and Xiaosha Chen and Xiaotao Nie and Xiaowen Sun and Xiaoxiang Wang and Xin Cheng and Xin Liu and Xin Xie and Xingchao Liu and Xingkai Yu and Xinnan Song and Xinxia Shan and Xinyi Zhou and Xinyu Yang and Xinyuan Li and Xuecheng Su and Xuheng Lin and Y. K. Li and Y. Q. Wang and Y. X. Wei and Y. X. Zhu and Yang Zhang and Yanhong Xu and Yanhong Xu and Yanping Huang and Yao Li and Yao Zhao and Yaofeng Sun and Yaohui Li and Yaohui Wang and Yi Yu and Yi Zheng and Yichao Zhang and Yifan Shi and Yiliang Xiong and Ying He and Ying Tang and Yishi Piao and Yisong Wang and Yixuan Tan and Yiyang Ma and Yiyuan Liu and Yongqiang Guo and Yu Wu and Yuan Ou and Yuchen Zhu and Yuduan Wang and Yue Gong and Yuheng Zou and Yujia He and Yukun Zha and Yunfan Xiong and Yunxian Ma and Yuting Yan and Yuxiang Luo and Yuxiang You and Yuxuan Liu and Yuyang Zhou and Z. F. Wu and Z. Z. Ren and Zehui Ren and Zhangli Sha and Zhe Fu and Zhean Xu and Zhen Huang and Zhen Zhang and Zhenda Xie and Zhengyan Zhang and Zhewen Hao and Zhibin Gou and Zhicheng Ma and Zhigang Yan and Zhihong Shao and Zhipeng Xu and Zhiyu Wu and Zhongyu Zhang and Zhuoshu Li and Zihui Gu and Zijia Zhu and Zijun Liu and Zilin Li and Ziwei Xie and Ziyang Song and Ziyi Gao and Zizheng Pan},
      year={2025},
      eprint={2412.19437},
      archivePrefix={arXiv},
      primaryClass={cs.CL},
      url={https://arxiv.org/abs/2412.19437}, 
}

@misc{team2024qwen2,
      title={Qwen2 Technical Report}, 
      author={An Yang and Baosong Yang and Binyuan Hui and Bo Zheng and Bowen Yu and Chang Zhou and Chengpeng Li and Chengyuan Li and Dayiheng Liu and Fei Huang and Guanting Dong and Haoran Wei and Huan Lin and Jialong Tang and Jialin Wang and Jian Yang and Jianhong Tu and Jianwei Zhang and Jianxin Ma and Jianxin Yang and Jin Xu and Jingren Zhou and Jinze Bai and Jinzheng He and Junyang Lin and Kai Dang and Keming Lu and Keqin Chen and Kexin Yang and Mei Li and Mingfeng Xue and Na Ni and Pei Zhang and Peng Wang and Ru Peng and Rui Men and Ruize Gao and Runji Lin and Shijie Wang and Shuai Bai and Sinan Tan and Tianhang Zhu and Tianhao Li and Tianyu Liu and Wenbin Ge and Xiaodong Deng and Xiaohuan Zhou and Xingzhang Ren and Xinyu Zhang and Xipin Wei and Xuancheng Ren and Xuejing Liu and Yang Fan and Yang Yao and Yichang Zhang and Yu Wan and Yunfei Chu and Yuqiong Liu and Zeyu Cui and Zhenru Zhang and Zhifang Guo and Zhihao Fan},
      year={2024},
      eprint={2407.10671},
      archivePrefix={arXiv},
      primaryClass={cs.CL},
      url={https://arxiv.org/abs/2407.10671}, 
}

@inproceedings{paterson1994longest,
 author = {Paterson, Mike and Dan{\v{c}}{\'\i}k, Vlado},
 booktitle = {International symposium on mathematical foundations of computer science},
 organization = {Springer},
 pages = {127--142},
 title = {Longest common subsequences},
 year = {1994}
}

@article{buckland1994relationship,
 author = {Buckland, Michael and Gey, Fredric},
 journal = {Journal of the American society for information science},
 number = {1},
 pages = {12--19},
 publisher = {Wiley Online Library},
 title = {The relationship between recall and precision},
 volume = {45},
 year = {1994}
}

@misc{openai2023gpt4,
 author = {Josh Achiam and Steven Adler and Sandhini Agarwal and Lama Ahmad and Ilge Akkaya and Florencia Leoni Aleman and Diogo Almeida and Janko Altenschmidt and Sam Altman and Shyamal Anadkat and Red Avila and Igor Babuschkin and Suchir Balaji and Valerie Balcom and Paul Baltescu and Haiming Bao and Mohammad Bavarian and Jeff Belgum and Irwan Bello and Jake Berdine and Gabriel Bernadett-Shapiro and Christopher Berner and Lenny Bogdonoff and Oleg Boiko and Madelaine Boyd and Anna-Luisa Brakman and Greg Brockman and Tim Brooks and Miles Brundage and Kevin Button and Trevor Cai and Rosie Campbell and Andrew Cann and Brittany Carey and Chelsea Carlson and Rory Carmichael and Brooke Chan and Che Chang and Fotis Chantzis and Derek Chen and Sully Chen and Ruby Chen and Jason Chen and Mark Chen and Ben Chess and Chester Cho and Casey Chu and Hyung Won Chung and Dave Cummings and Jeremiah Currier and Yunxing Dai and Cory Decareaux and Thomas Degry and Noah Deutsch and Damien Deville and Arka Dhar and David Dohan and Steve Dowling and Sheila Dunning and Adrien Ecoffet and Atty Eleti and Tyna Eloundou and David Farhi and Liam Fedus and Niko Felix and Simón Posada Fishman and Juston Forte and Isabella Fulford and Leo Gao and Elie Georges and Christian Gibson and Vik Goel and Tarun Gogineni and Gabriel Goh and Rapha Gontijo-Lopes and Jonathan Gordon and Morgan Grafstein and Scott Gray and Ryan Greene and Joshua Gross and Shawn Guo and Chris Hallacy and Jesse Han and Jeff Harris and Yuchen He and Mike Heaton and Johannes Heidecke and Chris Hesse and Alan Hickey and Wade Hickey and Peter Hoeschele and Brandon Houghton and Kenny Hsu and Shengli Hu and Xin Hu and Joanne Jang and Angela Jiang and Roger Jiang and Haozhun Jin and Denny Jin and Shino Jomoto and Billie Jonn and Heewoo Jun and Tomer Kaftan and Łukasz Kaiser and Ali Kamali and Ingmar Kanitscheider and Nitish Shirish Keskar and Tabarak Khan and Logan Kilpatrick and Jong Wook Kim and Christina Kim and Yongjik Kim and Jan Hendrik Kirchner and Jamie Kiros and Matt Knight and Daniel Kokotajlo and Łukasz Kondraciuk and Andrew Kondrich and Aris Konstantinidis and Kyle Kosic and Gretchen Krueger and Vishal Kuo and Michael Lampe and Ikai Lan and Teddy Lee and Jan Leike and Jade Leung and Daniel Levy and Chak Ming Li and Rachel Lim and Molly Lin and Stephanie Lin and Mateusz Litwin and Theresa Lopez and Ryan Lowe and Patricia Lue and Anna Makanju and Kim Malfacini and Sam Manning and Todor Markov and Yaniv Markovski and Bianca Martin and Katie Mayer and Andrew Mayne and Bob McGrew and Scott Mayer McKinney and Christine McLeavey and Paul McMillan and Jake McNeil and David Medina and Aalok Mehta and Jacob Menick and Luke Metz and Andrey Mishchenko and Pamela Mishkin and Vinnie Monaco and Evan Morikawa and Daniel Mossing and Tong Mu and Mira Murati and Oleg Murk and David Mély and Ashvin Nair and Reiichiro Nakano and Rajeev Nayak and Arvind Neelakantan and Richard Ngo and Hyeonwoo Noh and Long Ouyang and Cullen O'Keefe and Jakub Pachocki and Alex Paino and Joe Palermo and Ashley Pantuliano and Giambattista Parascandolo and Joel Parish and Emy Parparita and Alex Passos and Mikhail Pavlov and Andrew Peng and Adam Perelman and Filipe de Avila Belbute Peres and Michael Petrov and Henrique Ponde de Oliveira Pinto and Michael and Pokorny and Michelle Pokrass and Vitchyr H. Pong and Tolly Powell and Alethea Power and Boris Power and Elizabeth Proehl and Raul Puri and Alec Radford and Jack Rae and Aditya Ramesh and Cameron Raymond and Francis Real and Kendra Rimbach and Carl Ross and Bob Rotsted and Henri Roussez and Nick Ryder and Mario Saltarelli and Ted Sanders and Shibani Santurkar and Girish Sastry and Heather Schmidt and David Schnurr and John Schulman and Daniel Selsam and Kyla Sheppard and Toki Sherbakov and Jessica Shieh and Sarah Shoker and Pranav Shyam and Szymon Sidor and Eric Sigler and Maddie Simens and Jordan Sitkin and Katarina Slama and Ian Sohl and Benjamin Sokolowsky and Yang Song and Natalie Staudacher and Felipe Petroski Such and Natalie Summers and Ilya Sutskever and Jie Tang and Nikolas Tezak and Madeleine B. Thompson and Phil Tillet and Amin Tootoonchian and Elizabeth Tseng and Preston Tuggle and Nick Turley and Jerry Tworek and Juan Felipe Cerón Uribe and Andrea Vallone and Arun Vijayvergiya and Chelsea Voss and Carroll Wainwright and Justin Jay Wang and Alvin Wang and Ben Wang and Jonathan Ward and Jason Wei and CJ Weinmann and Akila Welihinda and Peter Welinder and Jiayi Weng and Lilian Weng and Matt Wiethoff and Dave Willner and Clemens Winter and Samuel Wolrich and Hannah Wong and Lauren Workman and Sherwin Wu and Jeff Wu and Michael Wu and Kai Xiao and Tao Xu and Sarah Yoo and Kevin Yu and Qiming Yuan and Wojciech Zaremba and Rowan Zellers and Chong Zhang and Marvin Zhang and Shengjia Zhao and Tianhao Zheng and Juntang Zhuang and William Zhuk and Barret Zoph},
 journal = {ArXiv preprint},
 title = {GPT-4 Technical Report},
 url = {https://arxiv.org/abs/2303.08774},
 volume = {abs/2303.08774},
 year = {2023}
}

@misc{bai2022constitutional,
 author = {Yuntao Bai and Saurav Kadavath and Sandipan Kundu and Amanda Askell and Jackson Kernion and Andy Jones and Anna Chen and Anna Goldie and Azalia Mirhoseini and Cameron McKinnon and Carol Chen and Catherine Olsson and Christopher Olah and Danny Hernandez and Dawn Drain and Deep Ganguli and Dustin Li and Eli Tran-Johnson and Ethan Perez and Jamie Kerr and Jared Mueller and Jeffrey Ladish and Joshua Landau and Kamal Ndousse and Kamile Lukosuite and Liane Lovitt and Michael Sellitto and Nelson Elhage and Nicholas Schiefer and Noemi Mercado and Nova DasSarma and Robert Lasenby and Robin Larson and Sam Ringer and Scott Johnston and Shauna Kravec and Sheer El Showk and Stanislav Fort and Tamera Lanham and Timothy Telleen-Lawton and Tom Conerly and Tom Henighan and Tristan Hume and Samuel R. Bowman and Zac Hatfield-Dodds and Ben Mann and Dario Amodei and Nicholas Joseph and Sam McCandlish and Tom Brown and Jared Kaplan},
 journal = {ArXiv preprint},
 title = {Constitutional {AI}: Harmlessness from {AI} Feedback},
 url = {https://arxiv.org/abs/2212.08073},
 volume = {abs/2212.08073},
 year = {2022}
}

@misc{askell2021general,
 author = {Amanda Askell and Yuntao Bai and Anna Chen and Dawn Drain and Deep Ganguli and Tom Henighan and Andy Jones and Nicholas Joseph and Ben Mann and Nova DasSarma and Nelson Elhage and Zac Hatfield-Dodds and Danny Hernandez and Jackson Kernion and Kamal Ndousse and Catherine Olsson and Dario Amodei and Tom Brown and Jack Clark and Sam McCandlish and Chris Olah and Jared Kaplan},
 journal = {ArXiv preprint},
 title = {A General Language Assistant as a Laboratory for Alignment},
 url = {https://arxiv.org/abs/2112.00861},
 volume = {abs/2112.00861},
 year = {2021}
}

@inproceedings{guu2020realmretrievalaugmentedlanguagemodel,
  author       = {Kelvin Guu and
                  Kenton Lee and
                  Zora Tung and
                  Panupong Pasupat and
                  Ming{-}Wei Chang},
  title        = {Retrieval Augmented Language Model Pre-Training},
  booktitle    = {Proceedings of the 37th International Conference on Machine Learning,
                  {ICML} 2020, 13-18 July 2020, Virtual Event},
  series       = {Proceedings of Machine Learning Research},
  pages        = {3929--3938},
  publisher    = {{PMLR}},
  year         = {2020},
  url          = {http://proceedings.mlr.press/v119/guu20a.html},
  bibsource    = {dblp computer science bibliography, https://dblp.org}
}

@inproceedings{lewis2021retrievalaugmentedgenerationknowledgeintensivenlp,
 author = {Patrick S. H. Lewis and
Ethan Perez and
Aleksandra Piktus and
Fabio Petroni and
Vladimir Karpukhin and
Naman Goyal and
Heinrich K{\"{u}}ttler and
Mike Lewis and
Wen{-}tau Yih and
Tim Rockt{\"{a}}schel and
Sebastian Riedel and
Douwe Kiela},
 bibsource = {dblp computer science bibliography, https://dblp.org},
 booktitle = {Advances in Neural Information Processing Systems 33: Annual Conference
on Neural Information Processing Systems 2020, NeurIPS 2020, December
6-12, 2020, virtual},
 editor = {Hugo Larochelle and
Marc'Aurelio Ranzato and
Raia Hadsell and
Maria{-}Florina Balcan and
Hsuan{-}Tien Lin},
 title = {Retrieval-Augmented Generation for Knowledge-Intensive {NLP} Tasks},
 url = {https://proceedings.neurips.cc/paper/2020/hash/6b493230205f780e1bc26945df7481e5-Abstract.html},
 year = {2020}
}

@inproceedings{borgeaud2022improvinglanguagemodelsretrieving,
 author = {Sebastian Borgeaud and
Arthur Mensch and
Jordan Hoffmann and
Trevor Cai and
Eliza Rutherford and
Katie Millican and
George van den Driessche and
Jean{-}Baptiste Lespiau and
Bogdan Damoc and
Aidan Clark and
Diego de Las Casas and
Aurelia Guy and
Jacob Menick and
Roman Ring and
Tom Hennigan and
Saffron Huang and
Loren Maggiore and
Chris Jones and
Albin Cassirer and
Andy Brock and
Michela Paganini and
Geoffrey Irving and
Oriol Vinyals and
Simon Osindero and
Karen Simonyan and
Jack W. Rae and
Erich Elsen and
Laurent Sifre},
 bibsource = {dblp computer science bibliography, https://dblp.org},
 booktitle = {International Conference on Machine Learning, {ICML} 2022, 17-23 July
2022, Baltimore, Maryland, {USA}},
 editor = {Kamalika Chaudhuri and
Stefanie Jegelka and
Le Song and
Csaba Szepesv{\'{a}}ri and
Gang Niu and
Sivan Sabato},
 pages = {2206--2240},
 publisher = {{PMLR}},
 series = {Proceedings of Machine Learning Research},
 title = {Improving Language Models by Retrieving from Trillions of Tokens},
 url = {https://proceedings.mlr.press/v162/borgeaud22a.html},
 volume = {162},
 year = {2022}
}

@inproceedings{asai2023selfraglearningretrievegenerate,
 author = {Akari Asai and
Zeqiu Wu and
Yizhong Wang and
Avirup Sil and
Hannaneh Hajishirzi},
 bibsource = {dblp computer science bibliography, https://dblp.org},
 booktitle = {The Twelfth International Conference on Learning Representations,
{ICLR} 2024, Vienna, Austria, May 7-11, 2024},
 publisher = {OpenReview.net},
 title = {Self-RAG: Learning to Retrieve, Generate, and Critique through Self-Reflection},
 url = {https://openreview.net/forum?id=hSyW5go0v8},
 year = {2024}
}

@inproceedings{qin2023toolllmfacilitatinglargelanguage,
 author = {Yujia Qin and
Shihao Liang and
Yining Ye and
Kunlun Zhu and
Lan Yan and
Yaxi Lu and
Yankai Lin and
Xin Cong and
Xiangru Tang and
Bill Qian and
Sihan Zhao and
Lauren Hong and
Runchu Tian and
Ruobing Xie and
Jie Zhou and
Mark Gerstein and
Dahai Li and
Zhiyuan Liu and
Maosong Sun},
 bibsource = {dblp computer science bibliography, https://dblp.org},
 booktitle = {The Twelfth International Conference on Learning Representations,
{ICLR} 2024, Vienna, Austria, May 7-11, 2024},
 publisher = {OpenReview.net},
 title = {ToolLLM: Facilitating Large Language Models to Master 16000+ Real-world
APIs},
 url = {https://openreview.net/forum?id=dHng2O0Jjr},
 year = {2024}
}

@inproceedings{schick2023toolformerlanguagemodelsteach,
 author = {Timo Schick and
Jane Dwivedi{-}Yu and
Roberto Dess{\`{\i}} and
Roberta Raileanu and
Maria Lomeli and
Eric Hambro and
Luke Zettlemoyer and
Nicola Cancedda and
Thomas Scialom},
 bibsource = {dblp computer science bibliography, https://dblp.org},
 booktitle = {Advances in Neural Information Processing Systems 36: Annual Conference
on Neural Information Processing Systems 2023, NeurIPS 2023, New Orleans,
LA, USA, December 10 - 16, 2023},
 editor = {Alice Oh and
Tristan Naumann and
Amir Globerson and
Kate Saenko and
Moritz Hardt and
Sergey Levine},
 title = {Toolformer: Language Models Can Teach Themselves to Use Tools},
 url = {http://papers.nips.cc/paper\_files/paper/2023/hash/d842425e4bf79ba039352da0f658a906-Abstract-Conference.html},
 year = {2023}
}

@misc{nakano2022webgptbrowserassistedquestionansweringhuman,
 author = {Reiichiro Nakano and Jacob Hilton and Suchir Balaji and Jeff Wu and Long Ouyang and Christina Kim and Christopher Hesse and Shantanu Jain and Vineet Kosaraju and William Saunders and Xu Jiang and Karl Cobbe and Tyna Eloundou and Gretchen Krueger and Kevin Button and Matthew Knight and Benjamin Chess and John Schulman},
 journal = {ArXiv preprint},
 title = {WebGPT: Browser-assisted question-answering with human feedback},
 url = {https://arxiv.org/abs/2112.09332},
 volume = {abs/2112.09332},
 year = {2021}
}

@inproceedings{yao2023reactsynergizingreasoningacting,
 author = {Shunyu Yao and
Jeffrey Zhao and
Dian Yu and
Nan Du and
Izhak Shafran and
Karthik R. Narasimhan and
Yuan Cao},
 bibsource = {dblp computer science bibliography, https://dblp.org},
 booktitle = {The Eleventh International Conference on Learning Representations,
{ICLR} 2023, Kigali, Rwanda, May 1-5, 2023},
 publisher = {OpenReview.net},
 title = {ReAct: Synergizing Reasoning and Acting in Language Models},
 url = {https://openreview.net/pdf?id=WE\_vluYUL-X},
 year = {2023}
}

@misc{ye2026cramfitmoretraining,
 author = {Jiayuan Ye and Vitaly Feldman and Kunal Talwar},
 journal = {ArXiv preprint},
 title = {Cram Less to Fit More: Training Data Pruning Improves Memorization of Facts},
 url = {https://arxiv.org/abs/2604.08519},
 volume = {abs/2604.08519},
 year = {2026}
}

@inproceedings{liang2025reasoningrag12,
    title = "Reasoning {RAG} via System 1 or System 2: A Survey on Reasoning Agentic Retrieval-Augmented Generation for Industry Challenges",
    author = "Liang, Jintao  and
      Sugang  and
      Lin, Huifeng  and
      Wu, You  and
      Zhao, Rui  and
      Li, Ziyue",
    editor = "Inui, Kentaro  and
      Sakti, Sakriani  and
      Wang, Haofen  and
      Wong, Derek F.  and
      Bhattacharyya, Pushpak  and
      Banerjee, Biplab  and
      Ekbal, Asif  and
      Chakraborty, Tanmoy  and
      Singh, Dhirendra Pratap",
    booktitle = "Proceedings of the 14th International Joint Conference on Natural Language Processing and the 4th Conference of the Asia-Pacific Chapter of the Association for Computational Linguistics",
    month = dec,
    year = "2025",
    address = "Mumbai, India",
    publisher = "The Asian Federation of Natural Language Processing and The Association for Computational Linguistics",
    url = "https://aclanthology.org/2025.findings-ijcnlp.122/",
    doi = "10.18653/v1/2025.findings-ijcnlp.122",
    pages = "1954--1966",
    ISBN = "979-8-89176-303-6"
}

@misc{singh2026agenticretrievalaugmentedgenerationsurvey,
 author = {Aditi Singh and Abul Ehtesham and Saket Kumar and Tala Talaei Khoei and Athanasios V. Vasilakos},
 journal = {ArXiv preprint},
 title = {Agentic Retrieval-Augmented Generation: A Survey on Agentic RAG},
 url = {https://arxiv.org/abs/2501.09136},
 volume = {abs/2501.09136},
 year = {2025}
}

@inproceedings{sanh2022multitaskpromptedtrainingenables,
 author = {Victor Sanh and
Albert Webson and
Colin Raffel and
Stephen H. Bach and
Lintang Sutawika and
Zaid Alyafeai and
Antoine Chaffin and
Arnaud Stiegler and
Arun Raja and
Manan Dey and
M Saiful Bari and
Canwen Xu and
Urmish Thakker and
Shanya Sharma Sharma and
Eliza Szczechla and
Taewoon Kim and
Gunjan Chhablani and
Nihal V. Nayak and
Debajyoti Datta and
Jonathan Chang and
Mike Tian{-}Jian Jiang and
Han Wang and
Matteo Manica and
Sheng Shen and
Zheng Xin Yong and
Harshit Pandey and
Rachel Bawden and
Thomas Wang and
Trishala Neeraj and
Jos Rozen and
Abheesht Sharma and
Andrea Santilli and
Thibault F{\'{e}}vry and
Jason Alan Fries and
Ryan Teehan and
Teven Le Scao and
Stella Biderman and
Leo Gao and
Thomas Wolf and
Alexander M. Rush},
 bibsource = {dblp computer science bibliography, https://dblp.org},
 booktitle = {The Tenth International Conference on Learning Representations, {ICLR}
2022, Virtual Event, April 25-29, 2022},
 publisher = {OpenReview.net},
 title = {Multitask Prompted Training Enables Zero-Shot Task Generalization},
 url = {https://openreview.net/forum?id=9Vrb9D0WI4},
 year = {2022}
}

@inproceedings{kojima2023largelanguagemodelszeroshot,
 author = {Takeshi Kojima and
Shixiang Shane Gu and
Machel Reid and
Yutaka Matsuo and
Yusuke Iwasawa},
 bibsource = {dblp computer science bibliography, https://dblp.org},
 booktitle = {Advances in Neural Information Processing Systems 35: Annual Conference
on Neural Information Processing Systems 2022, NeurIPS 2022, New Orleans,
LA, USA, November 28 - December 9, 2022},
 editor = {Sanmi Koyejo and
S. Mohamed and
A. Agarwal and
Danielle Belgrave and
K. Cho and
A. Oh},
 title = {Large Language Models are Zero-Shot Reasoners},
 url = {http://papers.nips.cc/paper\_files/paper/2022/hash/8bb0d291acd4acf06ef112099c16f326-Abstract-Conference.html},
 year = {2022}
}

@inproceedings{brown2020languagemodelsfewshotlearners,
 author = {Tom B. Brown and
Benjamin Mann and
Nick Ryder and
Melanie Subbiah and
Jared Kaplan and
Prafulla Dhariwal and
Arvind Neelakantan and
Pranav Shyam and
Girish Sastry and
Amanda Askell and
Sandhini Agarwal and
Ariel Herbert{-}Voss and
Gretchen Krueger and
Tom Henighan and
Rewon Child and
Aditya Ramesh and
Daniel M. Ziegler and
Jeffrey Wu and
Clemens Winter and
Christopher Hesse and
Mark Chen and
Eric Sigler and
Mateusz Litwin and
Scott Gray and
Benjamin Chess and
Jack Clark and
Christopher Berner and
Sam McCandlish and
Alec Radford and
Ilya Sutskever and
Dario Amodei},
 bibsource = {dblp computer science bibliography, https://dblp.org},
 booktitle = {Advances in Neural Information Processing Systems 33: Annual Conference
on Neural Information Processing Systems 2020, NeurIPS 2020, December
6-12, 2020, virtual},
 editor = {Hugo Larochelle and
Marc'Aurelio Ranzato and
Raia Hadsell and
Maria{-}Florina Balcan and
Hsuan{-}Tien Lin},
 title = {Language Models are Few-Shot Learners},
 url = {https://proceedings.neurips.cc/paper/2020/hash/1457c0d6bfcb4967418bfb8ac142f64a-Abstract.html},
 year = {2020}
}

@inproceedings{wei2022chain,
 author = {Jason Wei and
Xuezhi Wang and
Dale Schuurmans and
Maarten Bosma and
Brian Ichter and
Fei Xia and
Ed H. Chi and
Quoc V. Le and
Denny Zhou},
 bibsource = {dblp computer science bibliography, https://dblp.org},
 booktitle = {Advances in Neural Information Processing Systems 35: Annual Conference
on Neural Information Processing Systems 2022, NeurIPS 2022, New Orleans,
LA, USA, November 28 - December 9, 2022},
 editor = {Sanmi Koyejo and
S. Mohamed and
A. Agarwal and
Danielle Belgrave and
K. Cho and
A. Oh},
 title = {Chain-of-Thought Prompting Elicits Reasoning in Large Language Models},
 url = {http://papers.nips.cc/paper\_files/paper/2022/hash/9d5609613524ecf4f15af0f7b31abca4-Abstract-Conference.html},
 year = {2022}
}

@inproceedings{zhou2023least,
 author = {Denny Zhou and
Nathanael Sch{\"{a}}rli and
Le Hou and
Jason Wei and
Nathan Scales and
Xuezhi Wang and
Dale Schuurmans and
Claire Cui and
Olivier Bousquet and
Quoc V. Le and
Ed H. Chi},
 bibsource = {dblp computer science bibliography, https://dblp.org},
 booktitle = {The Eleventh International Conference on Learning Representations,
{ICLR} 2023, Kigali, Rwanda, May 1-5, 2023},
 publisher = {OpenReview.net},
 title = {Least-to-Most Prompting Enables Complex Reasoning in Large Language
Models},
 url = {https://openreview.net/pdf?id=WZH7099tgfM},
 year = {2023}
}

@inproceedings{dbpedia_ontology,
    title = "{DB}pedia: A Multilingual Cross-domain Knowledge Base",
    author = "Mendes, Pablo  and
      Jakob, Max  and
      Bizer, Christian",
    editor = "Calzolari, Nicoletta  and
      Choukri, Khalid  and
      Declerck, Thierry  and
      Do{\u{g}}an, Mehmet U{\u{g}}ur  and
      Maegaard, Bente  and
      Mariani, Joseph  and
      Moreno, Asuncion  and
      Odijk, Jan  and
      Piperidis, Stelios",
    booktitle = "Proceedings of the Eighth International Conference on Language Resources and Evaluation ({LREC}'12)",
    month = may,
    year = "2012",
    address = "Istanbul, Turkey",
    publisher = "European Language Resources Association (ELRA)",
    url = "https://aclanthology.org/L12-1323/",
    pages = "1813--1817"
}

@misc{ma2026improvingparametricknowledgeaccess,
 author = {Melody Ma and John Hewitt},
 journal = {ArXiv preprint},
 title = {Improving Parametric Knowledge Access in Reasoning Language Models},
 url = {https://arxiv.org/abs/2602.22193},
 volume = {abs/2602.22193},
 year = {2026}
}

@misc{gekhman2026thinkingrecallreasoningunlocks,
 author = {Zorik Gekhman and Roee Aharoni and Eran Ofek and Mor Geva and Roi Reichart and Jonathan Herzig},
 journal = {ArXiv preprint},
 title = {Thinking to Recall: How Reasoning Unlocks Parametric Knowledge in LLMs},
 url = {https://arxiv.org/abs/2603.09906},
 volume = {abs/2603.09906},
 year = {2026}
}

@inproceedings{lin2024mitigating,
 author = {Lin, Yong and Lin, Hangyu and Xiong, Wei and Diao, Shizhe and Liu, Jianmeng and Zhang, Jipeng and Pan, Rui and Wang, Haoxiang and Hu, Wenbin and Zhang, Hanning and Dong, Hanze and Pi, Renjie and Zhao, Han and Jiang, Nan and Ji, Heng and Yao, Yuan and Zhang, Tong},
 booktitle = {Proceedings of the 2024 Conference on Empirical Methods in Natural Language Processing},
 organization = {Association for Computational Linguistics},
 title = {Mitigating the Alignment Tax of {RLHF}},
 year = {2024}
}

@inproceedings{allen2024physics,
 author = {Zeyuan Allen{-}Zhu and
Yuanzhi Li},
 bibsource = {dblp computer science bibliography, https://dblp.org},
 booktitle = {Forty-first International Conference on Machine Learning, {ICML} 2024,
Vienna, Austria, July 21-27, 2024},
 publisher = {OpenReview.net},
 title = {Physics of Language Models: Part 3.1, Knowledge Storage and Extraction},
 url = {https://openreview.net/forum?id=5x788rqbcj},
 year = {2024}
}

@inproceedings{
agrawal2025gepa,
title={{GEPA}: Reflective Prompt Evolution Can Outperform Reinforcement Learning},
author={Lakshya A Agrawal and Shangyin Tan and Dilara Soylu and Noah Ziems and Rishi Khare and Krista Opsahl-Ong and Arnav Singhvi and Herumb Shandilya and Michael J Ryan and Meng Jiang and Christopher Potts and Koushik Sen and Alex Dimakis and Ion Stoica and Dan Klein and Matei Zaharia and Omar Khattab},
booktitle={The Fourteenth International Conference on Learning Representations, {ICLR} 2026, Rio de Janeiro, Brazil, April 23-27, 2026},
year={2026},
url={https://openreview.net/forum?id=RQm2KQTM5r},
 publisher = {OpenReview.net},
}

@article{wu2025invisible,
 author = {Wu, Fang and Xuan, Weihao and Lu, Ximing and Liu, Mingjie and Dong, Yi and Harchaoui, Zaid and Choi, Yejin},
 journal = {ArXiv preprint},
 title = {The Invisible Leash: Why RLVR May or May Not Escape Its Origin},
 url = {https://arxiv.org/abs/2507.14843},
 volume = {abs/2507.14843},
 year = {2025}
}

@inproceedings{berglund2024reversal,
 author = {Lukas Berglund and
Meg Tong and
Maximilian Kaufmann and
Mikita Balesni and
Asa Cooper Stickland and
Tomasz Korbak and
Owain Evans},
 bibsource = {dblp computer science bibliography, https://dblp.org},
 booktitle = {The Twelfth International Conference on Learning Representations,
{ICLR} 2024, Vienna, Austria, May 7-11, 2024},
 publisher = {OpenReview.net},
 title = {The Reversal Curse: LLMs trained on "A is B" fail to learn "B is A"},
 url = {https://openreview.net/forum?id=GPKTIktA0k},
 year = {2024}
}

@inproceedings{
yu2025dapo,
title={{DAPO}: An Open-Source {LLM} Reinforcement Learning System at Scale},
author={Qiying Yu and Zheng Zhang and Ruofei Zhu and Yufeng Yuan and Xiaochen Zuo and YuYue and Weinan Dai and Tiantian Fan and Gaohong Liu and Juncai Liu and LingJun Liu and Xin Liu and Haibin Lin and Zhiqi Lin and Bole Ma and Guangming Sheng and Yuxuan Tong and Chi Zhang and Mofan Zhang and Ru Zhang and Wang Zhang and Hang Zhu and Jinhua Zhu and Jiaze Chen and Jiangjie Chen and Chengyi Wang and Hongli Yu and Yuxuan Song and Xiangpeng Wei and Hao Zhou and Jingjing Liu and Wei-Ying Ma and Ya-Qin Zhang and Lin Yan and Yonghui Wu and Mingxuan Wang},
booktitle={The Thirty-ninth Annual Conference on Neural Information Processing Systems 2025, {NeurIPS} 2025, San Diego, CA, USA, December 2-7, 2025},
publisher={OpenReview.net},
year={2026},
url={https://openreview.net/forum?id=2a36EMSSTp}
}

@article{wang2025octothinker,
 author = {Wang, Zengzhi and Zhou, Fan and Li, Xuefeng and Liu, Pengfei},
 journal = {ArXiv preprint},
 title = {Octothinker: Mid-training incentivizes reinforcement learning scaling},
 url = {https://arxiv.org/abs/2506.20512},
 volume = {abs/2506.20512},
 year = {2025}
}

@inproceedings{ghosh2024closer,
 author = {Sreyan Ghosh and
Chandra Kiran Reddy Evuru and
Sonal Kumar and
Ramaneswaran S. and
Deepali Aneja and
Zeyu Jin and
Ramani Duraiswami and
Dinesh Manocha},
 bibsource = {dblp computer science bibliography, https://dblp.org},
 booktitle = {Forty-first International Conference on Machine Learning, {ICML} 2024,
Vienna, Austria, July 21-27, 2024},
 publisher = {OpenReview.net},
 title = {A Closer Look at the Limitations of Instruction Tuning},
 url = {https://openreview.net/forum?id=XkHJo8iXGQ},
 year = {2024}
}

@inproceedings{gekhman2024does,
    title = "Does Fine-Tuning {LLM}s on New Knowledge Encourage Hallucinations?",
    author = "Gekhman, Zorik  and
      Yona, Gal  and
      Aharoni, Roee  and
      Eyal, Matan  and
      Feder, Amir  and
      Reichart, Roi  and
      Herzig, Jonathan",
    editor = "Al-Onaizan, Yaser  and
      Bansal, Mohit  and
      Chen, Yun-Nung",
    booktitle = "Proceedings of the 2024 Conference on Empirical Methods in Natural Language Processing",
    month = nov,
    year = "2024",
    address = "Miami, Florida, USA",
    publisher = "Association for Computational Linguistics",
    url = "https://aclanthology.org/2024.emnlp-main.444/",
    doi = "10.18653/v1/2024.emnlp-main.444",
    pages = "7765--7784"
}

@article{yuan2024towards,
 author = {Yuan, Jiaqing and Pan, Lin and Hang, Chung-Wei and Guo, Jiang and Jiang, Jiarong and Min, Bonan and Ng, Patrick and Wang, Zhiguo},
 journal = {ArXiv preprint},
 title = {Towards a holistic evaluation of llms on factual knowledge recall},
 url = {https://arxiv.org/abs/2404.16164},
 volume = {abs/2404.16164},
 year = {2024}
}

@inproceedings{
sorensen2025spectrum,
title={Spectrum Tuning: Post-Training for Distributional Coverage and In-Context Steerability},
author={Taylor Sorensen and Benjamin Newman and Jared Moore and Chan Young Park and Jillian Fisher and Niloofar Mireshghallah and Liwei Jiang and Yejin Choi},
booktitle={The Fourteenth International Conference on Learning Representations, {ICLR} 2026, Rio de Janeiro, Brazil, April 23-27, 2026},
year={2026},
url={https://openreview.net/forum?id=ulvp7cbZeU},
 publisher = {OpenReview.net},
}

@inproceedings{
chu2025sft,
title={{SFT} Memorizes, {RL} Generalizes: A Comparative Study of Foundation Model Post-training},
author={Tianzhe Chu and Yuexiang Zhai and Jihan Yang and Shengbang Tong and Saining Xie and Dale Schuurmans and Quoc V Le and Sergey Levine and Yi Ma},
booktitle={Forty-second International Conference on Machine Learning, {ICML 2025}, Vancouver, Canada, July 13-19, 2025.},
year={2025},
url={https://openreview.net/forum?id=dYur3yabMj},
publisher = {OpenReview.net}
}

@inproceedings{
khattab2023dspy,
title={{DSP}y: Compiling Declarative Language Model Calls into State-of-the-Art Pipelines},
author={Omar Khattab and Arnav Singhvi and Paridhi Maheshwari and Zhiyuan Zhang and Keshav Santhanam and Sri Vardhamanan A and Saiful Haq and Ashutosh Sharma and Thomas T. Joshi and Hanna Moazam and Heather Miller and Matei Zaharia and Christopher Potts},
booktitle={The Twelfth International Conference on Learning Representations, {ICLR} 2024, Vienna, Austria, May 7-11, 2024},
year={2024},
url={https://openreview.net/forum?id=sY5N0zY5Od},
publisher = {OpenReview.net},
}

@article{ziems2025multi,
 author = {Ziems, Noah and Soylu, Dilara and Agrawal, Lakshya A and Miller, Isaac and Lai, Liheng and Qian, Chen and Song, Kaiqiang and Jiang, Meng and Klein, Dan and Zaharia, Matei and others},
 journal = {ArXiv preprint},
 title = {Multi-module GRPO: Composing policy gradients and prompt optimization for language model programs},
 url = {https://arxiv.org/abs/2508.04660},
 volume = {abs/2508.04660},
 year = {2025}
}

@article{huan2025does,
 author = {Huan, Maggie and Li, Yuetai and Zheng, Tuney and Xu, Xiaoyu and Kim, Seungone and Du, Minxin and Poovendran, Radha and Neubig, Graham and Yue, Xiang},
 journal = {ArXiv preprint},
 title = {Does Math Reasoning Improve General LLM Capabilities? Understanding Transferability of LLM Reasoning},
 url = {https://arxiv.org/abs/2507.00432},
 volume = {abs/2507.00432},
 year = {2025}
}

@inproceedings{skean2025layer,
  author       = {Oscar Skean and
                  Md Rifat Arefin and
                  Dan Zhao and
                  Niket Patel and
                  Jalal Naghiyev and
                  Yann LeCun and
                  Ravid Shwartz{-}Ziv},
  editor       = {Aarti Singh and
                  Maryam Fazel and
                  Daniel Hsu and
                  Simon Lacoste{-}Julien and
                  Felix Berkenkamp and
                  Tegan Maharaj and
                  Kiri Wagstaff and
                  Jerry Zhu},
  title        = {Layer by Layer: Uncovering Hidden Representations in Language Models},
  booktitle    = {Forty-second International Conference on Machine Learning, {ICML}
                  2025, Vancouver, BC, Canada, July 13-19, 2025},
  series       = {Proceedings of Machine Learning Research},
  publisher    = {{PMLR} / OpenReview.net},
  year         = {2025},
  url          = {https://proceedings.mlr.press/v267/skean25a.html},
  bibsource    = {dblp computer science bibliography, https://dblp.org}
}

@inproceedings{mukherjee2025reinforcement,
 author = {Sagnik Mukherjee and Lifan Yuan and Dilek Hakkani-T{\"u}r and Hao Peng},
 booktitle = {The Thirty-ninth Annual Conference on Neural Information Processing Systems},
 title = {Reinforcement Learning Finetunes Small Subnetworks in Large Language Models},
 url = {https://openreview.net/forum?id=0NdS4xCngO},
 year = {2025}
}

@inproceedings{agarwal2024policy,
 author = {Rishabh Agarwal and
Nino Vieillard and
Yongchao Zhou and
Piotr Stanczyk and
Sabela Ramos Garea and
Matthieu Geist and
Olivier Bachem},
 bibsource = {dblp computer science bibliography, https://dblp.org},
 booktitle = {The Twelfth International Conference on Learning Representations,
{ICLR} 2024, Vienna, Austria, May 7-11, 2024},
 publisher = {OpenReview.net},
 title = {On-Policy Distillation of Language Models: Learning from Self-Generated
Mistakes},
 url = {https://openreview.net/forum?id=3zKtaqxLhW},
 year = {2024}
}

@article{liu2025knowledge,
 author = {Li, Junliang and Wang, Yucheng and Chen, Yan and Ran, Yu and Zhang, Ruiqing and Liu, Jing and Wu, Hua and Wang, Haifeng},
 journal = {ArXiv preprint},
 title = {Knowledge-Level Consistency Reinforcement Learning: Dual-Fact Alignment for Long-Form Factuality},
 url = {https://arxiv.org/abs/2509.23765},
 volume = {abs/2509.23765},
 year = {2025}
}

@inproceedings{hu2024understanding,
 author = {Robert Kirk and
Ishita Mediratta and
Christoforos Nalmpantis and
Jelena Luketina and
Eric Hambro and
Edward Grefenstette and
Roberta Raileanu},
 bibsource = {dblp computer science bibliography, https://dblp.org},
 booktitle = {The Twelfth International Conference on Learning Representations,
{ICLR} 2024, Vienna, Austria, May 7-11, 2024},
 publisher = {OpenReview.net},
 title = {Understanding the Effects of {RLHF} on {LLM} Generalisation and Diversity},
 url = {https://openreview.net/forum?id=PXD3FAVHJT},
 year = {2024}
}

@article{sorensen2024beyond,
 author = {Phan, Hoang and Yang, Xianjun and Yao, Kevin and Zhang, Jingyu and Bi, Shengjie and Tang, Xiaocheng and Khabsa, Madian and Liu, Lijuan and Lei, Deren},
 journal = {ArXiv preprint},
 title = {Beyond Reasoning Gains: Mitigating General Capabilities Forgetting in Large Reasoning Models},
 url = {https://arxiv.org/abs/2510.21978},
 volume = {abs/2510.21978},
 year = {2025}
}

@article{liu2025deepseek,
  author       = {Daya Guo and
                  Dejian Yang and
                  Haowei Zhang and
                  Junxiao Song and
                  Peiyi Wang and
                  Qihao Zhu and
                  Runxin Xu and
                  Ruoyu Zhang and
                  Shirong Ma and
                  Xiao Bi and
                  Xiaokang Zhang and
                  Xingkai Yu and
                  Yu Wu and
                  Z. F. Wu and
                  Zhibin Gou and
                  Zhihong Shao and
                  Zhuoshu Li and
                  Ziyi Gao and
                  Aixin Liu and
                  Bing Xue and
                  Bingxuan Wang and
                  Bochao Wu and
                  Bei Feng and
                  Chengda Lu and
                  Chenggang Zhao and
                  Chengqi Deng and
                  Chong Ruan and
                  Damai Dai and
                  Deli Chen and
                  Dongjie Ji and
                  Erhang Li and
                  Fangyun Lin and
                  Fucong Dai and
                  Fuli Luo and
                  Guangbo Hao and
                  Guanting Chen and
                  Guowei Li and
                  Hao Zhang and
                  Hanwei Xu and
                  Honghui Ding and
                  Huazuo Gao and
                  Hui Qu and
                  Hui Li and
                  Jianzhong Guo and
                  Jiashi Li and
                  Jingchang Chen and
                  Jingyang Yuan and
                  Jinhao Tu and
                  Junjie Qiu and
                  Junlong Li and
                  J. L. Cai and
                  Jiaqi Ni and
                  Jian Liang and
                  Jin Chen and
                  Kai Dong and
                  Kai Hu and
                  Kaichao You and
                  Kaige Gao and
                  Kang Guan and
                  Kexin Huang and
                  Kuai Yu and
                  Lean Wang and
                  Lecong Zhang and
                  Liang Zhao and
                  Litong Wang and
                  Liyue Zhang and
                  Lei Xu and
                  Leyi Xia and
                  Mingchuan Zhang and
                  Minghua Zhang and
                  Minghui Tang and
                  Mingxu Zhou and
                  Meng Li and
                  Miaojun Wang and
                  Mingming Li and
                  Ning Tian and
                  Panpan Huang and
                  Peng Zhang and
                  Qiancheng Wang and
                  Qinyu Chen and
                  Qiushi Du and
                  Ruiqi Ge and
                  Ruisong Zhang and
                  Ruizhe Pan and
                  Runji Wang and
                  R. J. Chen and
                  R. L. Jin and
                  Ruyi Chen and
                  Shanghao Lu and
                  Shangyan Zhou and
                  Shanhuang Chen and
                  Shengfeng Ye and
                  Shiyu Wang and
                  Shuiping Yu and
                  Shunfeng Zhou and
                  Shuting Pan and
                  S. S. Li and
                  Shuang Zhou and
                  Shaoqing Wu and
                  Tao Yun and
                  Tian Pei and
                  Tianyu Sun and
                  Tao Wang and
                  Wangding Zeng and
                  Wen Liu and
                  Wenfeng Liang and
                  Wenjun Gao and
                  Wenqin Yu and
                  Wentao Zhang and
                  W. L. Xiao and
                  Wei An and
                  Xiaodong Liu and
                  Xiaohan Wang and
                  Xiaokang Chen and
                  Xiaotao Nie and
                  Xin Cheng and
                  Xin Liu and
                  Xin Xie and
                  Xingchao Liu and
                  Xinyu Yang and
                  Xinyuan Li and
                  Xuecheng Su and
                  Xuheng Lin and
                  X. Q. Li and
                  Xiangyue Jin and
                  Xiaojin Shen and
                  Xiaosha Chen and
                  Xiaowen Sun and
                  Xiaoxiang Wang and
                  Xinnan Song and
                  Xinyi Zhou and
                  Xianzu Wang and
                  Xinxia Shan and
                  Y. K. Li and
                  Y. Q. Wang and
                  Y. X. Wei and
                  Yang Zhang and
                  Yanhong Xu and
                  Yao Li and
                  Yao Zhao and
                  Yaofeng Sun and
                  Yaohui Wang and
                  Yi Yu and
                  Yichao Zhang and
                  Yifan Shi and
                  Yiliang Xiong and
                  Ying He and
                  Yishi Piao and
                  Yisong Wang and
                  Yixuan Tan and
                  Yiyang Ma and
                  Yiyuan Liu and
                  Yongqiang Guo and
                  Yuan Ou and
                  Yuduan Wang and
                  Yue Gong and
                  Yuheng Zou and
                  Yujia He and
                  Yunfan Xiong and
                  Yuxiang Luo and
                  Yuxiang You and
                  Yuxuan Liu and
                  Yuyang Zhou and
                  Y. X. Zhu and
                  Yanping Huang and
                  Yaohui Li and
                  Yi Zheng and
                  Yuchen Zhu and
                  Yunxian Ma and
                  Ying Tang and
                  Yukun Zha and
                  Yuting Yan and
                  Z. Z. Ren and
                  Zehui Ren and
                  Zhangli Sha and
                  Zhe Fu and
                  Zhean Xu and
                  Zhenda Xie and
                  Zhengyan Zhang and
                  Zhewen Hao and
                  Zhicheng Ma and
                  Zhigang Yan and
                  Zhiyu Wu and
                  Zihui Gu and
                  Zijia Zhu and
                  Zijun Liu and
                  Zilin Li and
                  Ziwei Xie and
                  Ziyang Song and
                  Zizheng Pan and
                  Zhen Huang and
                  Zhipeng Xu and
                  Zhongyu Zhang and
                  Zhen Zhang},
  title        = {DeepSeek-R1 incentivizes reasoning in LLMs through reinforcement learning},
  journal      = {Nat.},
  volume       = {645},
  number       = {8081},
  pages        = {633--638},
  year         = {2025},
  url          = {https://doi.org/10.1038/s41586-025-09422-z},
  doi          = {10.1038/S41586-025-09422-Z},
  bibsource    = {dblp computer science bibliography, https://dblp.org}
}

@inproceedings{mallen2023trustlanguagemodelsinvestigating,
 address = {Toronto, Canada},
 author = {Mallen, Alex  and
Asai, Akari  and
Zhong, Victor  and
Das, Rajarshi  and
Khashabi, Daniel  and
Hajishirzi, Hannaneh},
 booktitle = {Proceedings of the 61st Annual Meeting of the Association for Computational Linguistics (Volume 1: Long Papers)},
 doi = {10.18653/v1/2023.acl-long.546},
 editor = {Rogers, Anna  and
Boyd-Graber, Jordan  and
Okazaki, Naoaki},
 pages = {9802--9822},
 publisher = {Association for Computational Linguistics},
 title = {When Not to Trust Language Models: Investigating Effectiveness of Parametric and Non-Parametric Memories},
 url = {https://aclanthology.org/2023.acl-long.546},
 year = {2023}
}

@misc{wipo2026ipc,
 author = {{World Intellectual Property Organization}},
 howpublished = {\url{https://www.wipo.int/en/web/classification-ipc/}},
 note = {Accessed: 2026},
 title = {International Patent Classification ({IPC}), Version 2026.01},
 year = {2026}
}

@misc{kimiteam2026kimik25visualagentic,
 author = {Kimi Team and Tongtong Bai and Yifan Bai and Yiping Bao and S. H. Cai and Yuan Cao and Y. Charles and H. S. Che and Cheng Chen and Guanduo Chen and Huarong Chen and Jia Chen and Jiahao Chen and Jianlong Chen and Jun Chen and Kefan Chen and Liang Chen and Ruijue Chen and Xinhao Chen and Yanru Chen and Yanxu Chen and Yicun Chen and Yimin Chen and Yingjiang Chen and Yuankun Chen and Yujie Chen and Yutian Chen and Zhirong Chen and Ziwei Chen and Dazhi Cheng and Minghan Chu and Jialei Cui and Jiaqi Deng and Muxi Diao and Hao Ding and Mengfan Dong and Mengnan Dong and Yuxin Dong and Yuhao Dong and Angang Du and Chenzhuang Du and Dikang Du and Lingxiao Du and Yulun Du and Yu Fan and Shengjun Fang and Qiulin Feng and Yichen Feng and Garimugai Fu and Kelin Fu and Hongcheng Gao and Tong Gao and Yuyao Ge and Shangyi Geng and Chengyang Gong and Xiaochen Gong and Zhuoma Gongque and Qizheng Gu and Xinran Gu and Yicheng Gu and Longyu Guan and Yuanying Guo and Xiaoru Hao and Weiran He and Wenyang He and Yunjia He and Chao Hong and Hao Hu and Jiaxi Hu and Yangyang Hu and Zhenxing Hu and Ke Huang and Ruiyuan Huang and Weixiao Huang and Zhiqi Huang and Tao Jiang and Zhejun Jiang and Xinyi Jin and Yu Jing and Guokun Lai and Aidi Li and C. Li and Cheng Li and Fang Li and Guanghe Li and Guanyu Li and Haitao Li and Haoyang Li and Jia Li and Jingwei Li and Junxiong Li and Lincan Li and Mo Li and Weihong Li and Wentao Li and Xinhang Li and Xinhao Li and Yang Li and Yanhao Li and Yiwei Li and Yuxiao Li and Zhaowei Li and Zheming Li and Weilong Liao and Jiawei Lin and Xiaohan Lin and Zhishan Lin and Zichao Lin and Cheng Liu and Chenyu Liu and Hongzhang Liu and Liang Liu and Shaowei Liu and Shudong Liu and Shuran Liu and Tianwei Liu and Tianyu Liu and Weizhou Liu and Xiangyan Liu and Yangyang Liu and Yanming Liu and Yibo Liu and Yuanxin Liu and Yue Liu and Zhengying Liu and Zhongnuo Liu and Enzhe Lu and Haoyu Lu and Zhiyuan Lu and Junyu Luo and Tongxu Luo and Yashuo Luo and Long Ma and Yingwei Ma and Shaoguang Mao and Yuan Mei and Xin Men and Fanqing Meng and Zhiyong Meng and Yibo Miao and Minqing Ni and Kun Ouyang and Siyuan Pan and Bo Pang and Yuchao Qian and Ruoyu Qin and Zeyu Qin and Jiezhong Qiu and Bowen Qu and Zeyu Shang and Youbo Shao and Tianxiao Shen and Zhennan Shen and Juanfeng Shi and Lidong Shi and Shengyuan Shi and Feifan Song and Pengwei Song and Tianhui Song and Xiaoxi Song and Hongjin Su and Jianlin Su and Zhaochen Su and Lin Sui and Jinsong Sun and Junyao Sun and Tongyu Sun and Flood Sung and Yunpeng Tai and Chuning Tang and Heyi Tang and Xiaojuan Tang and Zhengyang Tang and Jiawen Tao and Shiyuan Teng and Chaoran Tian and Pengfei Tian and Ao Wang and Bowen Wang and Chensi Wang and Chuang Wang and Congcong Wang and Dingkun Wang and Dinglu Wang and Dongliang Wang and Feng Wang and Hailong Wang and Haiming Wang and Hengzhi Wang and Huaqing Wang and Hui Wang and Jiahao Wang and Jinhong Wang and Jiuzheng Wang and Kaixin Wang and Linian Wang and Qibin Wang and Shengjie Wang and Shuyi Wang and Si Wang and Wei Wang and Xiaochen Wang and Xinyuan Wang and Yao Wang and Yejie Wang and Yipu Wang and Yiqin Wang and Yucheng Wang and Yuzhi Wang and Zhaoji Wang and Zhaowei Wang and Zhengtao Wang and Zhexu Wang and Zihan Wang and Zizhe Wang and Chu Wei and Ming Wei and Chuan Wen and Zichen Wen and Chengjie Wu and Haoning Wu and Junyan Wu and Rucong Wu and Wenhao Wu and Yuefeng Wu and Yuhao Wu and Yuxin Wu and Zijian Wu and Chenjun Xiao and Jin Xie and Xiaotong Xie and Yuchong Xie and Yifei Xin and Bowei Xing and Boyu Xu and Jianfan Xu and Jing Xu and Jinjing Xu and L. H. Xu and Lin Xu and Suting Xu and Weixin Xu and Xinbo Xu and Xinran Xu and Yangchuan Xu and Yichang Xu and Yuemeng Xu and Zelai Xu and Ziyao Xu and Junjie Yan and Yuzi Yan and Guangyao Yang and Hao Yang and Junwei Yang and Kai Yang and Ningyuan Yang and Ruihan Yang and Xiaofei Yang and Xinlong Yang and Ying Yang and Yi Yang and Yi Yang and Zhen Yang and Zhilin Yang and Zonghan Yang and Haotian Yao and Dan Ye and Wenjie Ye and Zhuorui Ye and Bohong Yin and Chengzhen Yu and Longhui Yu and Tao Yu and Tianxiang Yu and Enming Yuan and Mengjie Yuan and Xiaokun Yuan and Yang Yue and Weihao Zeng and Dunyuan Zha and Haobing Zhan and Dehao Zhang and Hao Zhang and Jin Zhang and Puqi Zhang and Qiao Zhang and Rui Zhang and Xiaobin Zhang and Y. Zhang and Yadong Zhang and Yangkun Zhang and Yichi Zhang and Yizhi Zhang and Yongting Zhang and Yu Zhang and Yushun Zhang and Yutao Zhang and Yutong Zhang and Zheng Zhang and Chenguang Zhao and Feifan Zhao and Jinxiang Zhao and Shuai Zhao and Xiangyu Zhao and Yikai Zhao and Zijia Zhao and Huabin Zheng and Ruihan Zheng and Shaojie Zheng and Tengyang Zheng and Junfeng Zhong and Longguang Zhong and Weiming Zhong and M. Zhou and Runjie Zhou and Xinyu Zhou and Zaida Zhou and Jinguo Zhu and Liya Zhu and Xinhao Zhu and Yuxuan Zhu and Zhen Zhu and Jingze Zhuang and Weiyu Zhuang and Ying Zou and Xinxing Zu},
 journal = {ArXiv preprint},
 title = {Kimi K2.5: Visual Agentic Intelligence},
 url = {https://arxiv.org/abs/2602.02276},
 volume = {abs/2602.02276},
 year = {2026}
}

@misc{kimiteam2026kimik2openagentic,
 author = {Kimi Team and Yifan Bai and Yiping Bao and Y. Charles and Cheng Chen and Guanduo Chen and Haiting Chen and Huarong Chen and Jiahao Chen and Ningxin Chen and Ruijue Chen and Yanru Chen and Yuankun Chen and Yutian Chen and Zhuofu Chen and Jialei Cui and Hao Ding and Mengnan Dong and Angang Du and Chenzhuang Du and Dikang Du and Yulun Du and Yu Fan and Yichen Feng and Kelin Fu and Bofei Gao and Chenxiao Gao and Hongcheng Gao and Peizhong Gao and Tong Gao and Yuyao Ge and Shangyi Geng and Qizheng Gu and Xinran Gu and Longyu Guan and Haiqing Guo and Jianhang Guo and Xiaoru Hao and Tianhong He and Weiran He and Wenyang He and Yunjia He and Chao Hong and Hao Hu and Yangyang Hu and Zhenxing Hu and Weixiao Huang and Zhiqi Huang and Zihao Huang and Tao Jiang and Zhejun Jiang and Xinyi Jin and Yongsheng Kang and Guokun Lai and Cheng Li and Fang Li and Haoyang Li and Ming Li and Wentao Li and Yang Li and Yanhao Li and Yiwei Li and Zhaowei Li and Zheming Li and Hongzhan Lin and Xiaohan Lin and Zongyu Lin and Chengyin Liu and Chenyu Liu and Hongzhang Liu and Jingyuan Liu and Junqi Liu and Liang Liu and Shaowei Liu and T. Y. Liu and Tianwei Liu and Weizhou Liu and Yangyang Liu and Yibo Liu and Yiping Liu and Yue Liu and Zhengying Liu and Enzhe Lu and Haoyu Lu and Lijun Lu and Yashuo Luo and Shengling Ma and Xinyu Ma and Yingwei Ma and Shaoguang Mao and Jie Mei and Xin Men and Yibo Miao and Siyuan Pan and Yebo Peng and Ruoyu Qin and Zeyu Qin and Bowen Qu and Zeyu Shang and Lidong Shi and Shengyuan Shi and Feifan Song and Jianlin Su and Zhengyuan Su and Lin Sui and Xinjie Sun and Flood Sung and Yunpeng Tai and Heyi Tang and Jiawen Tao and Qifeng Teng and Chaoran Tian and Chensi Wang and Dinglu Wang and Feng Wang and Hailong Wang and Haiming Wang and Jianzhou Wang and Jiaxing Wang and Jinhong Wang and Shengjie Wang and Shuyi Wang and Si Wang and Xinyuan Wang and Yao Wang and Yejie Wang and Yiqin Wang and Yuxin Wang and Yuzhi Wang and Zhaoji Wang and Zhengtao Wang and Zhengtao Wang and Zhexu Wang and Chu Wei and Qianqian Wei and Haoning Wu and Wenhao Wu and Xingzhe Wu and Yuxin Wu and Chenjun Xiao and Jin Xie and Xiaotong Xie and Weimin Xiong and Boyu Xu and Jinjing Xu and L. H. Xu and Lin Xu and Suting Xu and Weixin Xu and Xinran Xu and Yangchuan Xu and Ziyao Xu and Jing Xu and Jing Xu and Junjie Yan and Yuzi Yan and Hao Yang and Xiaofei Yang and Yi Yang and Ying Yang and Zhen Yang and Zhilin Yang and Zonghan Yang and Haotian Yao and Xingcheng Yao and Wenjie Ye and Zhuorui Ye and Bohong Yin and Longhui Yu and Enming Yuan and Hongbang Yuan and Mengjie Yuan and Siyu Yuan and Haobing Zhan and Dehao Zhang and Hao Zhang and Wanlu Zhang and Xiaobin Zhang and Yadong Zhang and Yangkun Zhang and Yichi Zhang and Yizhi Zhang and Yongting Zhang and Yu Zhang and Yutao Zhang and Yutong Zhang and Zheng Zhang and Haotian Zhao and Yikai Zhao and Zijia Zhao and Huabin Zheng and Shaojie Zheng and Longguang Zhong and Jianren Zhou and Xinyu Zhou and Zaida Zhou and Jinguo Zhu and Zhen Zhu and Weiyu Zhuang and Xinxing Zu},
 journal = {ArXiv preprint},
 title = {Kimi K2: Open Agentic Intelligence},
 url = {https://arxiv.org/abs/2507.20534},
 volume = {abs/2507.20534},
 year = {2025}
}

@inproceedings{
liu2025infinigramscalingunboundedngram,
title={Infini-gram: Scaling Unbounded n-gram Language Models to a Trillion Tokens},
author={Jiacheng Liu and Sewon Min and Luke Zettlemoyer and Yejin Choi and Hannaneh Hajishirzi},
booktitle={First Conference on Language Modeling},
year={2024},
url={https://openreview.net/forum?id=u2vAyMeLMm}
}

\clearpage
\appendix

\section*{Appendix}
\section{Related Work}
\label{sec:related_work}

\subsection{The Alignment Tax and Factual Degradation}
\label{sec:alignment_tax}

The trade-off between alignment and factual accuracy has been extensively explored. \citet{lin2024mitigating} introduced the concept of ``alignment tax'', showing systematic degradation on factual benchmarks as RLHF reward strength increases. \citet{openai2023gpt4} similarly reported that RLHF ``does not improve exam performance (without active effort, it actually degrades it)'' and can reduce calibration. Mechanistic analyses in \citet{ghosh2024closer} reveal that instruction tuning primarily adjusts style rather than new knowledge. In particular, responses generated from pre-trained knowledge consistently outperform those from models learning new knowledge through instruction tuning. Both \citet{liu2025knowledge} and \citet{hu2024understanding} show that base models' parametric knowledge originates from pretraining while aligned models learn how to express it. In addition, \citet{liu2025knowledge} shows that training directly from base models mitigates knowledge forgetting and alignment tax incurred by SFT-based distillation. Recent work by \citet{sorensen2024beyond} shows a trade-off in 
Reinforcement Learning from Verifiable Rewards (RLVR): while optimizing for 
verifiable rewards improves reasoning on targeted tasks, it can degrade general 
capabilities acquired from pretraining and increase hallucinations.

While these studies document factual degradation from alignment, our work reveals a contrasting phenomenon: reasoning models \emph{outperform} their base counterparts on structured knowledge recall. This suggests that alignment tax may not uniformly affect all forms of parametric knowledge retrieval, especially when retrieval needs systematic navigation rather than direct factual recall.

\subsection{Retrieval-augmented generation (RAG)} 
RAG grounds model outputs by retrieving relevant documents at test time and conditioning generation on the retrieved context~\citep{guu2020realmretrievalaugmentedlanguagemodel, lewis2021retrievalaugmentedgenerationknowledgeintensivenlp, borgeaud2022improvinglanguagemodelsretrieving}. A related line of work extends RAG to tool-calling and agentic settings~\citep{nakano2022webgptbrowserassistedquestionansweringhuman, asai2023selfraglearningretrievegenerate, qin2023toolllmfacilitatinglargelanguage, schick2023toolformerlanguagemodelsteach, yao2023reactsynergizingreasoningacting, 
liang2025reasoningrag12, 
singh2026agenticretrievalaugmentedgenerationsurvey, ye2026cramfitmoretraining}, equipping models with external tools, APIs, and web search alongside planning, reflection, and multi-agent collaboration. These systems can dynamically decide when to retrieve, what evidence to gather, and how to integrate it across reasoning steps. Our work is orthogonal: instead of augmenting LLMs through external 
retrieval, we ask whether models can reliably access factual knowledge already stored 
in their parameters. This isolates parametric knowledge access from retrieval quality, 
tool selection, context construction, and the reliability of external pipelines.

\subsection{RL and Prompting for Reasoning}
\noindent RL and prompting have emerged as the two central approaches for eliciting long-form reasoning behaviors in LLMs. Systems such as OpenAI o1 and DeepSeek-R1 show that RL can encourage models to produce extended chains of thought, self-reflect, backtrack, and verify intermediate steps \citep{openai2023gpt4, shao2024deepseekmathpushinglimitsmathematical, liu2025deepseek, lambert2025tulu3pushingfrontiers, xie2025logicrlunleashingllmreasoning, wang2025reinforcementlearningreasoninglarge, yu2025dapo}. Consequently, this leads to strong gains on mathematics, coding, and other verifiable reasoning tasks. Optimized prompting can match or exceed fine-tuned and reasoning models on reasoning tasks \citep{brown2020languagemodelsfewshotlearners, sanh2022multitaskpromptedtrainingenables, wei2022chain, kojima2023largelanguagemodelszeroshot, zhou2023least, agrawal2025gepa}. 

Our work differs from both lines of research. While prior work studies RL and prompting as ways to improve reasoning, we investigate whether RL also improves a model's ability to retrieve parametric knowledge. We do not draw a hard boundary between reasoning and navigation; rather, we aim to isolate what RL improves in our setting—the traversal procedure used to access information, or the underlying knowledge itself. To this end, we use prompting strictly as a diagnostic tool. Because the prompt only tells a model how to navigate a knowledge hierarchy, success relies on accessing existing parametric memory, allowing us to attribute performance gaps to inefficient knowledge access rather than missing information. 
\subsection{Reasoning for Knowledge Retrieval}
\noindent Reasoning over parametric knowledge recall has gained attention. \citet{ma2026improvingparametricknowledgeaccess} show that RL on QA tasks improves factual retrieval on various out-of-distribution benchmarks, but do not isolate whether the gains come from improved knowledge access or from data exposure. Through a frequency-stratified, relation-disjoint experiment, we show that RL on rare facts surfaces common, previously non-extractable knowledge, thus decoupling access from exposure. \citet{gekhman2026thinkingrecallreasoningunlocks} study single-hop factual questions and identify computational buffering and factual priming as the drivers behind reasoning-aided recall. While their work focuses on expanding the knowledge boundary via pass@k and test-time recall improvements, we focus on reliable knowledge access via majority-voted accuracy and systematically compare how SFT, RL, and distillation affect this knowledge access. Finally, while these works study unstructured factual recall, our work investigates explicit and implicit hierarchical navigation within structured and unstructured knowledge. We also isolate SFT/RL/distillation's internal mechanisms through a layer-wise representational analysis.

\subsection{Knowledge Storage versus Knowledge Access}
\label{sec:knowledge_access}
Recent evidence suggests that improving how a model accesses existing knowledge may be equally important as expanding what it knows. \citet{allen2024physics} and \citet{berglund2024reversal} show that models trained on ``A is B'' often fail to answer the reverse query ``B is A''. This indicates that factual knowledge can be directionally inaccessible even when it was seen during training. An analogous pattern has been observed for reasoning: RL improves the sampling efficiency of correct solution paths rather than expanding the base model's reasoning boundary \citep{dang2025assessing,liu2025understandingr1zeroliketrainingcritical, wu2025invisible, yue2025limitofrlvr,zhao2025echochamberrlposttraining}. Whether a similar access-versus-storage distinction holds for parametric factual knowledge under RL remains unstudied. To address this gap, we systematically compare the effect of RL against other post-training methods including distillation and SFT, using prompting interventions and layerwise representation analysis. We further validate our findings through majority-vote response distributions, knowledge-injection experiments, and distillation failure-mode analysis.

\section{Technical Appendices and Supplementary Material}
\subsection{Terminology}
\label{app:terminology}
Table~\ref{tab:terminology} defines the two recurring notions---\emph{knowledge access} and \emph{hierarchical traversal}---used throughout the paper.

\begin{table}[t]
\centering
\small
\begin{tabular}{@{}>{\raggedright\arraybackslash}p{0.26\columnwidth}p{0.66\columnwidth}@{}}
\toprule
\textbf{Term} & \textbf{Definition} \\
\midrule
Knowledge access (retrieval) & A model's \emph{ability} to surface knowledge already encoded in its parameters; a fact can be stored yet inaccessible. \\
\addlinespace[2pt]
Hierarchical traversal (navigation) & The \emph{procedure} of navigating explicit taxonomies or implicit entity relations step by step (e.g., chapter $\rightarrow$ category $\rightarrow$ code) to reach the queried fact. \\
\bottomrule
\end{tabular}
\caption{Terminology used throughout the paper.}
\label{tab:terminology}
\end{table}

\subsection{Zero-Shot Prompt Templates}
\label{app: prompts}
We present the three zero-shot prompt templates used in IPC and MedConceptsQA in Figures~\ref{fig:prompt_direct},~\ref{fig:prompt_cot},~and~\ref{fig:prompt_structured}:
\begin{itemize}[itemsep=0pt, parsep=0pt, topsep=2pt]
    \item \textbf{Direct QA}: the model outputs only a single answer letter with no explanation.
    \item \textbf{CoT}: the model provides a final answer together with a free-form explanation.
    \item \textbf{Structured}: the model is instructed to recall the hierarchical breakdown
    of relevant facts or codes before reaching the final answer.
\end{itemize}

\subsection{Evaluation Protocol for Structured Prompting Experiments}
\label{app:evaluation}
For PopQA, we use exact-match over open-ended answers; for IPC and MedConceptsQA, we use multiple-choice accuracy. To avoid any confound from in-context demonstrations on knowledge access, we evaluate all model pairs in a zero-shot setting. Due to compute constraints, we only evaluate on 5,000 randomly selected questions from PopQA.

\subsection{Dataset Construction Details}
\subsubsection{IPC-Lookup} 
\label{app:ipc_lookup}
\textbf{Dataset Construction.} Two representative questions from the IPC dataset are shown in Figures~\ref{fig:ipc_example_1} and~\ref{fig:ipc_example_2}. Each question
presents a patent code and four candidate descriptions; the correct answer requires recalling the precise code description within the IPC taxonomy. Distractors are drawn from structurally adjacent nodes (sharing the same
parent or grandparent), avoiding coarse domain elimination. Questions containing code strings in the description are excluded to prevent leakage. All correct answers are randomly assigned to A/B/C/D to avoid position bias.
All questions in the dataset were constructed manually and cross-checked by two annotators. Two example questions are provided below.

\begin{figure}[H]
\begin{tcolorbox}[
    colback=pink!15!white,
    colframe=pink!80!black,
    title=\textbf{Example 1 — IPC-Lookup E01B 11/50}
]
\small
\textbf{Question:} What does the patent code E01B 11/50 in the International Patent Classification describe?

\vspace{0.5em}
\textbf{A.} Joint constructions for relatively movable rails, e.g. rails on turntables, traversers, or swing bridges\\
\textbf{B.} Non-dismountable rail joints; Welded joints; e.g., joints made by electric welding\\
\textbf{C.} Special arrangements for supporting rail ends\\
\textbf{D.} Electrically-insulating rail joints

\vspace{0.5em}
\textbf{Correct Answer: A} \quad \textit{(Subgroup of E01B 11/00)}
\end{tcolorbox}
\caption{IPC example question for code \texttt{E01B 11/00}.}
\label{fig:ipc_example_1}
\end{figure}

\begin{figure}[H]
\begin{tcolorbox}[
    colback=pink!15!white,
    colframe=pink!80!black,
    title=\textbf{Example 2 — IPC-Lookup H01B 7/20}
]
\small
\textbf{Question:} What does the patent code \texttt{H01B 7/20} in the
International Patent Classification describe?

\vspace{0.5em}
\textbf{A.} Extensible conductors or cables, e.g.\ self-coiling cords\\
\textbf{B.} Protection against damage caused by external factors, by wear,
mechanical force or pressure; metal tubes, e.g.\ lead sheaths\\
\textbf{C.} Submarine cables\\
\textbf{D.} Rigid-tube cables

\vspace{0.5em}
\textbf{Correct Answer: B} \quad \textit{(Subgroup of H01B 7/00)}
\end{tcolorbox}
\caption{IPC example question for code \texttt{H01B 7/20}.}
\label{fig:ipc_example_2}
\end{figure}

\subsubsection{IPC-CAR}
\label{app:ipc_car}
We use the NCA task to probe the fine-grained recall capabilities of instruct and reasoning models because it forces the model to traverse upward paths in the hierarchy rather than recognize a single code. The dataset was generated by prompting Claude 4.6 Sonnet with the parsing rules, task specification, and constraints shown in Figure~\ref{app:ipc_car_prompt}. Distractors are sampled from neighbors of the correct answer. Two annotators then cross-check every question for hierarchical paths and answer correctness. We construct questions with depths ranging from 2 to 6, with 20 questions per depth and 100 questions in total. Answer options are randomly permuted across A/B/C/D to avoid position bias. Two example questions at retrieval depths 3 and 6 are shown in Figures~\ref{fig:ipc_car_example1} and~\ref{fig:ipc_car_example2}.

\begin{figure}[H]
\begin{tcolorbox}[
    colback=pink!15!white,
    colframe=pink!80!black,
    title=\textbf{Example 1 --- IPC-CAR Depth 3}
]
\small
\textbf{Question:} Nearest common ancestor of A47L 5/24 and A47L 5/30 is: \\
A) A47L 5/22 \quad B) A47L 5/16 \quad C) A47L 5/12 \quad D) A47L 5/14
\vspace{0.5em}

\textbf{Hierarchical Paths:}
\begin{itemize}[itemsep=0pt, parsep=0pt, topsep=2pt]
    \item A47L 5/24 $\rightarrow$ A47L 5/22
    \item A47L 5/30 $\rightarrow$ A47L 5/28 $\rightarrow$ A47L 5/22
\end{itemize}
\textbf{Answer:} A
\end{tcolorbox}
\caption{IPC common-ancestor example at depth 3.}
\label{fig:ipc_car_example1}
\end{figure}

\begin{figure}[H]
\begin{tcolorbox}[
    colback=pink!15!white,
    colframe=pink!80!black,
    title=\textbf{Example 2 --- IPC-CAR Depth 6}
]
\small
\textbf{Question:} Nearest common ancestor of A61K 31/713 and A61K 31/7064 is: \\
A) A61K 31/095 \quad B) A61K 31/70 \quad C) A61K 35/22 \quad D) A61K 8/73
\vspace{0.5em}

\textbf{Hierarchical Paths:}
\begin{itemize}[itemsep=0pt, parsep=0pt, topsep=2pt]
    \item A61K 31/713 $\rightarrow$ A61K 31/7088 $\rightarrow$ A61K 31/70
    \item A61K 31/7064 $\rightarrow$ A61K 31/706 $\rightarrow$ A61K 31/7052 $\rightarrow$ A61K 31/7042 $\rightarrow$ A61K 31/70
\end{itemize}
\textbf{Answer:} B
\end{tcolorbox}
\caption{IPC common-ancestor example at depth 6.}
\label{fig:ipc_car_example2}
\end{figure}

\subsubsection{DBO-CAR}
\label{app:dbo_car}
We construct 500 questions across depths 2--6 (100 per depth), following
the same procedure as IPC-CAR. Figures~\ref{fig:dbo_car_example1} and~\ref{fig:dbo_car_example2} are two representative examples from DBO-CAR.

\begin{figure}[H]
\begin{tcolorbox}[
    colback=pink!15!white,
    colframe=pink!80!black,
    title=\textbf{Example 1 --- DBO-CAR \texttt{Depth}$=2$}
]
\small
\textbf{Question:} Nearest common ancestor of \texttt{SoftballLeague}
and \texttt{MixedMartialArtsLeague} is:
\vspace{0.5em}

\textbf{A.} \texttt{SambaSchool} \\
\textbf{B.} \texttt{SportsLeague} \\
\textbf{C.} \texttt{InternationalOrganisation} \\
\textbf{D.} \texttt{Broadcaster}
\vspace{0.5em}

\textbf{Hierarchical Paths:}
\begin{itemize}[itemsep=0pt, parsep=0pt, topsep=2pt]
    \item \texttt{SoftballLeague} $\rightarrow$ \texttt{SportsLeague}
    \item \texttt{MixedMartialArtsLeague} $\rightarrow$ \texttt{SportsLeague}
\end{itemize}

\textbf{Answer:} B
\end{tcolorbox}
\caption{DBO-CAR common-ancestor example at \texttt{depth} 2.}
\label{fig:dbo_car_example1}
\end{figure}

\begin{figure}[H]
\begin{tcolorbox}[
    colback=pink!15!white,
    colframe=pink!80!black,
    title=\textbf{Example 2 --- DBO-CAR \texttt{Depth} 5}
]
\small
\textbf{Question:} Nearest common ancestor of
\texttt{CultivatedVariety} and \texttt{Psychologist} is:
\vspace{0.5em}

\textbf{A.} \texttt{Person} \\
\textbf{B.} \texttt{Animal} \\
\textbf{C.} \texttt{Eukaryote} \\
\textbf{D.} \texttt{Bacteria}
\vspace{0.5em}

\textbf{Hierarchical Paths:}
\begin{itemize}[itemsep=0pt, parsep=0pt, topsep=2pt]
    \item \texttt{CultivatedVariety} $\rightarrow$ \texttt{Plant} $\rightarrow$ \texttt{Eukaryote}
    \item \texttt{Psychologist} $\rightarrow$ \texttt{Person} $\rightarrow$ \texttt{Animal} $\rightarrow$ \texttt{Eukaryote}
\end{itemize}

\textbf{Answer:} C
\end{tcolorbox}
\caption{DBO-CAR common-ancestor example at \texttt{depth} 5.}
\label{fig:dbo_car_example2}
\end{figure}
\subsection{Controlled Knowledge Injection through RL}
\label{sec:rl_exp}
\begin{figure}[t]
    \centering
    \includegraphics[width=0.85\linewidth]{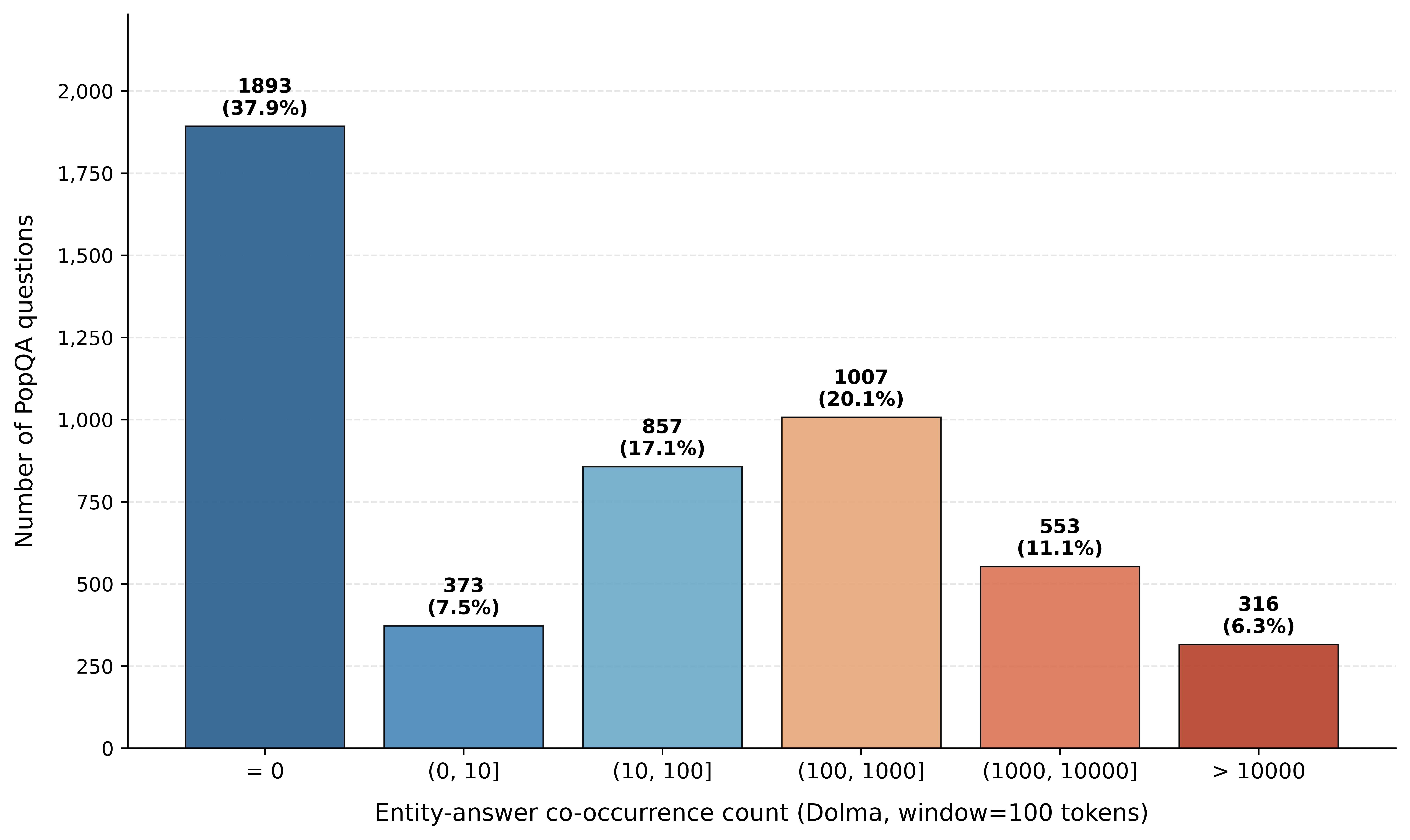}
    \caption{Distribution of PopQA entity-answer co-occurrence counts in the Dolma pre-training corpus.}
    \label{fig:popqa_cooc_distribution}
\end{figure}

\textbf{Entity-Answer Co-occurrence Rates.} First we randomly sample 5{,}000  PopQA questions, then we extract the subject entity and relation using DeepSeek-V3.1-NonThink with the prompt shown in Figure~\ref{fig:popqa_extraction_prompt}. We use Infini-gram to count entity-answer co-occurrences in the Dolma pre-training corpus within a window of \texttt{max\_diff\_tokens}=100 (the default). Questions are binned by co-occurrence count into four groups: $a = 0$, $0 < a \leq 10$, $10 < a \leq 100$, $100 < a \leq 1{,}000$, $1{,}000 < a \leq 10{,}000$ and $a > 10{,}000$. The $a = 0$ bin isolates unseen facts and serves as the source for the low-frequency, inaccessible training set; the $a > 100$ bin is merged from the original $(100, 1{,}000]$, $(1{,}000, 10{,}000]$ and $(10{,}000, \infty)$ ranges to ensure a sufficiently large high-frequency, inaccessible evaluation set. Per-bin counts are shown in Figure~\ref{fig:popqa_cooc_distribution}.

\noindent \textbf{Inaccessible Subgroups.} We evaluate Qwen3-8B (Instruct) on the high-frequency and low-frequency bins, retaining only questions that the model answers incorrectly across three independent runs. They are subsequently filtered to contain disjoint sets of relation as shown in Table~\ref{tab:relation_disjoint}. This yields 731 questions as the low-frequency, inaccessible training set and 208 questions as the high-frequency, inaccessible held-out evaluation set. The identical pipeline is repeated for Qwen3-30B-A3B.

\noindent \textbf{SFT/RL training details.} The learning rate for each model is selected by a grid search over $1\!\times\!10^{-4}$, $5\!\times\!10^{-4}$, $1\!\times\!10^{-5}$, $4\!\times\!10^{-5}$,
$1\!\times\!10^{-6}$, $5\!\times\!10^{-6}$, $8\!\times\!10^{-6}$. Each model is evaluated on a randomly chosen subset of 150 questions from low-frequency, inaccessible training set for 30 RL steps. For each learning rate, we compute the mean over the last third of steps as the selection metric instead of the overall reward, avoiding early reward spikes. The reward is a graded contains-check against the list of acceptable answers: 1.0 if any synonym appears in the response and 0.0 for an incorrect response.  Advantages are computed using Group Relative Policy Optimization (GRPO). The RL hyperparameters are reported in Table~\ref{tab:rl_hyperparams} (same for both the learning rate search and the main RL training). SFT uses the same setup but instead trains the model to predict a randomly sampled candidate answer given the question. For the main training, the selected rate is $4\!\times\!10^{-5}$ for Qwen3-8B-Instruct and $5\!\times\!10^{-4}$ for Qwen3-30B-A3B-Instruct; all runs converge within 140 steps and we report results at this checkpoint. 

\addtocounter{table}{2}
\begin{table*}[!t]
\centering
\small
\begin{tabular}{@{}lrr@{}}
\toprule
\textbf{Comparison} & \textbf{$\Delta$ Acc. (pp)} & \textbf{95\% paired-bootstrap CI (pp)} \\
\midrule
Static RFT $-$ SFT & $+1.44$ & $[-0.96,\ 3.85]$ \\
Iterated RFT $-$ Static RFT & $+7.21$ & $[3.37,\ 11.54]$ \\
Epoch-refresh GRPO $-$ no-refresh GRPO & $+4.33$ & $[0.48,\ 8.65]$ \\
Online GRPO $-$ epoch-refresh GRPO & $+2.40$ & $[-2.40,\ 7.21]$ \\
Online GRPO $-$ no-refresh GRPO & $+6.73$ & $[2.88,\ 11.06]$ \\
Online GRPO $-$ Iterated RFT & $+0.48$ & $[-3.85,\ 4.81]$ \\
\bottomrule
\end{tabular}
\caption{Held-out accuracy with bootstrap confidence intervals.}
\label{tab:rl_decomposition_pairwise}
\end{table*}
\addtocounter{table}{-3}

\begin{table}[h]
\centering
\caption{Per-relation counts for the held-out test set and the RL training set after enforcing disjoint relations.}
\label{tab:relation_disjoint}
\small
\begin{tabular}{lrr}
\toprule
\textbf{Relation} & \texttt{test} & \texttt{train} \\
\midrule
director       &   0 & 180 \\
screenwriter   &   0 & 163 \\
genre          &   0 & 161 \\
producer       &   0 & 114 \\
author         &   0 & 113 \\
\midrule
composer       & 100 &   0 \\
father         &  27 &   0 \\
capital        &  19 &   0 \\
capital of     &  18 &   0 \\
place of birth &  11 &   0 \\
mother         &   8 &   0 \\
occupation     &   8 &   0 \\
religion       &   7 &   0 \\
sport          &   5 &   0 \\
country        &   5 &   0 \\
\midrule
\textbf{Total} & \textbf{208} & \textbf{731} \\
\bottomrule
\end{tabular}
\end{table}

\begin{table}[t]
\centering
\caption{RL training hyperparameters for the controlled knowledge-injection experiment.}
\label{tab:rl_hyperparams}
\small
\begin{tabular}{@{}lp{0.55\linewidth}@{}}
\toprule
\textbf{Hyperparameter} & \textbf{Value} \\
\midrule
\multicolumn{2}{l}{\textit{LoRA configuration}} \\
\quad LoRA rank               & 32 \\
\midrule
\multicolumn{2}{l}{\textit{RL training}} \\
\quad Algorithm               & GRPO \\
\quad Batch size              & 16 \\
\quad Group size              & 16 \\
\quad Number of epochs        & 3 \\
\quad Sampling temperature    & 0.7 \\
\quad Evaluation temperature  & 0.0 \\
\quad LR scheduler            & constant \\
\quad Optimizer & Adam ($\beta_1{=}0.9$, $\beta_2{=}0.95$, \\
                & \phantom{Adam (}$\epsilon{=}10^{-8}$) \\
\quad Random seed            & 42 \\
\quad Max token      & 10{,}000 tokens \\
\bottomrule
\end{tabular}
\end{table}

\noindent \textbf{Wilson Confidence Intervals.}
\label{app:wilson_intervals}
For each method, we treat the correctness of held-out questions as a binary vector $y_i\in\{0,1\}$. For the accuracies
in Figure~\ref{fig:rl_decomposition}, we compute
$\hat p=x/n$, where $x=\sum_{i=1}^{n}y_i$ and $n=208$, and report a 95\%
Wilson score interval. With $z=1.96$, let
\begin{equation}
\begin{aligned}
c &= \frac{\hat p+z^2/(2n)}{1+z^2/n}, \\
h &= \frac{z\sqrt{\hat p(1-\hat p)/n+z^2/(4n^2)}}{1+z^2/n}.
\end{aligned}
\end{equation}
The reported interval is $[c-h,c+h]$. 

\noindent \textbf{Paired Bootstrap Confidence Intervals.}
For comparison between methods A and B, we retain their binary correctness
vectors on the same $n=208$ held-out questions and compute the
accuracy difference
\begin{equation}
\widehat{\Delta}
= \frac{1}{n}\sum_{i=1}^{n}
\left(y_i^{(A)}-y_i^{(B)}\right).
\end{equation}
Then we draw question indices with
replacement and recompute the difference for that bootstrap sample. 
We repeat this procedure for 100{,}000 samples using seed 20260825 and report
the 2.5th and 97.5th percentiles of the resulting differences as the 95\%
paired-bootstrap interval. 

\subsection{Layerwise Similarity Metric: Formal Definition}
\label{app:similarity_metric}
This section gives the formal definition of the cosine similarity metric used in Section~\ref{sec:representation_analysis},  including the set of source pairs over which inter-model and intra-model comparisons
are computed.

\noindent \textbf{Comparison Metric.} For each layer $\ell \in \{1, \ldots, L\}$, we use cosine similarity, a measure of directional alignment, to define representation similarity:
\begin{equation}
{\text{cos}}^{(a,b)}(\ell)
= \frac{1}{N} \sum_{i=1}^{N}
\frac{\mathbf{h}_{\ell}^{(a)}(i)^{\top}\mathbf{h}_{\ell}^{(b)}(i)}
{\|\mathbf{h}_{\ell}^{(a)}(i)\|_2\,\|\mathbf{h}_{\ell}^{(b)}(i)\|_2}. 
\end{equation}

\noindent Here, $\mathbf{h}_{\ell}^{(s)}(i)$ denotes the layer-$\ell$ hidden representation for probe $i$ from a source $s$. The set of sources 
$\mathcal{S} = \{\text{Q}^{\text{base}}, \text{A}^{\text{base}}, \text{Q}^{\text{post-trained}}, \text{A}^{\text{post-trained}}\}$ includes representations for both the question (Q) and answer (A) components from the base and post-trained (instruct/reasoning/distilled) models. Pair $(a,b)$ represents either inter-model (e.g., $\text{Q}^\text{base} \text{ vs } \text{Q}^{\text{post-trained}}, \text{A}^\text{base} \text{ vs } \text{A}^{\text{post-trained}}$) or intra-model comparisons (e.g., $\text{Q}^\text{base} \text{ vs } \text{A}^{\text{base}}, \text{Q}^\text{post-trained} \text{ vs } \text{A}^{\text{post-trained}}$). 

\section{Extended Experimental Results}
\label{app:technical}
\subsection{Extended Model Performance Across Prompt Types}

Tables~\ref{tab:reasoning_comparison_popqa}--\ref{tab:reasoning_comparison_medconceptsqa} report majority vote and mean accuracy for all model pairs across the three prompt types on PopQA, IPC, and MedConceptsQA. Across all datasets, the instruct--reasoning gap consistently narrows as prompt complexity increases.

\subsection{Reasoning Trajectories Analysis}
\label{app:cot_analysis}
Figures~\ref{fig:reasoning_traces_kimi_popqa}--\ref{fig:reasoning_traces} show reasoning traces across prompts for DeepSeek-V3, Qwen3-30B-A3B-Instruct, Qwen3-235B-A22B-Instruct, Kimi-K2.5-Instruct and their respective reasoning models. On MedConceptsQA, DeepSeek-V3 and Qwen3-235B-A22B-Instruct both fail under direct QA and CoT but succeed under structured prompting via explicit hierarchical traversal. DeepSeek-R1 and Qwen3-235B-A22B-Think arrive at the correct answer spontaneously without any structural guidance, demonstrating that reasoning models internalize the traversal policy that structured prompting supplies externally. Similar patterns hold on PopQA (Kimi-K2.5 pair) and IPC (Qwen3-30B-A3B pair), where the instruct model recovers only under structured prompting while its reasoning counterpart succeeds directly. Note that under direct QA, reasoning models output a short final answer, so we report their internal reasoning instead. 

\subsection{Additional Depth-Stratified Results on DBO-CAR}
\label{app:dbo_car_results}
Figure~\ref{fig:dbo_car_v3_r1} reports DeepSeek-V3 vs DeepSeek-R1 on
DBO-CAR. We observe a directionally consistent crossover: V3 leads at
shallow levels (PathMatch gap $-0.083$ at \texttt{depth}$=2$) while
R1 overtakes at the deepest level ($+0.054$ at \texttt{depth}$=6$). Final-answer accuracy saturates across both models, only
path matching score surfaces the depth-dependent divergence.

\subsection{Additional Layerwise Cosine Similarity Results}
\label{app:cos_sim_app}
\textbf{Qwen2.5 Series.} Tables~\ref{tab:layerwise_similarity} and~\ref{tab:within_model_qa_similarity} quantifies the layerwise cosine
similarity findings from Section~\ref{sec:representation_analysis} on the
\texttt{ICD9PROC} vocabulary, comparing Qwen2.5-32B (base) against its
instruct, reasoning, and distilled variants, as well as similarity between question (Q)
and answer (A) representations within each variant. We partition the 64 transformer layers
into early (1--21), middle (22--43), and late (44--64) thirds, and
report the mean similarity within each band for questions (Q) and
answers (A) separately, along with the per-component overall mean and
its minimum across layers. Across both tables, question representations diverge more than answer representations and the distilled variant diverges most,
supporting our claim that post-training reshapes question
processing rather than stored knowledge.

\begin{figure*}[h]
\centering
\includegraphics[width=\textwidth]{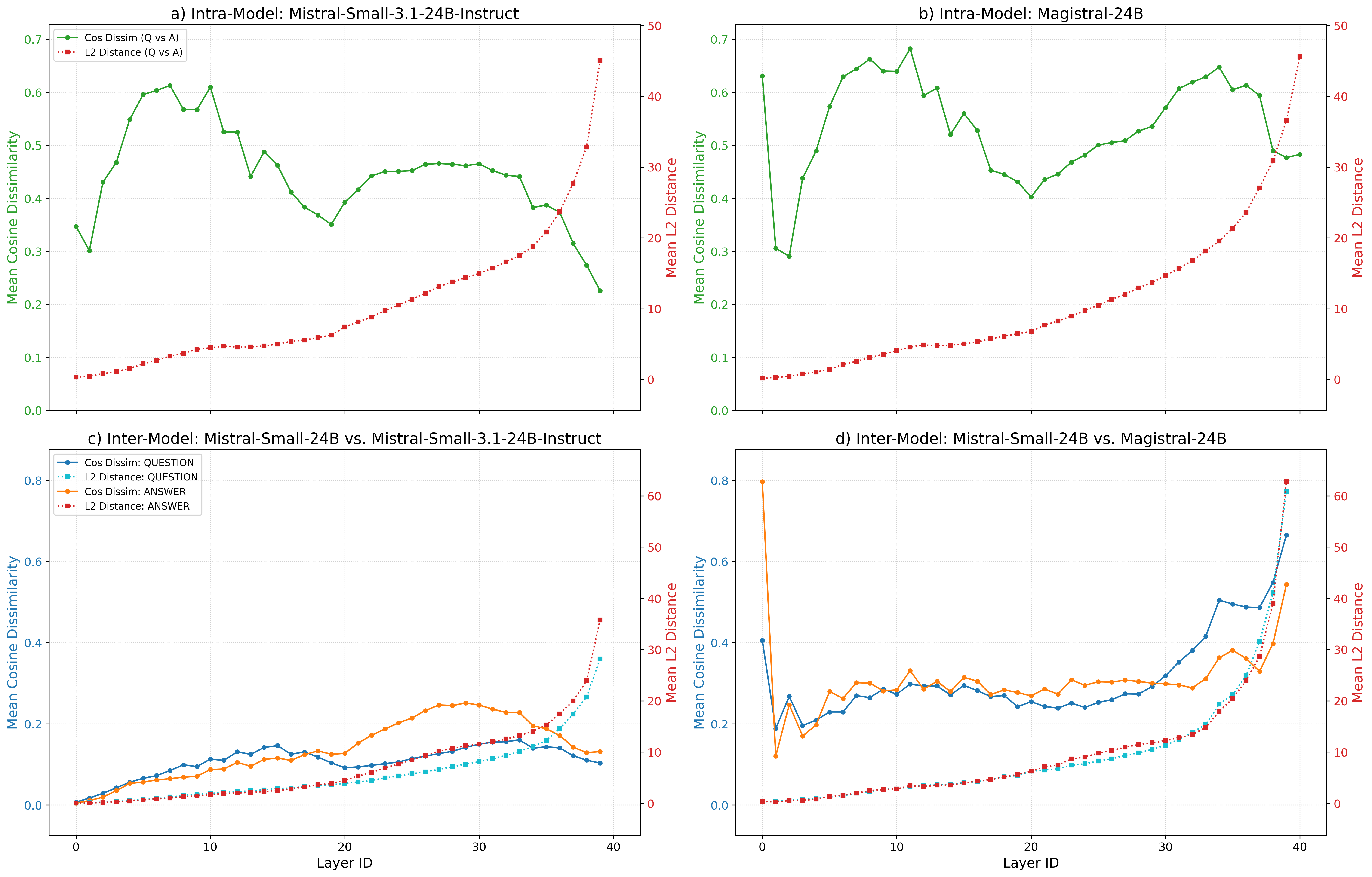}
\caption{Layerwise cosine similarity of question (Q) and answer (A) representations on ICD9PROC for Mistral-Small-24B family.}
\label{fig:appendix_divergence_mistral}
\end{figure*}

\noindent \textbf{Mistral Series.} Figure~\ref{fig:appendix_divergence_mistral} shows the corresponding analysis for the Mistral-Small model family, comparing the base (-Base-2503), instruction-tuned (-Instruct-2503), and reasoning (Magistral-Small-2507) variants across MedConceptsQA vocabulary \texttt{ICD9PROC}.

\subsection{Failure Modes of Distilled Models Under Structured Prompting}

Table~\ref{tab:qwen32b_both_datasets} reports additional results for the Qwen2.5-32B model series on MedConceptsQA, IPC, and PopQA, comparing the base, instruct, reasoning, and distilled variants across all three prompt types. Across datasets, the distilled model consistently underperforms the reasoning model and, in several cases, fails to match even the instruct or base variants, even under structured prompting. This shows the ineffectiveness of distillation on knowledge retrieval tasks.

Table~\ref{tab:distill_errors} presents representative reasoning trajectories from DeepSeek-R1-Distill-Qwen2.5-32B for each identified failure mode under structured prompting. The first category is surface-level self-correction, where the model imitates backtracking (e.g., ``Wait...'') while remaining anchored to its initial hypothesis. The second category shows unsuccessful backtracking, where the model reconsiders multiple options but converges on the wrong one. This further supports our conclusion that distillation, lacking the trial-and-error exploration inherent to RL, fails to acquire the navigation skills necessary to retrieve the parametric knowledge.

\section{Experimental Artifacts and Responsible Use}
\label{app:artifacts_responsible_use}

\subsection{Artifacts License and Intended Use}

We use several public datasets, models, and analytic tools. PopQA is released under an MIT license. MedConceptsQA is distributed on Hugging Face under an Apache license 2.0. The DBpedia Ontology is a community-curated ontology distributed under Creative Commons Attribution-ShareAlike 3.0 License (CC BY-SA 3.0). The International Patent Classification (IPC) is provided by WIPO as a public classification system for organizing and searching patent documents. We use IPC Version 2026.01 only for research evaluation, we only find its python library under GNU GPL v3 License. Infini-gram is used to estimate corpus co-occurrence frequency for entity--answer pairs, it is under Apache license 2.0. Qwen open-weight models are under Apache-2.0-style licenses, DeepSeek-R1/V3 and Kimi models are released under the MIT License; we use them only for research evaluation and do not redistribute any model weights. We cite all original artifacts in the main paper. Our use of artifacts is intended for research on language model evaluation and analysis only. 

The artifacts used and created cover several domains of factual knowledge. PopQA evaluates open-ended real world entity-relation recall; MedConceptsQA evaluates clinical code-description recall over medical ontologies; IPC-Lookup and IPC-CAR evaluate patent-code and patent-hierarchy recall; DBO-CAR evaluates hierarchy traversal over DBpedia ontology classes, including people, organizations, places, and species. All tasks are evaluated in English. The artifacts are used for studying factual recall and are not designed to represent demographic groups.

\subsection{Model Size and Computational Budget}
\label{app:compute_budget}

We evaluate models ranging from 8B to large-scale mixture-of-experts models, including Qwen2.5-32B, Qwen3-8B, Qwen3-30B-A3B, Qwen3-235B-A22B, DeepSeek-V3/V3.1/R1 with 671B parameters in total and 37B activated, and Kimi-K2/K2.5 with 1T parameters in total and 32B activated. We use around 100 GPU hours in total for running inference experiments, dataset construction, doing RL/SFT experiments. Experiments are done on serverless distributed GPU training, hence we do not have exact GPU type or hours used. 

\subsection{Use of AI assistants.}
We used AI assistants for literature search and for improving grammar and clarity. Everything reviewed and verified by the authors.

\onecolumn
\begin{figure}[H]
\begin{tcolorbox}[
    colback=pink!15!white,
    colframe=pink!80!black,
    title=\textbf{Prompt — Entity and Relation Extraction from PopQA Questions}
]
\small
\begin{verbatim}
You are an information extraction expert. Given a factual question, identify:
1. The subject ENTITY the question is about (a named entity: person, place, work, etc.).
2. The RELATION (attribute or property) being asked about the entity.

Respond ONLY with a JSON object of the form:
{"entity": "<entity string>", "relation": "<relation string>"}

Use a short, lowercase, hyphen-or-space-separated noun phrase for the relation
(e.g. "composer", "genre", "place of birth", "occupation"). Use the entity
exactly as it appears in the question (preserve casing and spelling).

###
Example 1:
Question: Who was the composer of America?
Output: {"entity": "America", "relation": "composer"}
###

###
Example 2:
Question: What genre is Holiday?
Output: {"entity": "Holiday", "relation": "genre"}
###

###
Example 3:
Question: In what city was Albert Einstein born?
Output: {"entity": "Albert Einstein", "relation": "place of birth"}
###

###
Example 4:
Question: What is the capital of France?
Output: {"entity": "France", "relation": "capital"}
###

###
Example 5:
Question: What is George Orwell's occupation?
Output: {"entity": "George Orwell", "relation": "occupation"}
###

...

Now extract the entity and relation from the following question.
Output ONLY the JSON object, no additional text.

Question: {question}
Output:
\end{verbatim}
\end{tcolorbox}
\caption{Few-shot prompt used to extract the subject entity and relation from each PopQA question before Infini-gram co-occurrence probing. }
\label{fig:popqa_extraction_prompt}
\end{figure}


\onecolumn
\begin{figure}[H]
\begin{tcolorbox}[
    colback=red!5!white,
    colframe=red!75!black,
    title=\textbf{Prompt Template 1: Final Answer Only}
]
\small
Answer the question. Output only the final answer, with no explanation.

\vspace{0.5em}
\noindent
\textit{[... Insert Question Text Here ...]}
\vspace{0.5em}

\noindent
\textbf{Answer (only the final answer):}
\end{tcolorbox}
\caption{Direct QA prompt for PopQA, IPC, and MedConceptsQA.}
\label{fig:prompt_direct}
\end{figure}

\begin{figure}[H]
\begin{tcolorbox}[
  colback  = blue!5!white,
  colframe = blue!75!black,
  title    = \textbf{Prompt Template 2: CoT with Explanation}
]
\small

You are a \textit{[medical / patent / knowledge]} research assistant.
Read the following question carefully.
Your task is to:

\begin{enumerate}
  \item \textit{MCQ only:} Answer with one of A/B/C/D, which corresponds
    to the four options. \textit{PopQA:} Provide the correct answer.
  \item For convenience, give the response in the format
    \texttt{Answer: \ldots},
    with an additional explanation in the format
    \texttt{Explanation: \ldots}.
\end{enumerate}

\vspace{0.5em}
\noindent \textit{[{\dots} Insert Question Text Here {\dots}]}
\vspace{0.5em}

Respond in the following format:

\noindent \textbf{Answer:} \\
\noindent \textbf{Explanation:}

\end{tcolorbox}
\caption{CoT prompt. The role label and answer placeholder adapt to the
  dataset: \textit{medical research assistant} (MedConceptsQA),
  \textit{patent research assistant} (IPC), or
  \textit{research assistant} (PopQA). Here MCQ datasets refer to IPC and MedConceptsQA. The structural format is otherwise
  identical.}
\label{fig:prompt_cot}
\end{figure}

\clearpage
\clearpage
\begin{tcolorbox}[
  enhanced jigsaw,                   
  breakable,
  colback  = green!5!white,
  colframe = green!75!black,
  title    = \textbf{Prompt Template 3: Structured Recall},
  pad at break*=1mm,
]
\small
\textbf{MCQ (IPC / MedConceptsQA):}
\vspace{0.3em}
You are a \textit{[medical / patent]} classification expert.
For each option, first \textbf{recall the general category and structural
breakdown of the \textit{[medical / patent]} code}, then explain
\textbf{why it might be wrong}. Finally pick the correct one.

\vspace{0.5em}
\noindent \textit{[{\dots} Insert Question Text Here {\dots}]}
\vspace{0.5em}

\noindent \textbf{Steps to follow:}
\begin{enumerate}
  \item Recall the general category and structural breakdown of the code.
  \item Evaluate each option (A--D) briefly.
  \item Choose the best option and justify.
\end{enumerate}

\noindent \textbf{Answer format:}
\begin{verbatim}
Step 1:  …
Step 2A: …
Step 2B: …
Step 2C: …
Step 2D: …
Final Answer: [A/B/C/D] because …
\end{verbatim}

\hrule
\vspace{0.8em}

\textbf{Open QA (PopQA):}
\vspace{0.3em}

You are a knowledge assistant.
Your task is to answer a factual question by first retrieving the
relevant information from memory in a structured way.

\vspace{0.5em}
\noindent \textit{[{\dots} Insert Question Text Here {\dots}]}
\vspace{0.5em}

\noindent \textbf{Steps to follow:}
\begin{enumerate}
  \item Identify the type of fact being asked about
    (e.g., date, person, place, attribute, relation, event).
  \item Recall the relevant fact or facts needed to answer the question.
  \item Provide the final answer and justify it.
\end{enumerate}

\noindent \textbf{Answer format:}
\begin{verbatim}
Step 1: …
Step 2: …
Step 3: …
Final answer: <short factual answer> because 
…
\end{verbatim}
\end{tcolorbox}
\captionof{figure}{Structured navigation prompt. For MCQ datasets (IPC,
  MedConceptsQA), the model performs hierarchical code recall followed by
  per-option elimination. For PopQA, the model identifies the fact type,
  recalls the relevant knowledge, and then states the final answer.
  Italicised labels in square brackets denote dataset-specific substitutions.}
\label{fig:prompt_structured}

\clearpage
\begin{tcolorbox}[colback=gray!5!white, colframe=gray!75!black, title=IPC-CAR Generation Prompt, breakable]
\small
\textbf{Task:} Generate IPC Common-Ancestor Dataset

\textbf{Background: IPC Code Hierarchy.} The International Patent Classification (IPC) is a tree-structured taxonomy. In the source PDF, each entry looks like:

\begin{verbatim}
1/00  Hand tools (edge trimmers for lawns A01G 3/06) [1, 2006.01]
1/02  • Spades; Shovels [1, 2006.01]
1/04  • • with teeth [1, 2006.01]
1/06  • Hoes; Hand cultivators [1, 2006.01]
1/08  • • with a single blade [1, 2006.01]
1/10  • • with two or more blades [1, 2006.01]
1/12  • • with blades provided with teeth [1, 2006.01]
1/14  • • with teeth only [1, 2006.01]
\end{verbatim}

\textbf{Parsing rules:}
\begin{itemize}[itemsep=0pt, parsep=0pt, topsep=2pt]
    \item The number of dots (\textbullet) before the description indicates the depth in the tree.
    \item A code's parent is the nearest preceding code with exactly one fewer dot.
    \item A code with zero dots (e.g., \texttt{1/00}) is a section header; its parent is the subclass (e.g., \texttt{A01B}, which is a more general sectional header immediately above it).
    \item Example: \texttt{1/04} (2 dots) $\rightarrow$ parent is \texttt{1/02} (1 dot, immediately above). \texttt{1/02} (1 dot) $\rightarrow$ parent is \texttt{1/00} (0 dots).
\end{itemize}

\textbf{Goal.} Build a multiple-choice dataset that tests an LLM's ability to recall the nearest common ancestor (NCA) of two IPC codes by traversing this hierarchy from parametric memory alone.

\textbf{Question Format.} Each question has 4 options (A--D), one correct answer, and a reasoning trace showing the two upward paths to the NCA. Example:

\begin{verbatim}
{
  "C07C_313_32_C07C_313_22_P2g": {
    "QUESTION": "Nearest common ancestor of C07C 313/32 and
                 C07C 313/22 is: A) C07C 313/08 B) C07C 313/18
                 C) C07C 313/00 D) C07C 313/12",
    "REASONING": {
      "H_PATHS": [
        "C07C 313/32 -> C07C 313/30 -> C07C 313/26 -> C07C 313/18",
        "C07C 313/22 -> C07C 313/18"
      ]
    },
    "ANSWER": "B",
    "QUESTION_TYPE": "P2g_Common_ancestor",
    "PATENT_CODE": "C07C 313/32, C07C 313/22",
    "TASK_COMPLEXITY": "MM",
    "max_levels": 3,
    "add_levels": 4
  }
}
\end{verbatim}

\textbf{Dataset Specification.}
\begin{itemize}[itemsep=0pt, parsep=0pt, topsep=2pt]
    \item \textbf{Levels:} Generate questions for traversal depths of 2, 3, 4, 5, and 6 levels, where the level equals the total number of edges traversed across both upward paths, excluding the two leaf nodes shown in the question.
    \item \textbf{Per-level count:} 20 questions per level.
\end{itemize}

\textbf{Deliverable.} Write Python code that:
\begin{enumerate}[itemsep=0pt, parsep=0pt, topsep=2pt]
    \item Parses the IPC PDF (or its text extraction) into a tree using the dot-depth rules above.
    \item For each (category, level) pair, samples leaf-code pairs whose NCA satisfies Constraint 1.
    \item Constructs 3 distractors per question.
    \item Emits the final dataset as JSON in the format shown in the example.
\end{enumerate}

\label{app:ipc_car_prompt}
\end{tcolorbox}
\captionof{figure}{The LLM prompt for generating IPC-CAR dataset.}

\onecolumn
\begin{table}[t]
\centering
\caption{Performance comparison of Instruct vs. Reasoning models on PopQA. The first column indicates the dataset. Models are evaluated across three prompt templates (QA, CoT, Structured). Metrics shown are majority voting accuracy (Maj. Vote Acc.) and mean accuracy (Mean Acc.). Mean accuracy is reported as Mean Acc. \textsubscript{\small(Std.)}, with the standard deviation in subscripted parentheses. For each model pair, a $\Delta$ row shows the gap from the reasoning model for both Maj. Acc. (orange) and Mean Acc. (yellow). Bold values indicate the best performance within each model pair. $\Delta$ values are highlighted, with darker shades indicating larger gaps.}
\label{tab:reasoning_comparison_popqa}
\scalebox{0.65}{%
\begin{tabular}{l ll | ccc | ccc}
\toprule
\multirow{2}{*}{\textbf{Dataset}} & \multirow{2}{*}{\textbf{Model}} & \multirow{2}{*}{\textbf{Model Type}} & \multicolumn{3}{c|}{\textbf{Maj. Vote Acc.}} & \multicolumn{3}{c}{\textbf{Mean Acc.\textsubscript{\small(Std.)}}} \\
\cmidrule(lr){4-6} \cmidrule(lr){7-9}
& & & \textbf{QA} & \textbf{CoT} & \textbf{Structured} & \textbf{QA} & \textbf{CoT} & \textbf{Structured} \\
\midrule
\multirow{24}{*}{\textbf{PopQA}}
& \multirow{3}{*}{\textbf{Qwen2.5-32B}}     & Instruct  & 0.252          & 0.288          & 0.310          & 0.251\textsubscript{\small(.0009)} & 0.291\textsubscript{\small(.0018)} & 0.310\textsubscript{\small(.0003)} \\
& & Reasoning & \textbf{0.386} & \textbf{0.380} & \textbf{0.375} & \textbf{0.386\textsubscript{\small(.0010)}} & \textbf{0.379\textsubscript{\small(.0014)}} & \textbf{0.378\textsubscript{\small(.0030)}} \\
& & $\Delta$  & \cellcolor{orange!62}\textbf{+0.134} & \cellcolor{orange!52}\textbf{+0.092} & \cellcolor{orange!42}\textbf{+0.065} & \cellcolor{yellow!62}\textbf{+0.135} & \cellcolor{yellow!50}\textbf{+0.088} & \cellcolor{yellow!42}\textbf{+0.068} \\
\cmidrule{2-9}
& \multirow{3}{*}{\textbf{Qwen3-8B}}        & Instruct  & 0.205          & 0.252          & \textbf{0.262}          & 0.204\textsubscript{\small(.0004)} & 0.251\textsubscript{\small(.0013)} & 0.263\textsubscript{\small(.0016)} \\
& & Reasoning & \textbf{0.237} & \textbf{0.267} & \textbf{0.262} & \textbf{0.234\textsubscript{\small(.0042)}} & \textbf{0.267\textsubscript{\small(.0016)}} & \textbf{0.265\textsubscript{\small(.0015)}} \\
& & $\Delta$  & \cellcolor{orange!18}\textbf{+0.032} & \cellcolor{orange!12}\textbf{+0.015} & \cellcolor{orange!10}\textbf{+0.000} & \cellcolor{yellow!18}\textbf{+0.029} & \cellcolor{yellow!12}\textbf{+0.016} & \cellcolor{yellow!10}\textbf{+0.002} \\
\cmidrule{2-9}
& \multirow{3}{*}{\textbf{Qwen3-30B-A3B}}   & Instruct  & 0.231          & 0.279          & 0.302          & 0.226\textsubscript{\small(.0010)} & 0.277\textsubscript{\small(.0026)} & 0.304\textsubscript{\small(.0014)} \\
& & Reasoning & \textbf{0.288} & \textbf{0.317} & \textbf{0.312} & \textbf{0.283\textsubscript{\small(.0033)}} & \textbf{0.318\textsubscript{\small(.0029)}} & \textbf{0.313\textsubscript{\small(.0011)}} \\
& & $\Delta$  & \cellcolor{orange!32}\textbf{+0.058} & \cellcolor{orange!22}\textbf{+0.037} & \cellcolor{orange!12}\textbf{+0.010} & \cellcolor{yellow!28}\textbf{+0.057} & \cellcolor{yellow!25}\textbf{+0.041} & \cellcolor{yellow!12}\textbf{+0.010} \\
\cmidrule{2-9}
& \multirow{3}{*}{\textbf{Qwen3-235B-A22B}} & Instruct  & 0.429          & 0.491          & 0.541          & 0.426\textsubscript{\small(.0005)} & 0.490\textsubscript{\small(.0010)} & 0.542\textsubscript{\small(.0007)} \\
& & Reasoning & \textbf{0.494} & \textbf{0.562} & \textbf{0.567} & \textbf{0.490\textsubscript{\small(.0020)}} & \textbf{0.566\textsubscript{\small(.0027)}} & \textbf{0.566\textsubscript{\small(.0021)}} \\
& & $\Delta$  & \cellcolor{orange!42}\textbf{+0.065} & \cellcolor{orange!50}\textbf{+0.071} & \cellcolor{orange!18}\textbf{+0.026} & \cellcolor{yellow!42}\textbf{+0.064} & \cellcolor{yellow!50}\textbf{+0.076} & \cellcolor{yellow!18}\textbf{+0.024} \\
\cmidrule{2-9}
& \multirow{3}{*}{\textbf{DeepSeek-V3}}     & Instruct  & 0.430          & 0.505          & \textbf{0.569}          & 0.422\textsubscript{\small(.0030)} & 0.504\textsubscript{\small(.0023)} & 0.569\textsubscript{\small(.0022)} \\
& & Reasoning & \textbf{0.513} & \textbf{0.557} & \textbf{0.569} & \textbf{0.502\textsubscript{\small(.0023)}} & \textbf{0.559\textsubscript{\small(.0027)}} & \textbf{0.570\textsubscript{\small(.0020)}} \\
& & $\Delta$  & \cellcolor{orange!50}\textbf{+0.083} & \cellcolor{orange!32}\textbf{+0.052} & \cellcolor{orange!10}\textbf{+0.000} & \cellcolor{yellow!50}\textbf{+0.080} & \cellcolor{yellow!32}\textbf{+0.055} & \cellcolor{yellow!10}\textbf{+0.001} \\
\cmidrule{2-9}
& \multirow{3}{*}{\textbf{DeepSeek-V3.1}}   & Instruct  & \textbf{0.429}          & 0.506          & 0.565          & \textbf{0.424}\textsubscript{\small(.0013)} & 0.506\textsubscript{\small(.0016)} & \textbf{0.566}\textsubscript{\small(.0017)} \\
& & Reasoning & \textbf{0.429} & \textbf{0.508} & \textbf{0.567} & \textbf{0.424\textsubscript{\small(.0015)}} & \textbf{0.511\textsubscript{\small(.0031)}} & 0.565\textsubscript{\small(.0019)} \\
& & $\Delta$  & \cellcolor{orange!10}\textbf{+0.003} & \cellcolor{orange!10}\textbf{+0.002} & \cellcolor{orange!10}\textbf{+0.002} & \cellcolor{yellow!10}\textbf{+0.001} & \cellcolor{yellow!10}\textbf{+0.004} & \cellcolor{yellow!10}\textbf{-0.001} \\
\cmidrule{2-9}
& \multirow{3}{*}{\textbf{Kimi-K2}}  & Instruct  & 0.507          & 0.533          & 0.566          & 0.503\textsubscript{\small(.0033)} & 0.536\textsubscript{\small(.0023)} & 0.570\textsubscript{\small(.0040)} \\
& & Reasoning & \textbf{0.580} & \textbf{0.662} & \textbf{0.652} & \textbf{0.574\textsubscript{\small(.0018)}} & \textbf{0.663\textsubscript{\small(.0007)}} & \textbf{0.653\textsubscript{\small(.0021)}} \\
& & $\Delta$  & \cellcolor{orange!38}\textbf{+0.073} & \cellcolor{orange!62}\textbf{+0.129} & \cellcolor{orange!42}\textbf{+0.086} & \cellcolor{yellow!38}\textbf{+0.071} & \cellcolor{yellow!60}\textbf{+0.127} & \cellcolor{yellow!42}\textbf{+0.083} \\
\cmidrule{2-9}
& \multirow{3}{*}{\textbf{Kimi-K2.5}}       & Instruct  & 0.507          & 0.573          & 0.649          & 0.503\textsubscript{\small(.0011)} & 0.572\textsubscript{\small(.0009)} & 0.647\textsubscript{\small(.0012)} \\
& & Reasoning & \textbf{0.597} & \textbf{0.700} & \textbf{0.682} & \textbf{0.595\textsubscript{\small(.0021)}} & \textbf{0.697\textsubscript{\small(.0026)}} & \textbf{0.685\textsubscript{\small(.0020)}} \\
& & $\Delta$  & \cellcolor{orange!50}\textbf{+0.100} & \cellcolor{orange!62}\textbf{+0.127} & \cellcolor{orange!38}\textbf{+0.033} & \cellcolor{yellow!52}\textbf{+0.107} & \cellcolor{yellow!60}\textbf{+0.125} & \cellcolor{yellow!22}\textbf{+0.038} \\
\bottomrule
\end{tabular}}
\end{table}

\begin{table}[H]
\centering
\caption{Performance comparison of Instruct vs. Reasoning models on IPC Codes.}
\label{tab:reasoning_comparison_ipc}
\scalebox{0.65}{%
\begin{tabular}{l ll | ccc | ccc}
\toprule
\multirow{2}{*}{\textbf{Dataset}} & \multirow{2}{*}{\textbf{Model}} & \multirow{2}{*}{\textbf{Model Type}} & \multicolumn{3}{c|}{\textbf{Maj. Vote Acc.}} & \multicolumn{3}{c}{\textbf{Mean Acc.\textsubscript{\small(Std.)}}} \\
\cmidrule(lr){4-6} \cmidrule(lr){7-9}
& & & \textbf{QA} & \textbf{CoT} & \textbf{Structured} & \textbf{QA} & \textbf{CoT} & \textbf{Structured} \\
\midrule
\multirow{24}{*}{\textbf{IPC Codes}}
& \multirow{3}{*}{\textbf{Qwen2.5-32B}} & Instruct & 0.300 & \textbf{0.370} & \textbf{0.400} & 0.313\textsubscript{\small(.017)} & \textbf{0.380\textsubscript{\small(.014)}} & \textbf{0.383\textsubscript{\small(.048)}} \\
& & Reasoning & \textbf{0.350} & 0.360 & 0.360 & \textbf{0.370\textsubscript{\small(.014)}} & 0.370\textsubscript{\small(.016)} & 0.353\textsubscript{\small(.017)} \\
& & $\Delta$ & \cellcolor{orange!28}\textbf{+0.050} & \cellcolor{orange!10}\textbf{-0.010} & \cellcolor{orange!10}\textbf{-0.040} & \cellcolor{yellow!32}\textbf{+0.057} & \cellcolor{yellow!10}\textbf{-0.010} & \cellcolor{yellow!10}\textbf{-0.030} \\
\cmidrule{2-9}
& \multirow{3}{*}{\textbf{Qwen3-8B}} & Instruct & 0.266 & 0.277 & 0.372 & 0.280\textsubscript{\small(.010)} & 0.284\textsubscript{\small(.005)} & 0.340\textsubscript{\small(.048)} \\
& & Reasoning & \textbf{0.298} & \textbf{0.394} & \textbf{0.394} & \textbf{0.316\textsubscript{\small(.010)}} & \textbf{0.344\textsubscript{\small(.010)}} & \textbf{0.372\textsubscript{\small(.023)}} \\
& & $\Delta$ & \cellcolor{orange!18}\textbf{+0.032} & \cellcolor{orange!50}\textbf{+0.117} & \cellcolor{orange!12}\textbf{+0.022} & \cellcolor{yellow!20}\textbf{+0.036} & \cellcolor{yellow!35}\textbf{+0.060} & \cellcolor{yellow!18}\textbf{+0.032} \\
\cmidrule{2-9}
& \multirow{3}{*}{\textbf{Qwen3-30B-A3B}} & Instruct & 0.309 & 0.319 & 0.436 & 0.312\textsubscript{\small(.005)} & 0.312\textsubscript{\small(.013)} & \textbf{0.411\textsubscript{\small(.027)}} \\
& & Reasoning & \textbf{0.426} & \textbf{0.372} & \textbf{0.457} & \textbf{0.394\textsubscript{\small(.009)}} & \textbf{0.383\textsubscript{\small(.031)}} & \textbf{0.411\textsubscript{\small(.020)}} \\
& & $\Delta$ & \cellcolor{orange!50}\textbf{+0.117} & \cellcolor{orange!28}\textbf{+0.053} & \cellcolor{orange!12}\textbf{+0.021} & \cellcolor{yellow!42}\textbf{+0.082} & \cellcolor{yellow!38}\textbf{+0.071} & \cellcolor{yellow!10}\textbf{+0.000} \\
\cmidrule{2-9}
& \multirow{3}{*}{\textbf{Qwen3-235B-A22B}} & Instruct & 0.372 & 0.383 & \textbf{0.596} & 0.351\textsubscript{\small(.015)} & 0.379\textsubscript{\small(.005)} & \textbf{0.550\textsubscript{\small(.027)}} \\
& & Reasoning & \textbf{0.457} & \textbf{0.404} & 0.574 & \textbf{0.454\textsubscript{\small(.018)}} & \textbf{0.422\textsubscript{\small(.028)}} & 0.518\textsubscript{\small(.022)} \\
& & $\Delta$ & \cellcolor{orange!42}\textbf{+0.085} & \cellcolor{orange!12}\textbf{+0.021} & \cellcolor{orange!10}\textbf{-0.022} & \cellcolor{yellow!50}\textbf{+0.103} & \cellcolor{yellow!22}\textbf{+0.043} & \cellcolor{yellow!10}\textbf{-0.032} \\
\cmidrule{2-9}
& \multirow{3}{*}{\textbf{DeepSeek-V3}} & Instruct & 0.415 & 0.596 & 0.553 & 0.415\textsubscript{\small(.009)} & 0.564\textsubscript{\small(.057)} & 0.518\textsubscript{\small(.020)} \\
& & Reasoning & \textbf{0.638} & \textbf{0.670} & \textbf{0.574} & \textbf{0.592\textsubscript{\small(.039)}} & \textbf{0.628\textsubscript{\small(.015)}} & \textbf{0.564\textsubscript{\small(.035)}} \\
& & $\Delta$ & \cellcolor{orange!50}\textbf{+0.223} & \cellcolor{orange!38}\textbf{+0.074} & \cellcolor{orange!12}\textbf{+0.021} & \cellcolor{yellow!45}\textbf{+0.177} & \cellcolor{yellow!18}\textbf{+0.064} & \cellcolor{yellow!25}\textbf{+0.046} \\
\cmidrule{2-9}
& \multirow{3}{*}{\textbf{DeepSeek-V3.1}} & Instruct & 0.468 & 0.500 & \textbf{0.606} & 0.447\textsubscript{\small(.040)} & 0.482\textsubscript{\small(.027)} & \textbf{0.592\textsubscript{\small(.027)}} \\
& & Reasoning & \textbf{0.479} & \textbf{0.511} & 0.585 & \textbf{0.454\textsubscript{\small(.013)}} & \textbf{0.518\textsubscript{\small(.013)}} & 0.553\textsubscript{\small(.017)} \\
& & $\Delta$ & \cellcolor{orange!12}\textbf{+0.011} & \cellcolor{orange!12}\textbf{+0.011} & \cellcolor{orange!10}\textbf{-0.021} & \cellcolor{yellow!12}\textbf{+0.007} & \cellcolor{yellow!22}\textbf{+0.036} & \cellcolor{yellow!10}\textbf{-0.039} \\
\cmidrule{2-9}
& \multirow{3}{*}{\textbf{Kimi-K2}} & Instruct & 0.511 & 0.521 & 0.543 & 0.511\textsubscript{\small(.000)} & 0.514\textsubscript{\small(.010)} & \textbf{0.560\textsubscript{\small(.028)}} \\
& & Reasoning & \textbf{0.575} & \textbf{0.585} & \textbf{0.564} & \textbf{0.543\textsubscript{\small(.009)}} & \textbf{0.564\textsubscript{\small(.009)}} & 0.525\textsubscript{\small(.013)} \\
& & $\Delta$ & \cellcolor{orange!32}\textbf{+0.064} & \cellcolor{orange!32}\textbf{+0.064} & \cellcolor{orange!12}\textbf{+0.021} & \cellcolor{yellow!18}\textbf{+0.032} & \cellcolor{yellow!28}\textbf{+0.050} & \cellcolor{yellow!10}\textbf{-0.035} \\
\cmidrule{2-9}
& \multirow{3}{*}{\textbf{Kimi-K2.5}} & Instruct & 0.521 & 0.500 & \textbf{0.553} & 0.482\textsubscript{\small(.010)} & 0.514\textsubscript{\small(.033)} & 0.543\textsubscript{\small(.031)} \\
& & Reasoning & \textbf{0.574} & \textbf{0.574} & \textbf{0.585} & \textbf{0.557\textsubscript{\small(.018)}} & \textbf{0.571\textsubscript{\small(.018)}} & \textbf{0.553}\textsubscript{\small(.035)} \\
& & $\Delta$ & \cellcolor{orange!28}\textbf{+0.053} & \cellcolor{orange!38}\textbf{+0.074} & \cellcolor{orange!10}\textbf{+0.032} & \cellcolor{yellow!38}\textbf{+0.075} & \cellcolor{yellow!32}\textbf{+0.057} & \cellcolor{yellow!10}\textbf{+0.010} \\
\bottomrule
\end{tabular}}
\end{table}

\begin{table}[H]
\centering
\caption{Performance comparison of Instruct vs. Reasoning models on MedConceptsQA.}
\label{tab:reasoning_comparison_medconceptsqa}
\scalebox{0.65}{%
\begin{tabular}{l ll | ccc | ccc}
\toprule
\multirow{2}{*}{\textbf{Dataset}} & \multirow{2}{*}{\textbf{Model}} & \multirow{2}{*}{\textbf{Model Type}} & \multicolumn{3}{c|}{\textbf{Maj. Vote Acc.}} & \multicolumn{3}{c}{\textbf{Mean Acc.\textsubscript{\small(Std.)}}} \\
\cmidrule(lr){4-6} \cmidrule(lr){7-9}
& & & \textbf{QA} & \textbf{CoT} & \textbf{Structured} & \textbf{QA} & \textbf{CoT} & \textbf{Structured} \\
\midrule
\multirow{24}{*}{\textbf{MedConceptsQA}} & \multirow{3}{*}{\textbf{Qwen2.5-32B}} & Instruct & 0.379 & 0.475 & 0.469 & 0.371\textsubscript{\small(.012)} & 0.449\textsubscript{\small(.010)} & 0.454\textsubscript{\small(.007)} \\
& & Reasoning & \textbf{0.482} & \textbf{0.513} & \textbf{0.505} & \textbf{0.470\textsubscript{\small(.012)}} & \textbf{0.487\textsubscript{\small(.009)}} & \textbf{0.481\textsubscript{\small(.005)}} \\
& & $\Delta$ & \cellcolor{orange!46}\textbf{+0.103} & \cellcolor{orange!34}\textbf{+0.038} & \cellcolor{orange!32}\textbf{+0.036} & \cellcolor{yellow!43}\textbf{+0.099} & \cellcolor{yellow!35}\textbf{+0.038} & \cellcolor{yellow!25}\textbf{+0.027} \\
\cmidrule{2-9}
& \multirow{3}{*}{\textbf{Qwen3-8B}} & Instruct & 0.338 & 0.337 & 0.406 & 0.339\textsubscript{\small(.007)} & 0.335\textsubscript{\small(.005)} & 0.386\textsubscript{\small(.011)} \\
& & Reasoning & \textbf{0.400} & \textbf{0.424} & \textbf{0.421} & \textbf{0.386\textsubscript{\small(.012)}} & \textbf{0.410\textsubscript{\small(.003)}} & \textbf{0.407\textsubscript{\small(.009)}} \\
& & $\Delta$ & \cellcolor{orange!35}\textbf{+0.062} & \cellcolor{orange!42}\textbf{+0.087} & \cellcolor{orange!22}\textbf{+0.015} & \cellcolor{yellow!32}\textbf{+0.047} & \cellcolor{yellow!38}\textbf{+0.075} & \cellcolor{yellow!22}\textbf{+0.021} \\
\cmidrule{2-9}
 & \multirow{3}{*}{\textbf{Qwen3-30B-A3B}} & Instruct & 0.398 & 0.404 & \textbf{0.489} & 0.399\textsubscript{\small(.005)} & 0.399\textsubscript{\small(.005)} & \textbf{0.485\textsubscript{\small(.006)}} \\
& & Reasoning & \textbf{0.498} & \textbf{0.506} & 0.476 & \textbf{0.494\textsubscript{\small(.007)}} & \textbf{0.501\textsubscript{\small(.014)}} & 0.477\textsubscript{\small(.008)} \\
& & $\Delta$ & \cellcolor{orange!50}\textbf{+0.100} & \cellcolor{orange!51}\textbf{+0.102} & \cellcolor{orange!10}\textbf{-0.013} & \cellcolor{yellow!47}\textbf{+0.095} & \cellcolor{yellow!51}\textbf{+0.102} & \cellcolor{yellow!10}\textbf{-0.008} \\
\cmidrule{2-9}
& \multirow{3}{*}{\textbf{Qwen3-235B-A22B}} & Instruct & 0.542 & 0.548 & \textbf{0.631} & 0.503\textsubscript{\small(.004)} & 0.528\textsubscript{\small(.005)} & \textbf{0.589\textsubscript{\small(.007)}} \\
& & Reasoning & \textbf{0.641} & \textbf{0.656} & 0.580 & \textbf{0.599\textsubscript{\small(.003)}} & \textbf{0.617\textsubscript{\small(.003)}} & 0.554\textsubscript{\small(.008)} \\
& & $\Delta$ & \cellcolor{orange!44}\textbf{+0.099} & \cellcolor{orange!46}\textbf{+0.108} & \cellcolor{orange!10}\textbf{-0.051} & \cellcolor{yellow!42}\textbf{+0.096} & \cellcolor{yellow!40}\textbf{+0.089} & \cellcolor{yellow!10}\textbf{-0.035} \\
\cmidrule{2-9}
& \multirow{3}{*}{\textbf{DeepSeek-V3}} & Instruct & 0.541 & 0.632 & 0.717 & 0.551\textsubscript{\small(.014)} & 0.636\textsubscript{\small(.049)} & 0.701\textsubscript{\small(.026)} \\
& & Reasoning & \textbf{0.778} & \textbf{0.790} & \textbf{0.792} & \textbf{0.830\textsubscript{\small(.006)}} & \textbf{0.774\textsubscript{\small(.013)}} & \textbf{0.775\textsubscript{\small(.026)}} \\
& & $\Delta$ & \cellcolor{orange!75}\textbf{+0.237} & \cellcolor{orange!57}\textbf{+0.158} & \cellcolor{orange!38}\textbf{+0.075} & \cellcolor{yellow!80}\textbf{+0.279} & \cellcolor{yellow!55}\textbf{+0.138} & \cellcolor{yellow!35}\textbf{+0.074} \\
\cmidrule{2-9}
& \multirow{3}{*}{\textbf{DeepSeek-V3.1}} & Instruct & 0.526 & 0.595 & \textbf{0.785} & 0.520\textsubscript{\small(.008)} & 0.593\textsubscript{\small(.009)} & 0.763\textsubscript{\small(.004)} \\
& & Reasoning
& \textbf{0.794}
& \textbf{0.805}
& 0.774
& \textbf{0.785}\textsubscript{\small(.011)}
& \textbf{0.772}\textsubscript{\small(.008)}
& \textbf{0.772}\textsubscript{\small(.008)} \\
& & $\Delta$ & \cellcolor{orange!85}\textbf{+0.268} & \cellcolor{orange!67}\textbf{+0.210} & \cellcolor{orange!10}\textbf{-0.011} & \cellcolor{yellow!85}\textbf{+0.265} & \cellcolor{yellow!57}\textbf{+0.179} & \cellcolor{yellow!10}\textbf{+0.009} \\ 
\cmidrule{2-9}
& \multirow{3}{*}{\textbf{Kimi-K2}} & Instruct & 0.595 & 0.634 & 0.750 & 0.593\textsubscript{\small(.003)} & 0.625\textsubscript{\small(.003)} & 0.735\textsubscript{\small(.011)} \\
& & Reasoning & \textbf{0.834} & \textbf{0.829} & \textbf{0.830} & \textbf{0.809\textsubscript{\small(.001)}} & \textbf{0.810\textsubscript{\small(.006)}} & \textbf{0.799\textsubscript{\small(.007)}} \\
& & $\Delta$ & \cellcolor{orange!75}\textbf{+0.239} & \cellcolor{orange!69}\textbf{+0.195} & \cellcolor{orange!40}\textbf{+0.080} & \cellcolor{yellow!72}\textbf{+0.216} & \cellcolor{yellow!65}\textbf{+0.185} & \cellcolor{yellow!32}\textbf{+0.064} \\
\cmidrule{2-9}
& \multirow{3}{*}{\textbf{Kimi-K2.5}} & Instruct & 0.699 & 0.734 & 0.875 & 0.690\textsubscript{\small(.006)} & 0.727\textsubscript{\small(.008)} & 0.864\textsubscript{\small(.008)} \\
& & Reasoning & \textbf{0.905} & \textbf{0.915} & \textbf{0.911} & \textbf{0.881\textsubscript{\small(.008)}} & \textbf{0.895\textsubscript{\small(.008)}} & \textbf{0.895\textsubscript{\small(.008)}} \\
& & $\Delta$ & \cellcolor{orange!72}\textbf{+0.206} & \cellcolor{orange!65}\textbf{+0.181} & \cellcolor{orange!28}\textbf{+0.036} & \cellcolor{yellow!68}\textbf{+0.191} & \cellcolor{yellow!60}\textbf{+0.168} & \cellcolor{yellow!25}\textbf{+0.031} \\
\bottomrule
\end{tabular}}
\end{table}

\onecolumn

\begin{figure}[H]
\centering
\begin{minipage}{\linewidth}
\caption{Reasoning trajectories for Kimi-K2.5 (non-thinking) and its reasoning variant across prompt templates on PopQA.}
\label{fig:reasoning_traces_kimi_popqa}

\begin{tcolorbox}[colback=orange!5!white, colframe=orange!90!black, title=\textbf{Question}]
\texttt{What is the capital of Poni Province?}\\[4pt]
\texttt{\textbf{Ground Truth: Gaoua} \textcolor{green!60!black}{$\checkmark$}}
\end{tcolorbox}

\begin{tcolorbox}[colback=red!5!white,
    colframe=red!75!black, title=\textbf{Kimi-K2.5-NonThink --- Direct QA}]
Boulounga \textcolor{red}{$\times$}
\end{tcolorbox}

\begin{tcolorbox}[colback=blue!5!white,
    colframe=blue!75!black, title=\textbf{Kimi-K2.5-NonThink --- CoT}]
\textbf{Answer:} Kedougou

\textbf{Explanation:}  The capital city of Poni Province is
Kedougou (also spelled Kédougou). The province shares its name with the Poni River
and is known for its agricultural activities and natural landscapes. \textcolor{red}{$\times$}
\end{tcolorbox}

\begin{tcolorbox}[colback=green!5!white,
    colframe=green!75!black, title=\textbf{Kimi-K2.5-NonThink --- Structured Prompting}]
\textcolor{red}{\textbf{Step 1:} This is a geography question asking about the capital of a place (Poni Province).\\[4pt]
\textbf{Step 2:} Poni Province is a province in Burkina Faso, a country in West Africa. The capital of Poni Province is Gaoua.\\[4pt]
\textbf{Step 3:} Gaoua serves as the administrative center of Poni Province in southwestern Burkina Faso.}\\[6pt]
\textbf{Final Answer: Gaoua} \textcolor{green!60!black}{$\checkmark$}
\end{tcolorbox}

\begin{tcolorbox}[colback=red!5!white,
    colframe=red!75!black, title=\textbf{Kimi-K2.5-Think --- Direct QA}]
\textcolor{red}{\textbf{Reasoning:} Poni Province is one of the 45 provinces of Burkina Faso,
located in the Sud-Ouest Region. The capital is indeed Gaoua.}\\[6pt]
\textbf{Final Answer: Gaoua} \textcolor{green!60!black}{$\checkmark$}
\end{tcolorbox}
\end{minipage}
\end{figure}

\begin{figure}[H]
\centering
\scalebox{0.84}{
\begin{minipage}{\linewidth}
\caption{Reasoning trajectories for Qwen3-30B-A3B (non-thinking) and its reasoning variant across prompt templates on IPC.}
\label{fig:reasoning_traces_qwen3_30b_ipc_b01f}
\begin{tcolorbox}[colback=orange!5!white, colframe=orange!90!black, title=\textbf{Question}]
\texttt{What does the patent code B01F 27/1145 in the International Patent Classification describe?}\\[4pt]
\texttt{A. Centrifugal stirrers, i.e.\ stirrers with radial outlets; Stirrers of the turbine type, e.g.\ with means to guide the flow}\\
\texttt{B. Helically shaped stirrers, ribbon shaped with an open space between the helical ribbon flight and the rotating axis}\\
\texttt{C. Propeller-shaped stirrers for producing an axial flow, e.g.\ shaped like a ship or aircraft propeller}\\
\texttt{D. Stirrers shaped as cylinders, balls or rollers}\\[4pt]
\texttt{\textbf{Ground Truth: B} \textcolor{green!60!black}{$\checkmark$}}
\end{tcolorbox}

\begin{tcolorbox}[colback=red!5!white,
    colframe=red!75!black, title=\textbf{Qwen3-30B-A3B-NonThink --- Direct QA}]
A \textcolor{red}{$\times$}
\end{tcolorbox}

\begin{tcolorbox}[colback=blue!5!white,
    colframe=blue!75!black, title=\textbf{Qwen3-30B-A3B-NonThink --- CoT}]
\textbf{Answer:} A\\[4pt]
\textbf{Explanation:} The patent code B01F 27/1145 refers to centrifugal stirrers
with radial outlets and stirrers of the turbine type, which may include means to
guide the flow. This classification covers stirrers that operate by creating a
radial flow, commonly seen in industrial mixing applications. \textcolor{red}{$\times$}
\end{tcolorbox}

\begin{tcolorbox}[colback=green!5!white,
    colframe=green!75!black, title=\textbf{Qwen3-30B-A3B-NonThink --- Structured Prompting}]
\textcolor{red}{\textbf{Step 1: B01 covers physical or chemical processes or apparatus; B01F covers mixing; B01F 27/1145 denotes helically shaped or ribbon shaped stirrers with an open space between the helical ribbon flight and the rotating axis.}}\\[4pt]
\textbf{Step 2A:} Option A (centrifugal/turbine-type stirrers with radial outlets)
falls under different subclasses such as
B01F 27/111. \textbf{Incorrect.}\\[4pt]
\textbf{Step 2B:} Option B (helically shaped, ribbon-type stirrers with open space
between ribbon flight and axis) is a precise match for B01F 27/1145.
\textbf{Correct.}\\[4pt]
\textbf{Step 2C:} Option C (propeller-shaped stirrers producing axial flow) is
typically classified under axial-flow impeller subclasses, not B01F 27/1145.
\textbf{Incorrect.}\\[4pt]
\textbf{Step 2D:} Option D (stirrers shaped as cylinders, balls, or rollers) does
not correspond to helical ribbon stirrers and is covered by code such as B01F 27/116.
\textbf{Incorrect.}\\[6pt]
\textbf{Final Answer: B} \textcolor{green!60!black}{$\checkmark$}
\end{tcolorbox}

\begin{tcolorbox}[colback=red!5!white,
    colframe=red!75!black, title=\textbf{Qwen3-30B-A3B-Think --- Direct QA}]
\textbf{Reasoning:} \textcolor{red}{\textbf{B01F relates to mixing apparatus.  B01F 27 refers to mixers with rotary devices. The subgroup 27/1145 specifically covers helical ribbon
stirrers---ribbon-shaped with an open space between the ribbon flight and the
rotating axis.}} Options A (centrifugal/turbine), C (propeller/axial), and D
(cylinders/balls/rollers) each describe distinct stirrer types classified
elsewhere. Option B is therefore the correct match.\\[6pt]
\textbf{Final Answer: B} \textcolor{green!60!black}{$\checkmark$}
\end{tcolorbox}
\end{minipage}
}
\end{figure}

\begin{figure}[H]
\centering
\begin{minipage}{\linewidth} 
\caption{Reasoning trajectories for Qwen3-235B-A22B-2507-Instruct and its reasoning model across prompt templates on MedConceptsQA.}
\label{fig:reasoning_traces_qwen3}

\begin{tcolorbox}[colback=orange!5!white, colframe=orange!90!black, title=\textbf{Question}]
\small
\texttt{What is the description of the medical code 66.97 in ICD9PROC?}\\[4pt]
\texttt{A. Bilateral partial
salpingectomy, not otherwise specified}\\
\texttt{B. Salpingotomy and salpingostomy}\\
\texttt{C. Other operations on
female genital organs } \\
\texttt{\textbf{D. Burying of fimbriae in uterine wall} \textcolor{green!60!black} {$\checkmark$}}
\end{tcolorbox}

\begin{tcolorbox}[colback=red!5!white,
    colframe=red!75!black, title=\textbf{Qwen3-235B-A22B-2507-Instruct --- Direct QA}]
\small
The code \textbf{66.97} in \textbf{ICD-9-PROC} stands for ``Other operations on female genital
organs." This code is used when a procedure on the female genital organs is performed
that is not specifically listed elsewhere in the coding system. Therefore, the correct answer is \[
\boxed{C}
\]
\end{tcolorbox}

\begin{tcolorbox}[colback=blue!5!white,
    colframe=blue!75!black, title=\textbf{Qwen3-235B-A22B-2507-Instruct --- CoT}]
\small
\textbf{Answer:} C

\textbf{Explanation:} \textbf{ICD-9-CM} procedure code \textbf{66.97} falls under the category ``Other operations on female genital organs."  It's a residual category for procedures not otherwise specified in the
coding system. This makes option C the accurate description. \\[6pt]
\textbf{Option Analysis:}\\[2pt]
The other options are incorrect: 
option A mentions bilateral partial salpingectomy (e.g., 66.6x) which would be a different code range. B refers to salpingotomy and salpingostomy procedures (66.01, 66.02), which have their own specific codes. Option D describes a very specific technique of burying fimbriae in the uterine wall, which
is not a standard procedure or code listed under 66.97.
\end{tcolorbox}

\begin{tcolorbox}[colback=green!5!white,
    colframe=green!75!black, title=\textbf{Qwen3-235B-A22B-2507-Instruct --- Structured Prompting}]
\small
\textcolor{red} {\textbf{The medical code 66.97 belongs to the ICD-9-CM (International Classification
of Diseases, 9th Revision, Clinical Modification) procedural coding system. It falls
under the broad category of ``Operations on the female genital organs" (ICD-9-CM codes
65–69). Chapter 66 refers to ``Other operations on the fallopian tubes and broad ligament.", procedure code 66.97 is "Burying of fimbriae in uterine wall". }} \\[6pt]
\textbf{Option Analysis:}\\[2pt]
\textbf{A. Bilateral partial salpingectomy, not otherwise specified} --- incorrect, typically coded under 66.6x (e.g., 66.62 for salpingectomy related procedures) \\[2pt]
\textbf{B. Salpingotomy and salpingostomy} --- incorrect, specific procedures like salpingotomy (an incision into the fallopian tube, often for ectopic pregnancy) usually
coded as 66.01 and 66.02. \\[2pt]
\textbf{C. Other operations on female genital organs} --- incorrect, very broad category and refers to codes in the 69.x range. \\[2pt]
\textbf{D. Burying of fimbriae in uterine wall} --- correct, classified under ICD-9-CM code 66.97.\\[6pt]
\textbf{Final Answer: D} 
\end{tcolorbox}

\begin{tcolorbox}[colback=red!5!white,
    colframe=red!75!black,  title=\textbf{Qwen3-235B-A22B-2507-Think --- Direct QA}]
\small
\textcolor{red}{\textbf{Code Structure: In \textbf{ICD-9-CM}, codes in the 65-71 range are for operations on the female genital organs. Within this chapter, code 66 is for ``Operations on fallopian tubes and broad ligaments." and 66.9 is ``Other operations on fallopian tubes." 66.97 is "Burying of fimbriae in uterine wall." }}\\[6pt]
\textbf{Option Analysis:}\\[2pt]
\textbf{A. Bilateral partial salpingectomy, not otherwise specified} --- incorrect, might fall under 66.62.\\[2pt]
\textbf{B. Salpingotomy and salpingostomy} --- incorrect. Salpingotomy is often 66.02 or included in 66.0, but not specifically 66.97. \\[2pt]
\textbf{C. Other operations on fallopian tubes} --- incorrect, C is a distractor because ``other operations" might be misinterpreted, but
66.97 is specific. Code 66.9 is ``Other operations on fallopian tubes," and 66.97 is a specific type
under that. \\[2pt]
\textbf{Final Answer: D} --- code \textbf{66.97} is ``Burying of fimbriae in uterine wall", where the fimbriated end of the fallopian tube is
buried into the uterine wall to prevent fluid accumulation or for sterilization.
\end{tcolorbox}

\end{minipage} 
\end{figure}

\begin{figure}[H]
\centering
\begin{minipage}{\linewidth} 
\caption{Reasoning trajectories for DeepSeek-V3 and DeepSeek-R1 across prompt templates on MedConceptsQA.}
\label{fig:reasoning_traces}

\begin{tcolorbox}[colback=orange!5!white, colframe=orange!90!black, title=\textbf{Question}]
\small
\texttt{What is the description of the medical code 57.95 in ICD9PROC?}\\[4pt]
\texttt{A. Other procedures for relief of elevated intraocular pressure}\\
\texttt{\textbf{B. Replacement of indwelling urinary catheter} \textcolor{green!60!black}{$\checkmark$}}\\
\texttt{C. Arthrodesis of other joint}\\
\texttt{D. Biopsy of heart}
\end{tcolorbox}

\begin{tcolorbox}[colback=red!5!white,
    colframe=red!75!black, title=\textbf{DeepSeek-V3 --- Direct QA}]
\small
A
\end{tcolorbox}

\begin{tcolorbox}[colback=blue!5!white,
    colframe=blue!75!black, title=\textbf{DeepSeek-V3 --- CoT}]
\small
\textbf{Answer:} A   
\textbf{Explanation:} The medical code \textbf{57.95} in \textbf{IPC-9-PROC} corresponds to ``other procedures for relief of elevated intraocular pressure." This code is used to classify surgical or procedural interventions aimed at reducing high pressure within the eye, such as in glaucoma treatment. The other options (b, c, d) are unrelated to this specific procedure.
\end{tcolorbox}

\begin{tcolorbox}[colback=green!5!white,
    colframe=green!75!black, title=\textbf{DeepSeek-V3 --- Structured Prompting}]
\small
Here is the breakdown for \textbf{IPC-9-PROC procedure code 57.95}:\\[4pt]
\textcolor{red} {\textbf{Code Structure: Category \textbf{57} refers to ``Other operations on the bladder'' in ICD-9-PROC Volume 3. Subcode \textbf{.95} indicates ``Replacement of indwelling urinary catheter.''}} \\[6pt]
\textbf{Option Analysis:}\\[2pt]
\textbf{A. Other procedures for relief of elevated intraocular pressure} --- incorrect. This falls under ophthalmology (codes in the \textbf{12--14} range), not bladder procedures.\\[2pt]
\textbf{B. Replacement of indwelling urinary catheter} --- correct. Matches the \textbf{57.95} code description for bladder-related catheter replacement.\\[2pt]
\textbf{C. Arthrodesis of other joint} --- incorrect. Arthrodesis is a musculoskeletal procedure (codes in the \textbf{81} range).\\[2pt]
\textbf{D. Biopsy of heart} --- incorrect. Heart biopsies are cardiovascular procedures (codes in the \textbf{37} range).\\[6pt]
\textbf{Final Answer: B} --- code \textbf{57.95} specifically refers to the replacement of an indwelling urinary catheter, a bladder-related procedure.
\end{tcolorbox}

\begin{tcolorbox}[colback=red!5!white,
    colframe=red!75!black,  title=\textbf{DeepSeek-R1 --- Direct QA}]
\small
\textcolor{red}{\textbf{Code Structure: Code \textbf{57.95} is an ICD-9-PROC procedure code. In ICD-9-PROC Volume 3, Chapter 11 covers procedures on the urinary system (codes 55--59). Within this chapter, category \textbf{57} and subcode \textbf{57.9} refer to ``Other operations on bladder,'' and subcode \textbf{57.95} is ``Replacement of indwelling urinary catheter.''}}\\[6pt]
\textbf{Option Analysis:}\\[2pt]
\textbf{A. Other procedures for relief of elevated intraocular pressure} --- incorrect. This is an ophthalmologic procedure falling under a different system entirely.\\[2pt]
\textbf{B. Replacement of indwelling urinary catheter} --- correct. Directly matches \textbf{57.95} under the bladder operations category.\\[2pt]
\textbf{C. Arthrodesis of other joint} --- incorrect. Arthrodesis is a musculoskeletal procedure (codes in the \textbf{80s} range).\\[2pt]
\textbf{D. Biopsy of heart} --- incorrect. Heart biopsies are cardiovascular procedures (codes in the \textbf{35--39} range).\\[6pt]
\textbf{Final Answer: B} --- code \textbf{57.95} specifically refers to the replacement of an indwelling urinary catheter, a bladder-related procedure.
\end{tcolorbox}

\end{minipage} 
\end{figure}


\begin{table}[t]
\centering
\caption{Layerwise mean cosine similarity between Qwen2.5-32B (base) and
its post-trained variants on \texttt{ICD9PROC} question (Q) and answer
(A) representations, averaged over $N{=}100$ probes. Mean cosine similarity is reported for early (1--21), middle (22--43), and late (44--64) thirds. Lower similarity indicates greater divergence from the base model. Across all
three variants, question representations diverge more than answer
representations, and the distilled model diverges most.}
\label{tab:layerwise_similarity}
\small
\begin{tabular}{llcccccc}
\toprule
Variant & Comp.\ & Early & Middle & Late & Overall & Min & Min Layer\\
\midrule
\multirow{2}{*}{Instruct}
        & Q & 0.990 & 0.970 & 0.938 & 0.966 & 0.729 & 64 \\
        & A & 0.994 & 0.976 & 0.958 & 0.976 & 0.871 & 64 \\
\midrule
\multirow{2}{*}{Reasoning}
        & Q & 0.982 & 0.942 & 0.923 & 0.950 & 0.810 & 64 \\
        & A & 0.987 & 0.951 & 0.944 & 0.961 & 0.876 & 64 \\
\midrule
\multirow{2}{*}{Distilled}
        & Q & 0.911 & 0.718 & 0.789 & 0.808 & 0.634 & 36 \\
        & A & 0.955 & 0.864 & 0.854 & 0.892 & 0.778 & 64 \\
\bottomrule
\end{tabular}
\end{table}

\begin{table}[t]
\centering
\caption{Within-model layerwise cosine similarity between question (Q)
and answer (A) representations on \texttt{ICD9PROC}, averaged over
$N{=}100$ probes. Higher similarity indicates that questions and answers
share representational structure at that layer. Across all models, Q
and A representations diverge substantially after the embedding layer
and the distilled model's late layers show the largest divergence.}
\label{tab:within_model_qa_similarity}
\small
\begin{tabular}{lcccccc}
\toprule
Variant & Early (1--21) & Middle (22--43) & Late (44--64) & Overall & Min & Min Layer\\
\midrule
Instruct      & 0.276 & 0.393 & 0.325 & 0.333 & 0.148 & 63 \\
Reasoning           & 0.285 & 0.410 & 0.310 & 0.336 & 0.161 & 63 \\
Distilled      & 0.317 & 0.363 & 0.203 & 0.295 & 0.091 & 63 \\
\bottomrule
\end{tabular}
\end{table}

\begin{table}[t]
\centering
\caption{Qwen2.5-32B model series accuracy on MedConceptsQA, IPC and PopQA. Mean Acc. shown as mean\textsubscript{\small(std)}.}
\label{tab:qwen32b_both_datasets}
\scriptsize
\begin{tabular}{ll ccc ccc}
\toprule
& & \multicolumn{3}{c}{\textbf{Maj. Vote Acc.}} & \multicolumn{3}{c}{\textbf{Mean Acc.\textsubscript{\small(Std.)}}} \\
\cmidrule(lr){3-5} \cmidrule(lr){6-8}
\textbf{Dataset} & \textbf{Model} & \textbf{QA} & \textbf{CoT} & \textbf{Structured} & \textbf{QA} & \textbf{CoT} & \textbf{Structured} \\
\midrule
\multirow{5}{*}{\textbf{MedConceptsQA}}
& Base      & 0.221 & 0.332 & 0.404 & 0.219\textsubscript{\small(.012)} & 0.260\textsubscript{\small(.071)} & 0.372\textsubscript{\small(.007)} \\
& Distilled & 0.375 & 0.380 & 0.447 & 0.351\textsubscript{\small(.009)} & 0.369\textsubscript{\small(.005)} & 0.420\textsubscript{\small(.002)} \\
& Instruct  & 0.379 & 0.475 & 0.469 & 0.371\textsubscript{\small(.012)} & 0.449\textsubscript{\small(.010)} & 0.454\textsubscript{\small(.007)} \\
& Reasoning & \textbf{0.482} & \textbf{0.513} & \textbf{0.505} & \textbf{0.470}\textsubscript{\small(.012)} & \textbf{0.487}\textsubscript{\small(.009)} & \textbf{0.481}\textsubscript{\small(.005)} \\
\midrule
\multirow{5}{*}{\textbf{IPC Codes}}
& Base      & \textbf{0.350} & 0.320 & 0.390 & 0.310\textsubscript{\small(.065)} & 0.270\textsubscript{\small(.008)} & 0.370\textsubscript{\small(.071)} \\
& Distilled & 0.340 & 0.340 & 0.330 & 0.369\textsubscript{\small(.050)} & 0.350\textsubscript{\small(.045)} & 0.330\textsubscript{\small(.014)} \\
& Instruct  & 0.300 & \textbf{0.370} & \textbf{0.400} & 0.313\textsubscript{\small(.017)} & \textbf{0.380}\textsubscript{\small(.014)} & \textbf{0.383}\textsubscript{\small(.048)} \\
& Reasoning & \textbf{0.350} & 0.360 & 0.360 & \textbf{0.370}\textsubscript{\small(.014)} & 0.370\textsubscript{\small(.016)} & 0.353\textsubscript{\small(.017)} \\
\midrule
\multirow{5}{*}{\textbf{PopQA}}
& Base      & 0.281 & 0.277 & 0.299 & 0.280\textsubscript{\small(.007)} & 0.279\textsubscript{\small(.002)} & 0.303\textsubscript{\small(.003)} \\
& Distilled & 0.332 & 0.334 & 0.328 & 0.334\textsubscript{\small(.021)} & 0.334\textsubscript{\small(.017)} & 0.325\textsubscript{\small(.022)} \\
& Instruct  & 0.302 & 0.289 & 0.310 & 0.297\textsubscript{\small(.003)} & 0.291\textsubscript{\small(.002)} & 0.310\textsubscript{\small(.000)} \\
& Reasoning & \textbf{0.386} & \textbf{0.380} & \textbf{0.375} & \textbf{0.386}\textsubscript{\small(.010)} & \textbf{0.379}\textsubscript{\small(.014)} & \textbf{0.378}\textsubscript{\small(.030)} \\
\bottomrule
\end{tabular}
\end{table}
\label{sec: distill_failure}

\onecolumn
\begin{table}[t]
\centering
\caption{Failure mode categorization from 40 incorrect responses of \texttt{DeepSeek-R1-Distill-Qwen2.5-32B} under structured prompting.}
\label{tab:distill_errors}
\small
\begin{tabular}{p{0.3cm} p{3.8cm} p{7.5cm} p{1.5cm}}
\toprule
\textbf{\#} & \textbf{Failure Mode Description} & \textbf{Redacted Reasoning Trajectory} & \textbf{Rate (\%)} \\
\midrule
1 & Surface-level self-correction: mimics self-correction patterns using phrases such as ``Wait\ldots'' or ``Alternatively\ldots'', but remains trapped in the initial wrong hypothesis. It performs pattern-matching against its first committed option to eliminate other options rather than genuinely revisiting alternative hypotheses. & The \textbf{ICD-9-PROC} code \textbf{00.09} falls under the ``Therapeutic Ultrasound'', a part of the ``Radiology'' section. This code refers to other types of therapeutic ultrasound not classified elsewhere. Option A relates to gastrointestinal procedures, specifically endoscopic treatments. \textbf{\textcolor{red}{It does not involve ultrasound, which is the focus of code 00.09, so it is incorrect.}} {[}...{]} \textbf{\textcolor{red}{Wait, but just to double-check, what's the exact description of 00.09?}} I think it's ``Other therapeutic ultrasound procedures." \textbf{\textcolor{red}{Option C concerns prosthetics or orthotics, which are unrelated to therapeutic ultrasound, so it is incorrect.}} {[}...{]} \textbf{Final Answer}: B, because code 00.09 specifically refers to other therapeutic ultrasound procedures.
 & 80 \\
\midrule
2 & Backtracks, jumps back and forth between multiple uncertain options before ultimately landing on the wrong one. & {[}...{]} I know that 77.24 is a code for "Removal of vaginal packing" or "Removal of intravenous catheter". \textbf{\textcolor{red}{Wait, no—77.24 is actually for "Removal of vaginal packing."}} But I need to make sure. \textbf{\textcolor{red}{Alternatively, perhaps it's a biopsy}}. In ICD-9PROC, codes for biopsies are often in the 28–29 series, but for specific areas, they might be under the body system. \textbf{\textcolor{red}{Wait, maybe 77.2x is for procedures on the Bartholin's gland}}. {[}...{]}  B is the correct answer and 77.24 is a removal of a vaginal packing. & 20 \\
\bottomrule
\end{tabular}
\end{table}

\begin{figure*}[t]
\centering
\begin{tikzpicture}

\begin{axis}[
    name=pmRaw,
    width=7cm, height=5.2cm,
    ybar, bar width=10pt,
    ymin=0, ymax=1.15,
    ytick={0,0.2,0.4,0.6,0.8,1.0},
    enlarge x limits=0.18,
    symbolic x coords={2,3,4,5,6},
    xtick=data,
    xlabel={\texttt{Depths}},
    ylabel={Path Matching},
    title={\textbf{Path Matching} (raw) $\uparrow$},
    title style={font=\small},
    legend style={
        at={(0.96,0.96)}, anchor=north east,
        font=\footnotesize, draw=none, fill=none,
        /tikz/every even column/.append style={column sep=6pt}
    },
    legend columns=2,
    ymajorgrids, grid style={gray!20},
    tick label style={font=\footnotesize},
    label style={font=\footnotesize},
]
\addplot[fill=v3blue, draw=v3blue!80!black] coordinates {
    (2,0.843) (3,0.757) (4,0.797) (5,0.730) (6,0.640)
};
\addplot[fill=r1coral, draw=r1coral!80!black] coordinates {
    (2,0.760) (3,0.757) (4,0.745) (5,0.701) (6,0.694)
};
\legend{ds\_v3, ds\_r1}
\node[font=\fontsize{6}{7}\selectfont, xshift=-4pt, yshift=1pt, anchor=south] at (axis cs:2,0.843) {0.843};
\node[font=\fontsize{6}{7}\selectfont, xshift=-6pt, yshift=2.2pt, anchor=south] at (axis cs:3,0.757) {0.757};
\node[font=\fontsize{6}{7}\selectfont, xshift=-6pt, yshift=0.2pt, anchor=south] at (axis cs:4,0.797) {0.797};
\node[font=\fontsize{6}{7}\selectfont, xshift=-6pt, yshift=0.1pt, anchor=south] at (axis cs:5,0.730) {0.730};
\node[font=\fontsize{6}{7}\selectfont, xshift=-6pt, yshift=0.5pt, anchor=south] at (axis cs:6,0.640) {0.640};

\node[font=\fontsize{6}{7}\selectfont, xshift=6pt, yshift=0.3pt, anchor=south] at (axis cs:2,0.760) {0.760};
\node[font=\fontsize{6}{7}\selectfont, xshift=6pt, yshift=0pt, anchor=south] at (axis cs:3,0.757) {0.757};
\node[font=\fontsize{6}{7}\selectfont, xshift=6pt, yshift=0pt, anchor=south] at (axis cs:4,0.745) {0.745};
\node[font=\fontsize{6}{7}\selectfont, xshift=6pt, yshift=0pt, anchor=south] at (axis cs:5,0.701) {0.701};
\node[font=\fontsize{6}{7}\selectfont, xshift=6pt, yshift=0pt, anchor=south] at (axis cs:6,0.694) {0.694};
\end{axis}

\begin{axis}[
    name=pmGap,
    at={($(pmRaw.east)+(1.7cm,0)$)}, anchor=west,
    width=7cm, height=5.2cm,
    ybar, bar width=14pt,
    ymin=-0.12, ymax=0.08,
    ytick={-0.10,-0.08,-0.06,-0.04,-0.02,0,0.02,0.04,0.06},
    scaled y ticks=false,
    yticklabel style={
        /pgf/number format/fixed, 
        /pgf/number format/precision=2, 
        /pgf/number format/zerofill
    },
    enlarge x limits={abs=0.7cm},
    symbolic x coords={2,3,4,5,6},
    xtick={2,3,4,5,6},
    xlabel={\texttt{Depths}},
    ylabel={Gap (R1 $-$ V3)},
    title={\textbf{Path Matching} gap},
    title style={font=\small},
    ymajorgrids, grid style={gray!20},
    extra y ticks={0}, extra y tick labels={},
    extra y tick style={grid=major, grid style={black, thick}},
    tick label style={font=\footnotesize},
    label style={font=\footnotesize},
]

\addplot[ybar, bar width=14pt, bar shift=0pt, draw=black!40, fill=gapred] coordinates {
    (2,-0.082) (3,-0.000) (4,-0.052) (5,-0.029) (6,nan)
};
\addplot[ybar, bar width=14pt, bar shift=0pt, draw=black!40, fill=gapgreen] coordinates {
    (2,nan) (3,nan) (4,nan) (5,nan) (6,0.054)
};

\node[font=\tiny] at (axis cs:2,-0.094) {$-$0.083};
\node[font=\tiny] at (axis cs:3, 0.012) {$-$0.000};
\node[font=\tiny] at (axis cs:4,-0.064) {$-$0.052};
\node[font=\tiny] at (axis cs:5,-0.041) {$-$0.029};
\node[font=\tiny] at (axis cs:6, 0.066) {+0.054};
\node[font=\tiny, gray] at (axis cs:2,-0.110) {n=100};
\node[font=\tiny, gray] at (axis cs:3,-0.110) {n=100};
\node[font=\tiny, gray] at (axis cs:4,-0.110) {n=100};
\node[font=\tiny, gray] at (axis cs:5,-0.110) {n=100};
\node[font=\tiny, gray] at (axis cs:6,-0.110) {n=100};
\end{axis}

\begin{axis}[
    name=accRaw,
    at={($(pmRaw.south)+(0,-1.9cm)$)}, anchor=north,
    width=7cm, height=5.2cm,
    ybar, bar width=10pt,
    ymin=0, ymax=1.15,
    ytick={0,0.2,0.4,0.6,0.8,1.0},
    enlarge x limits=0.18,
    symbolic x coords={2,3,4,5,6},
    xtick=data,
    xlabel={\texttt{Depths}},
    ylabel={Accuracy},
    title={\textbf{Accuracy} (raw) $\uparrow$},
    title style={font=\small},
    ymajorgrids, grid style={gray!20},
    tick label style={font=\footnotesize},
    label style={font=\footnotesize},
]
\addplot[fill=v3blue, draw=v3blue!80!black] coordinates {
    (2,0.920) (3,0.950) (4,0.970) (5,0.970) (6,0.930)
};
\addplot[fill=r1coral, draw=r1coral!80!black] coordinates {
    (2,0.930) (3,0.980) (4,0.920) (5,0.900) (6,0.910)
};
\node[font=\fontsize{6}{7}\selectfont, xshift=-6pt, yshift=0pt, anchor=south] at (axis cs:2,0.920) {0.920};
\node[font=\fontsize{6}{7}\selectfont, xshift=-4pt, yshift=0pt, anchor=south] at (axis cs:3,0.950) {0.950};
\node[font=\fontsize{6}{7}\selectfont, xshift=-4pt, yshift=2pt, anchor=south] at (axis cs:4,0.970) {0.970};
\node[font=\fontsize{6}{7}\selectfont, xshift=-6pt, yshift=0pt, anchor=south] at (axis cs:5,0.970) {0.970};
\node[font=\fontsize{6}{7}\selectfont, xshift=-6pt, yshift=0pt, anchor=south] at (axis cs:6,0.930) {0.930};

\node[font=\fontsize{6}{7}\selectfont, xshift=8pt, yshift=0pt, anchor=south] at (axis cs:2,0.930) {0.930};
\node[font=\fontsize{6}{7}\selectfont, xshift=8pt, yshift=0pt, anchor=south] at (axis cs:3,0.980) {0.980};
\node[font=\fontsize{6}{7}\selectfont, xshift=7pt, yshift=0.2pt, anchor=south] at (axis cs:4,0.920) {0.920};
\node[font=\fontsize{6}{7}\selectfont, xshift=7pt, yshift=0.2pt, anchor=south] at (axis cs:5,0.900) {0.900};
\node[font=\fontsize{6}{7}\selectfont, xshift=7pt, yshift=0.2pt, anchor=south] at (axis cs:6,0.910) {0.910};
\end{axis}

\begin{axis}[
name=accGap,
at={($(accRaw.east)+(1.7cm,0)$)}, anchor=west,
width=7cm, height=5.2cm,
ybar, bar width=14pt,
ymin=-0.10, ymax=0.06,
ytick={-0.09,-0.06,-0.03,0,0.03,0.06},
scaled y ticks=false,
yticklabel style={
    /pgf/number format/fixed,
    /pgf/number format/precision=2,
    /pgf/number format/zerofill
},
enlarge x limits={abs=0.7cm},
symbolic x coords={2,3,4,5,6},
xtick={2,3,4,5,6},
xlabel={\texttt{Depths}},
ylabel={Gap (R1 $-$ V3)},
title={\textbf{Accuracy} gap},
title style={font=\small},
ymajorgrids, grid style={gray!20},
extra y ticks={0}, extra y tick labels={},
extra y tick style={grid=major, grid style={black, thick}},
tick label style={font=\footnotesize},
label style={font=\footnotesize},
]

\addplot[ybar, bar width=14pt, bar shift=0pt, draw=black!40, fill=gapred] coordinates {
(2,nan) (3,nan) (4,-0.050) (5,-0.070) (6,-0.020)
};
\addplot[ybar, bar width=14pt, bar shift=0pt, draw=black!40, fill=gapgreen] coordinates {
(2,0.010) (3,0.030) (4,nan) (5,nan) (6,nan)
};

\node[font=\tiny] at (axis cs:2, 0.018) {+0.010};
\node[font=\tiny] at (axis cs:3, 0.038) {+0.030};
\node[font=\tiny] at (axis cs:4,-0.062) {$-$0.050};
\node[font=\tiny] at (axis cs:5,-0.082) {$-$0.070};
\node[font=\tiny] at (axis cs:6,-0.032) {$-$0.020};
\node[font=\tiny, gray] at (axis cs:2,-0.092) {n=100};
\node[font=\tiny, gray] at (axis cs:3,-0.092) {n=100};
\node[font=\tiny, gray] at (axis cs:4,-0.092) {n=100};
\node[font=\tiny, gray] at (axis cs:5,-0.092) {n=100};
\node[font=\tiny, gray] at (axis cs:6,-0.092) {n=100};
\end{axis}

\end{tikzpicture}
\caption{%
  Path matching score and final-answer accuracy of DeepSeek-V3 (Instruct)
  versus DeepSeek-R1 (Reasoning) on DBO-CAR, stratified by
  \texttt{depths}.
}
\label{fig:dbo_car_v3_r1}
\end{figure*}
\end{document}